%% file: main.tex
\documentclass[sigconf, nonacm]{acmart}

\newcommand\vldbdoi{XX.XX/XXX.XX}
\newcommand\vldbpages{XXX-XXX}
\newcommand\vldbvolume{15}
\newcommand\vldbissue{1}
\newcommand\vldbyear{2021}
\newcommand\vldbauthors{\authors}
\newcommand\vldbtitle{\singlelinetitle} 
\newcommand\vldbavailabilityurl{}
\newcommand\vldbpagestyle{plain} 
\input{defines.tex}
\begin{document}
\title{Certifiable Machine Unlearning for Linear Models \\ {\large [Experiments, Analysis \& Benchmarks]}}
\newcommand\singlelinetitle{Certifiable Machine Unlearning for Linear Models}
%%
%% The "author" command and its associated commands are used to define the authors and their affiliations.
\author{Ananth Mahadevan}
\affiliation{%
  \institution{University of Helsinki}
  % \city{Helsinki}
  % \country{Finland}
}
\email{ananth.mahadevan@helsinki.fi}

\author{Michael Mathioudakis}
\affiliation{%
  \institution{University of Helsinki}
  % \city{Helsinki}
  % \country{Finland}
}
\email{michael.mathioudakis@helsinki.fi}

%%
%% The abstract is a short summary of the work to be presented in the
%% article.
\begin{abstract}
\input{abstract.tex}
\end{abstract}

\maketitle

%%% do not modify the following VLDB block %%
%%% VLDB block start %%%
\pagestyle{\vldbpagestyle}
\begingroup\small\noindent\raggedright\textbf{PVLDB Reference Format:}\\
\vldbauthors. \vldbtitle. PVLDB, \vldbvolume(\vldbissue): \vldbpages, \vldbyear.\\
\href{https://doi.org/\vldbdoi}{doi:\vldbdoi}
\endgroup
\begingroup
\renewcommand\thefootnote{}\footnote{\noindent
This work is licensed under the Creative Commons BY-NC-ND 4.0 International License. Visit \url{https://creativecommons.org/licenses/by-nc-nd/4.0/} to view a copy of this license. For any use beyond those covered by this license, obtain permission by emailing \href{mailto:info@vldb.org}{info@vldb.org}. Copyright is held by the owner/author(s). Publication rights licensed to the VLDB Endowment. \\
\raggedright Proceedings of the VLDB Endowment, Vol. \vldbvolume, No. \vldbissue\ %
ISSN 2150-8097. \\
\href{https://doi.org/\vldbdoi}{doi:\vldbdoi} \\
}\addtocounter{footnote}{-1}\endgroup
%%% VLDB block end %%%

%%% do not modify the following VLDB block %%
%%% VLDB block start %%%
\ifdefempty{\vldbavailabilityurl}{}{
\vspace{.3cm}
\begingroup\small\noindent\raggedright\textbf{PVLDB Artifact Availability:}\\
The source code, data, and/or other artifacts have been made available at \url{\vldbavailabilityurl}.
\endgroup
}
%%% VLDB block end %%%
\sloppy 
\input{introduction.tex}

\input{related_work.tex} 
\input{unlearning.tex}
\input{setup.tex}
\input{deletion_distributions.tex}

\input{experiments.tex}
\input{retrain.tex}
\input{conclusion.tex}

\begin{submission}
  \clearpage
\end{submission}
\bibliographystyle{ACM-Reference-Format}
\bibliography{references}
\begin{fullpaper}
\clearpage
\appendix
\input{appendix.tex}
\end{fullpaper}
\end{document}

%% file: defines.tex
\usepackage{xspace}
\usepackage{amsmath}
\usepackage{balance}
\usepackage[show]{chato-notes}
\usepackage{subcaption}
\usepackage{multirow}
\usepackage{verbatim}
\usepackage{booktabs}
\usepackage{algorithmic}
\usepackage[ruled,linesnumbered]{algorithm2e}
\usepackage{tikz}
\usepackage{color}
\usepackage{multirow}
\usepackage{tabularx}
\usepackage{tikz}
\usetikzlibrary{calc,positioning,shapes.geometric}
\usepackage{cleveref}
\crefname{equation}{Eq.}{Eqs.}
\Crefname{equation}{Equation}{Equations}

\newif\iffullpaper

\fullpapertrue

\iffullpaper
\includecomment{fullpaper}
\excludecomment{submission}
\else
\excludecomment{fullpaper}
\includecomment{submission}
\fi

\newcommand{\para}[1]{\noindent\textbf{#1}}
\newcommand{\spara}[1]{\smallskip\noindent\textbf{#1}}

\DeclareMathOperator*{\argmin}{arg\,min}

\newcommand{\squishlist}{
 \begin{list}{$\bullet$}
  {  \setlength{\itemsep}{0pt}
     \setlength{\parsep}{3pt}
     \setlength{\topsep}{3pt}
     \setlength{\partopsep}{0pt}
     \setlength{\leftmargin}{2em}
     \setlength{\labelwidth}{1.5em}
     \setlength{\labelsep}{0.5em}
} }

\newcommand*{\squishend}{
  \end{list}
}

\newcommand*{\error}{\ensuremath{\accDis}\xspace}
\newcommand*{\accuracy}{\ensuremath{\acctest}\xspace}

\newcommand*{\D}{\ensuremath{\mathcal{D}}\xspace}
\newcommand*{\sample}{\ensuremath{\mathbf{x}}\xspace}
\newcommand*{\dimensions}{\ensuremath{d}\xspace}
\newcommand*{\labels}{\ensuremath{\mathbf{y}}\xspace}

\newcommand*{\Dtrain}{\ensuremath{\D_{\text{init} }}\xspace}
\newcommand*{\Dm}{\ensuremath{\mathcal{D}_{\nremovals}}\xspace}
\newcommand*{\Dprime}{\ensuremath{\D\setminus\Dm}\xspace}
\newcommand*{\Dtest}{\ensuremath{\mathcal{D}_{\text{test}}}\xspace}
\newcommand*{\nremovals}{\ensuremath{m}\xspace}
\newcommand*{\ntrain}{\ensuremath{n_{\text{init}}}\xspace}
\newcommand*{\ntest}{\ensuremath{n}_{\text{test}}\xspace}
\newcommand*{\obj}{\ensuremath{L}\xspace}

\newcommand*{\UM}{\ensuremath{u}\xspace}
\newcommand*{\ml}{ML\xspace}
\newcommand*{\sgd}{SGD\xspace}
\newcommand*{\w}{\ensuremath{\mathbf{w}}\xspace}
\newcommand*{\worig}{\ensuremath{\mathbf{w}^{*}}\xspace}

\newcommand*{\wEmployed}{\ensuremath{\w}\xspace}
\newcommand*{\wunlearned}{\ensuremath{\w^\UM}\xspace}

\newcommand*{\QoA}{\ensuremath{\tau}\xspace}
\newcommand*{\noiseParamter}{\ensuremath{\sigma}\xspace}
\newcommand*{\noisyobj}{\ensuremath{\obj_{\sigma}}\xspace}
\newcommand*{\infl}{\textsc{Influence}\xspace}

\newcommand*{\fisher}{\textsc{Fisher}\xspace}
\newcommand*{\fMatrix}{\ensuremath{F}\xspace}

\newcommand*{\deltagrad}{\textsc{DeltaGrad}\xspace}
\newcommand*{\iw}{\ensuremath{\w}\xspace}
\newcommand*{\dgapprox}{\textsc{DGApprox}\xspace}

\newcommand*{\bmnist}{\ensuremath{\text{\sc mnist}^{\text{b}}}\xspace}
\newcommand*{\mnist}{\text{\sc mnist}\xspace}
\newcommand*{\covtype}{\text{\sc covtype}\xspace}
\newcommand*{\higgs}{\text{\sc higgs}\xspace}
\newcommand*{\cifar}{\text{\sc cifar2}\xspace}
\newcommand*{\eps}{\text{\sc epsilon}\xspace}
\newcommand*{\T}{\top}

\newcommand*{\noiseb}{\ensuremath{\mathbf{b}}\xspace}

\newcommand{\sape}{\ensuremath{\text{\tt SAPE}}\xspace}
\newcommand{\acctest}{\ensuremath{\text{\tt Acc}_{\text{test}}}\xspace}
\newcommand{\accremoved}{\ensuremath{\text{\tt Acc}_{\text{del}}}\xspace}
\newcommand*{\accDrop}{\texttt{AccErr}\xspace}

\newcommand*{\accDis}{\texttt{AccDis}\xspace}

\newcommand{\unirand}{\text{{\tt uniform-random}}\xspace}
\newcommand{\targrand}{\text{{\tt targeted-random}}\xspace}
\newcommand{\uniinfo}{\text{{\tt uniform-informed}}\xspace}
\newcommand{\targinfo}{\text{{\tt targeted-informed}}\xspace}

\newcommand*{\winit}{\ensuremath{\worig_{\text{init}}}\xspace}
\newcommand*{\accdropinit}{\ensuremath{\accDrop_{\text{init}}}\xspace}
\newcommand*{\acctestinit}{\ensuremath{\acctest^{\text{init}}}\xspace}
\newcommand*{\slope}{\ensuremath{c}\xspace}
\newcommand*{\errorPred}{\ensuremath{\widetilde{\error}}\xspace}

%% file: abstract.tex
Machine unlearning is the task of updating machine learning (\ml) models after a subset of the training data they were trained on is deleted.
Methods for the task are desired to combine \emph{effectiveness} and \emph{efficiency}, i.e., they should effectively `unlearn' deleted data, but in a way that does not require excessive computational effort (e.g., a full retraining) for a small amount of deletions.
Such a combination is typically achieved by tolerating some amount of approximation in the unlearning.
In addition, laws and regulations in the spirit of ``the right to be forgotten'' have given rise to requirements for \emph{certifiability}, i.e., the ability to demonstrate that the deleted data has indeed been unlearned by the \ml model.

In this paper, we present an experimental study of the three state-of-the-art approximate unlearning methods for linear models and demonstrate the trade-offs between efficiency, effectiveness and certifiability offered by each method.
In implementing the study, we extend some of the existing works and describe a common \ml pipeline to compare and evaluate the unlearning methods on six real-world datasets and a variety of settings.
We provide insights into the effect of the quantity and distribution of the deleted data on \ml models and the performance of each unlearning method in different settings.
We also propose a practical online strategy to determine when the accumulated error from approximate unlearning is large enough to warrant a full retraining of the \ml model. 

%% file: introduction.tex
\section{Introduction}
\label{sec:introduction}
%%% The task and why it is important
\emph{Machine unlearning} is the task of updating a machine learning (\ml) model after the partial deletion of data on which the model had been trained, so that the model reflects the remaining data.
The task arises in the context of many database applications that involve training and using an \ml model while allowing data deletions to occur.
For example, consider an online store that maintains a database of ratings for its products, and uses the database to train a model that predicts customer preferences (e.g., a logistic regression model that predicts what rating a customer would assign to a given product).
If part of the database is deleted (e.g., if some users request their accounts to be removed), then a problem arises: how to update the \ml model to ``unlearn'' the deleted data.
It is crucial to address the problem appropriately, so that the computational effort for unlearning is in proportion to the effect of the deletion: a tiny amount of deletion should not trigger a full retraining of the \ml model, leading to potentially huge data-processing costs; but at the same time, data deletions should not be ignored to such extent that the \ml model does not reflect the remaining data anymore.

%%% What we do and why
In this work, we perform a comparative analysis of existing methods for {machine unlearning}. 
In doing so, we are motivated both by the practical importance of the task and the lack of a comprehensive comparison in the literature.
Our goal is to compare the performance of existing methods in a variety of settings in terms of certain desirable qualities.
%

%%% Explain what we do: the main qualities of interest
What are those qualities?
First, machine unlearning should be \textbf{efficient}, i.e., achieving small running time, and \textbf{effective}, i.e., achieving good accuracy. 
%
% However, there is a trade-off between these two qualities.
% %
% On one extreme, a brute-force approach is to simply retrain the \ml model after every deletion, using all the remaining data.
% %
% While this approach is effective and achieves good accuracy, it is costly to perform a complete retrain for every deletion, particularly if only a minor fraction of the data is deleted.
% %
% On the other extreme, a naive approach is to keep the \ml model unaltered even after the deletion of data.
% %
% While this approach is efficient, as it requires no effort, it is also ineffective, as the \ml model may no longer reflect the remaining data after the deletions.
% %
% Ideally, one would use a method that falls between these two extremes, combining effectiveness with efficiency.
%
Moreover, machine unlearning is sometimes required to be \textbf{certifiable}, i.e., guarantee that after data deletion the \ml model operates \emph{as if} the deleted data had never been observed.
Such a requirement may be stipulated by laws, e.g., in the spirit of the \emph{right to be forgotten} \cite{manteleroRightToBeForgtten} or the \emph{right of erasure} \cite{gdprLegislation} in EU laws; or even offered voluntarily by the application in order to address privacy concerns.
In the example of the online store, consider the case where some users request their data to be removed from its database: the online store should not only delete the data in the hosting database, but also ensure that the data are unlearned by any \ml model that was built from them.
Essentially, if an audit was performed, the employed \ml models should be found to have unlearned the deleted data as well as a model that is obtained with a brute-force, full retraining on the remaining data -- even if full retraining was not actually performed to unlearn the deleted data.
%
% In what follows, the amount of tolerated approximation is expressed as a threshold \error for the dissimilarity between the currently \emph{employed} and \emph{fully retrained} \ml models.
%

The aforementioned qualities exhibit pairwise trade-offs.
There is a trade-off between efficiency, on one hand, and effectiveness or certifiability on the other: that's because it takes time to optimize a model so as to reflect the underlying data or unlearn the deleted data.
%
% Moreover, certifiability determines how much room there is to trade between effectiveness and efficiency: smaller values of \error by definition require the currently employed model to imitate the performance of a fully retrained model, thus limiting how much efficiency one could potentially  gain at the expense of effectiveness. 
%
Moreover, there is a trade-off between certifiability and effectiveness: that's because unlearning the deleted data, thus ensuring certifiability, corresponds to learning from fewer data, thus decreasing accuracy.
In this study, we observe the three trade-offs experimentally -- and find that, because the compared methods involve different processing costs for different operations, they offer better or worse trade-offs in different settings.

\begin{figure*}
    \input{figures/pipeline.tex}
    \caption{The common \ml pipeline with the three stages of \emph{training}, \emph{inference} and \emph{unlearning}. First, an initial model \worig is trained on all data and used for inference; subsequently, whenever a part \Dm of the data is deleted, an updated model \wunlearned is obtained via machine unlearning. The pipeline restarts if the updated model is deemed inadequate.}
    \label{fig:pipeline}
\end{figure*}
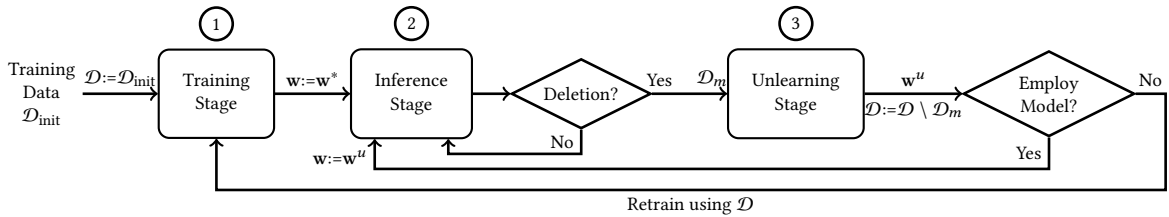

%%% Explain how we do it
% In this work, we set out to explore and demonstrate the performance of different approaches for machine unlearning in terms of efficiency, effectiveness, and certifiability, and the trade-offs they offer.
%
For the experimental evaluation, we implement a common \emph{\ml pipeline} (Figure~\ref{fig:pipeline}) for the compared methods.
The first stage trains an \emph{initial} \ml model from the data.
To limit the variable parts of our experimentation, we will be focusing on \textbf{linear classification} models, such as logistic regression, as they represent a large class of models that are commonly encountered in a wide range of settings.
In addition, we'll be assuming that the initial model is trained with \textbf{stochastic gradient descent} (\sgd), since \sgd and its variants are the standard algorithms for training general \ml models.
The second stage \emph{employs} the initial \ml model for inference, i.e., for classification.
During this stage, if data deletion occurs, then the pipeline proceeds to the third stage to unlearn the deleted data and produce an updated model.
%
% Taking into account the data deletion, the application updates the employed model using a method for machine unlearning.
%
After every such model update, the updated model is evaluated for certifiability.
If it fails, then the pipeline restarts and trains a new model from scratch on the remaining data; otherwise, it is employed in the inference stage and pipeline resumes.
%
% The evaluation ensures the employed model is certifiable and will pass in an audit.
%
When an audit is requested by an external auditor (not shown in Figure~\ref{fig:pipeline}) -- a full retraining of the \ml model is executed on the remaining data, and the \emph{fully retrained} model is compared to the currently employed model.
If the employed model is found to have unlearned the deleted data as well as the fully retrained model (within a threshold of disparity), then the audit is successful, meaning the pipeline has certifiably unlearned the deleted data so far and is allowed to resume.
%
% Otherwise, if the audit fails, it signifies that pipeline is has not unlearned the deleted data and certifiability was not ensured.
%
% Obviously, and in line with our earlier discussion, it is desirable to use a method for machine unlearning that is effective (leads to a model with good accuracy), efficient (executes the model updates fast) 

Given this pipeline, we evaluate three methods, namely \fisher, \infl, and \deltagrad, that follow largely different approaches for machine unlearning and represent the state of the art for our setting (linear classification models trained with \sgd).
\fisher updates the initial \ml model using the \emph{remaining} data to perform a corrective Newton step; it follows \citet{golatkarEternalSunshineSpotless2020}.
\infl updates the initial \ml model using the \emph{deleted} data to perform a corrective Newton step; it is defined in \citet{guoCertifiedDataRemoval2020}.
And \deltagrad updates the initial \ml model by correcting the \sgd steps that led to the initial model; it follows the method defined in \citet{wuDeltaGradRapidRetraining2020}.
Note that, in this work, we extend the original papers of \citet{wuDeltaGradRapidRetraining2020} and \citet{golatkarEternalSunshineSpotless2020}, so as to ensure that all the evaluated methods are equipped with mechanisms to control the trade-offs between efficiency, effectiveness and certifiability.
%
% -- which, however, we extend in this paper with a mechanism to trade effectiveness for certifiability.
%
% For a more elaborate discussion of the literature, see Section~\ref{sec:related}.
%

For the experimental evaluation, we implement the three methods and compare them in a large range of settings that adhere to the pipeline described above.
The aim of the experiments is to demonstrate the trade-offs that the three methods offer in terms of efficiency, effectiveness and  certifiability.
%
% In this regard, we evaluate three main aspects of the \ml pipeline.
%
% \textcolor{red}{First, we evaluate the effect of the quality and quantity of the deleted data on fully retrained models and then select these values for each dataset to facilitate comparison between unlearning methods.}
%
% Second, we evaluate the effect of the two mechanisms independently from each other that control the trade-offs between efficiency, effectiveness for each unlearning method.
%
% {We identify the influence of the  dimensionality of the dataset in the amount of efficiency traded-off for certifiability and the the influence of the noise injection mechanism on the decrease in efficiency for each method.}
%
% Finally, we compare the joint effect of both mechanisms on all the three trade-offs between efficiency, effectiveness and certifiability for each unlearning method.
First, we demonstrate that the trade-offs are much more pronounced for certain worst-case deletion distributions than for random deletions.
Subsequently, we observe that \fisher offers overall best certifiability, along with good effectiveness at lower efficiency than \infl especially for larger datasets; \infl offers overall best efficiency, along with good effectiveness at lower levels of certifiability; and \deltagrad offers stable albeit lower performance across all qualities.
Moreover, we observe that the efficiency of \fisher and \infl is much higher for datasets of lower dimensionality.
The patterns we observe in these experiments have a beneficial by-product: they allow us to define a practical approach to determine in online fashion (i.e., as the pipeline unfolds) when the accumulated error from approximate unlearning is large enough to require restarting the pipeline to perform a full retraining of the ML model.
%
% We propose a simple linear estimation of the certifiability error \error during the evaluation of the updated model, that allows us to decide when the pipeline should restart and obtain a new fully trained model.
%

To summarize, we make the following contributions:
\squishlist
    \item We define a novel framework to compare machine unlearning methods in terms of effectiveness, efficiency, and certifiability.
    \item We extend the methods of \citet{wuDeltaGradRapidRetraining2020} and \citet{golatkarEternalSunshineSpotless2020} with mechanisms to control performance trade-offs. 
    \item We offer the first experimental comparison of the competing methods in a large variety of settings. As an outcome, we obtain novel empirical insights about (1) the effect of the deletion distribution on the performance trade-offs, (2) the strengths of each method in terms of performance trade-offs.
    \item We propose a practical online strategy to determine an optimal time when to restart the training pipeline.
\squishend

As future work, a similar experimental study would address model updates for data addition rather than deletion.
For this work, we opted to focus on deletion to keep the paper well-contained, because certifiability is typically required in the case of deletion (e.g., when users request their data to be deleted from an application) and the methods we evaluate are tailored to certifiable deletion.

%% file: figures/pipeline.tex
\usetikzlibrary{shapes.geometric}

% positioning style from https://tex.stackexchange.com/questions/102250/how-to-position-one-node-relative-to-another-node-at-a-certain-angle-in-tikz-tak
\tikzset{
        position/.style args={#1:#2 from #3}{
            at=(#3.#1), anchor=#1+180, shift=(#1:#2)
        }
    }

% database shape from from https://tex.stackexchange.com/questions/442991/database-shape-in-tikz

\makeatletter
\tikzset{
    database top segment style/.style={draw},
    database middle segment style/.style={draw},
    database bottom segment style/.style={draw},
    database/.style={
        path picture={
            \path [database bottom segment style]
                (-\db@r,-0.5*\db@sh) 
                -- ++(0,-1*\db@sh) 
                arc [start angle=180, end angle=360,
                    x radius=\db@r, y radius=\db@ar*\db@r]
                -- ++(0,1*\db@sh)
                arc [start angle=360, end angle=180,
                    x radius=\db@r, y radius=\db@ar*\db@r];
            \path [database middle segment style]
                (-\db@r,0.5*\db@sh) 
                -- ++(0,-1*\db@sh) 
                arc [start angle=180, end angle=360,
                    x radius=\db@r, y radius=\db@ar*\db@r]
                -- ++(0,1*\db@sh)
                arc [start angle=360, end angle=180,
                    x radius=\db@r, y radius=\db@ar*\db@r];
            \path [database top segment style]
                (-\db@r,1.5*\db@sh) 
                -- ++(0,-1*\db@sh) 
                arc [start angle=180, end angle=360,
                    x radius=\db@r, y radius=\db@ar*\db@r]
                -- ++(0,1*\db@sh)
                arc [start angle=360, end angle=180,
                    x radius=\db@r, y radius=\db@ar*\db@r];
            \path [database top segment style]
                (0, 1.5*\db@sh) circle [x radius=\db@r, y radius=\db@ar*\db@r];
        },
        minimum width=2*\db@r + \pgflinewidth,
        minimum height=3*\db@sh + 2*\db@ar*\db@r + \pgflinewidth,
    },
    database segment height/.store in=\db@sh,
    database radius/.store in=\db@r,
    database aspect ratio/.store in=\db@ar,
    database segment height=0.1cm,
    database radius=0.25cm,
    database aspect ratio=0.35,
    database top segment/.style={
        database top segment style/.append style={#1}},
    database middle segment/.style={
        database middle segment style/.append style={#1}},
    database bottom segment/.style={
        database bottom segment style/.append style={#1}}
}
\makeatother

% \tikzset{font=\Large}
\begin{tikzpicture}[line width=1pt,scale=0.8, every node/.style={scale=0.8}]
    % helpers 
    % \node[circle,draw=red] at (0,0) {};
    % \draw [help lines] grid (3,2);
    % nodes 
    \node[align=center] (db1) at (0,0) {Training \\Data\\ \Dtrain} ;
    \node [
        rectangle,
        right=1cm of db1,thick,
        draw=black,
        rounded corners,
        inner sep=0.4cm,
        align=center,
    ] (training) {Training\\Stage};
    \node [
        circle,
        draw=black,
        above=0.1cm of training
    ] (step1) {1};
    \node [
        rectangle,
        right=1cm of training,
        thick,draw=black,
        rounded corners,
        inner sep=0.4cm,
        align=center
    ] (inference) {Inference\\Stage};
    \node [
        circle,
        draw=black,
        above=0.1cm of inference
    ] (step2) {2};
    % decision diamond
    \node [
        draw,
        diamond,
        aspect=2,
        right=0.5cm of inference,
        align=center
    ] (deletion) {Deletion?};
    \node [
        rectangle,
        right=1cm of deletion ,
        thick,
        draw=black,
        rounded corners,
        inner sep=0.4cm,
        align=center
        ] (unlearning) {Unlearning\\Stage};
    \node [
            circle,
            draw=black,
            above=0.1cm of unlearning
        ] (step3) {3};
    % decision diamond
    \node [draw, diamond, aspect=2, right=1.3cm of unlearning,align=center] (employ) {Employ\\Model?};

    % lines
    \draw[->] (db1) -- (training) node[midway,above] {\D:=\Dtrain};
    \draw[->] (training) -- (inference) node[midway,above] {\wEmployed:=\worig};
    \draw[->] (inference) -- (deletion);
    \draw[->]  (deletion) -- node[pos=0.5,left] {No} ++(0,-1cm) -| ($(inference.south west)!.8!(inference.south east)$) ;
    \draw[->] (deletion) -- (unlearning) node[pos=0.1,above] {Yes} node[pos=0.8,above] {\Dm};
    \draw[->] (unlearning) -- (employ) node[midway,above] {\wunlearned};
    \draw[draw=none] (unlearning) -- (employ) node[midway,below]  {\D:=\Dprime};
    \draw[->]  (employ) -- node[pos=0.5,left] {Yes} ++(0,-1.25cm) -| ($(inference.south west)!.2!(inference.south east)$) node[pos=0.7,left] {\wEmployed:=\wunlearned};
    \draw[->]  (employ.east) -- node[pos=0.5,above] {No} ++(0.5cm,0cm) -- ++(0,-1.6cm) -| (training.south) node[pos=0.25,below]{Retrain using \D};
    
\end{tikzpicture}

%% file: related_work.tex
\section{Related Work}
\label{sec:related}
Unlearning methods are classified as \emph{exact} or \emph{approximate}.

\para{Exact unlearning methods} produce \ml models that perform as fully retrained models.
By definition, these methods offer the highest certifiability as the produced models are effectively the same as ones obtained with retraining.
There exist several exact unlearning methods, typically for training algorithms that are model-specific and deterministic in nature.
For instance, \ml models such as support vector machines \cite{tsaiIncrementalDecrementalTraining2014,poggio2000incremental,karasuyamaMultipleIncrementalDecremental2010}, collaborative filtering, naive bayes \cite{schelterAmnesiaMachineLearning2020,caoMakingSystemsForget2015} $k$-nearest neighbors and ridge regression \cite{schelterAmnesiaMachineLearning2020} possess exact unlearning methods.
The efficiency for such exact methods varies.

For stochastic training algorithms such as \sgd, \citet{bourtouleMachineUnlearning2020} propose an exact unlearning approach, under the assumption that learning is performed in federated fashion.
In federated learning, separate \ml models are trained on separate data partitions and their predictions are aggregated during inference.
This partitioning of data allows for efficient retraining of \ml models on smaller fragments of data, leading to efficient unlearning when data are deleted.
However, for general \ml models trained with \sgd, the setting of federated learning comes with a potential cost on effectiveness that is difficult to quantify and control, because model optimization is not performed jointly on the full dataset.

% Exact methods are typically model-specific. Moreover, in practice, real-world models are trained with general optimization methods like \sgd,  has not , we choose to analyse approximate unlearning methods in this paper.

\para{Approximate unlearning methods} produce \ml models that are an approximation of the fully retrained model.
These methods typically aim to offer much larger efficiency through the relaxation of the effectiveness and certifiability requirements.
Most of them can be categorized into one of three groups.

The \emph{first group} \cite{golatkarEternalSunshineSpotless2020,golatkarForgettingOutsideBox2020,golatkarMixedPrivacyForgettingDeep2020} uses the remaining data of the training dataset to update the \ml model and control certifiability.
These methods use Fisher information \cite{jamesFisherMatrix2020} to retrain information of the remaining data and inject optimal noise in order to unlearn the deleted data.
The \emph{second group}~\cite{guoCertifiedDataRemoval2020,izzoApproximateDataDeletion2020,chaudhuriPrivacyPreservingLogistic2009} uses the deleted data to update \ml models during unlearning.
They perform a Newton step \cite{kohUnderstandingBlackboxPredictions2017} to approximate the influence of the deleted data on the \ml model and remove it.
To trade-off certifiability for effectiveness, they inject random noise to the training objective function \cite{chaudhuriPrivacyPreservingLogistic2009}.
The \emph{third group} \cite{wuDeltaGradRapidRetraining2020,wuPrIUProvenanceBasedApproach2020,neelDescenttoDeleteGradientBasedMethods2020,gravesAmnesiacMachineLearning2020} stores data and information during training and then utilize this when deletion occurs to update the model.
Specifically, these methods focus on approximating the \sgd steps that would have occurred if full retraining was performed.
To aid in this approximation, they store the intermediate quantities (e.g., gradients and model updates) produced by each \sgd step during training.
The amount of stored information and the approximation process raise an effectiveness vs efficiency trade-off.

Methods from the above three groups can be used to perform unlearning for classification models with \sgd, as long as the relevant quantities (e.g., the model gradients) are easy to compute for the model at hand. Apart from the above three groups, there are other approximate unlearning methods that do not fit the same template -- e.g., methods for specific \ml models, such as \citet{brophyDARTDataAddition2020} for random forest models, or for Bayesian modelling, such as \citet{nguyenVariationalBayesianUnlearning} for Bayesian learning -- and so we consider them outside the scope of this paper.

\para{In this paper}, we focus on approximate unlearning methods, because they are applicable to general \ml models, when training is performed with general and widely used optimization algorithms like \sgd. We implement three methods, \fisher, \infl, and \deltagrad which correspond to state-of-the-art unlearning methods from each of the aforementioned groups, \citet{golatkarEternalSunshineSpotless2020}, \citet{guoCertifiedDataRemoval2020}, and \citet{wuDeltaGradRapidRetraining2020} respectively.

%% file: unlearning.tex
\section{Machine Unlearning}
\label{sec:unlearning-methods}
%First paragraph
In this section, we present the common \ml pipeline over which the unlearning methods are evaluated in this study (Section~\ref{sec:pipeline}).
%
% What does a method consist of  
Subsequently, we describe the three unlearning methods (Sections~\ref{subsec:fisher}-\ref{subsec:guo}). 
Any of them can be used in the \ml pipeline to update the \ml model at the event of data deletion.
Lastly, we discuss the process of auditing the \ml pipeline for certifiability (Section~\ref{sec:audit}).

% Training stage and objective function
\subsection{The \ml pipeline}
\label{sec:pipeline}

The \ml pipeline describes the lifecycle of \ml models in our experimental framework -- i.e., how a model is trained, employed for inference, updated incrementally, and potentially fully retrained from scratch, while a series of data deletions occur (see Figure~\ref{fig:pipeline}).

\para{First stage: Training}
In this stage a \ml model is learned from the training dataset \D. 
In what follows, we'll assume that the \ml model is logistic regression, a simple and widely used model for classification.
Each data point consists of $d$ features $\sample \in \mathbb{R}^{\dimensions}$ and a categorical label \labels\ --- and we assume that initially there are $\ntrain$ entries in total, i.e., $\D = \Dtrain =\left\{\left(\sample_{i},\labels_{i}\right)\right\}_{i=1}^{\ntrain}$.
At any time in the lifecycle of the pipeline, \D will denote the \emph{currently available} training dataset, which is a subset of initial training data due to possible deletions, i.e., $\D \subseteq \Dtrain$.
%
% Training an \ml model over a dataset \D is achieved by an optimization algorithm that selects the model parameters \w that optimize an objective function \obj.
%
An \emph{objective function} measures the fitness of a \ml model's parameters \w on a dataset \D.
Following common practice for logistic regression, the objective function is
\begin{equation}
    \obj = \obj(\w,\D) = \frac{1}{|\D|}\sum_{i\in\D}\ell(\w^{T}\sample_{i},\labels_{i}) + \frac{\lambda}{2}\|\w\|_{2}^{2}, \label{eq:objectiveFunc}
\end{equation}
where the first term captures the average classification loss, with $\ell$ as the binary cross entropy for logistic regression; 
and the second term quantifies ridge regularization for a fixed value of parameter $\lambda$, the role of which is to prevent over-fitting.

Moreover, for training, we'll be using mini-batch \sgd, a general and widely-used optimization algorithm~\cite{boyd2004convex}.
\sgd iteratively minimizes the objective function over the training data: first, it initializes the model parameters to a random value $\w := \w_{0}$, and it improves them in iterative steps as follows,
\begin{equation}
    \w_{t+1} := \w_{t} - \eta_{t}\nabla \obj(\w_{t},\D)\label{eq:sgd-step}
\end{equation}
where $\eta_{t}$ is the learning rate at iteration $t$.
An appropriate number of iterations is taken for convergence, after which the resulting \ml model minimizes\footnote{For general models \sgd requires a few re-runs to limit the possibility of returning a local optimum with low objective value. However for logistic regression, used in this paper, the objective function is convex and \sgd leads to the global optimum.} the objective function \obj on the dataset \D.
Following common practice, \sgd is executed in mini-batch fashion, i.e., only a subset of $\D$ is used in each execution of \Cref{eq:sgd-step}.

As we'll see in the upcoming sections~\ref{subsec:fisher}-\ref{subsec:guo}, each unlearning method uses an adaptation of the objective function of \cref{eq:objectiveFunc}.
A model obtained from the training stage of the \ml pipeline is denoted\footnote{Slightly abusing notation, we use \w for both a model instance and its parameters.} with \worig.
When it is obtained using the initial dataset \Dtrain, then \worig is referred to as the \emph{initial trained} model; and when it is obtained using a subset $\D\subseteq\Dtrain$ of the training dataset, then it is referred to as the \emph{fully retrained} model.
This model, $\w:=\worig$ is sent to the second stage to be employed for inference.

\para{Second stage: Inference} The available model \w is employed for inference, i.e., to predict the class \labels of arbitrary data points \sample submitted as queries to the \ml model.
At any time during the second stage, a subset of the data may be deleted, which prompts the pipeline to proceed to the third stage.

% unlearning stage and unlearning algorithm
\para{Third stage: Unlearning} The third stage receives the currently employed model \wEmployed and the deleted subset of the training data, denoted with $\Dm = \left\{ \left( \sample_{i},\labels_{i} \right) \right\}_{i=1}^{\nremovals}$.
It executes the \emph{unlearning algorithm} so as to ``unlearn'' the deleted data \Dm.
The result of the unlearning is an {updated} model \wunlearned.

Once obtained, the updated model \wunlearned is evaluated on a test dataset, $\Dtest = \left\{ \left( \sample_{i},\labels_{i} \right) \right\}_{i=1}^{n_{\text{test}}}$ in terms of effectiveness and certifiability.
Following common practice, the test dataset \Dtest is disjoint from the training dataset \D\ --- and in real scenarios, \Dtest is typically independently collected, e.g., consisting of user queries (i.e., the data points \sample for which the \ml model is asked to predict \labels).

Effectiveness is measured as the model's accuracy on the test dataset \accuracy (i.e., the fraction of test data that it classifies correctly).
Furthermore, certifiability is measured as the  disparity \error between the updated \wunlearned and the fully retrained \worig \ml model, in terms of accuracy over the deleted data \Dm.
Intuitively, the disparity \error captures the amount of information that the updated model \wunlearned possesses about the deleted data \Dm in comparison to a fully retrained model \worig: if the disparity is small, then the updated model has `unlearned' the deleted data as well as a model that is retrained from scratch on the remaining data.
Note however that, while \wunlearned is readily available from the execution of the unlearning algorithm, the fully retrained model \worig is not: in fact, obtaining \worig after every deletion would defeat the purpose of obtaining \wunlearned in the first place.
Therefore, unlike our experimental study, in a practical setting disparity \error could not be directly measured exactly but it should be estimated.
% We note however that, in some practical scenarios, access to the deleted data \Dm may not be feasible, e.g., if they are deleted immediately from the database, before the certifiability of the updated model is assessed.
%
To deal with this challenge, we will experimentally show how to estimate \error using \accuracy which is easily obtainable (see \cref{sec:when-to-retrain}).

After evaluation on the test dataset, a decision is made about the updated model.
If its accuracy \accuracy is sufficiently high and disparity \error sufficiently low relative to some thresholds (determined, e.g., by the administrator of this pipeline), then the pipeline returns to the second stage and employs the updated model \w:=\wunlearned for inference -- otherwise, the pipeline returns to the first stage for a full retraining over the remaining data \Dprime.
Intuitively, the full retraining is triggered once a large volume of deletions lead to an updated model with degraded effectiveness or certifiability.

\spara{Controlling the trade-offs}
Each unlearning method is equipped with mechanisms to navigate trade-offs between efficiency, effectiveness, and certifiability.
%
% This trade-off then determines how much time the unlearning stage of the pipeline takes and the chances of the updates model passing evaluation and being employed. 
% %
% For instance, if we trade effectiveness or certifiability for efficiency, then we obtain an updated model quickly, however, it may not pass the evaluation and the pipeline would restart.
% %
% Alternatively, if we trade efficiency for effectiveness or certifiability, the unlearning stage takes longer to produce an updated model and it is more likely to pass evaluation and be employed for inference.
%
% Specifically, each unlearning method is associated with two parameters to control the trade-offs, namely the Quality of Approximation (\QoA) parameter and the noise parameter. 
%
The first mechanism trades efficiency, on one hand, for effectiveness and certifiability, on the other, and is controlled via an \emph{efficiency parameter} \QoA, specified separately for each unlearning method.
Lower values of \QoA indicate lower efficiency, and thus allowing longer running times to improve effectiveness and certifiability of the updated model.

The second mechanism trades effectiveness (high accuracy \accuracy ) for certifiability (low disparity \error).
This trade-off arises because of two reasons. 
First, because unlearning the deleted data, thus achieving low disparity \error, is equivalent to learning from fewer data, thus leading to lower accuracy \accuracy. 
Second, because while the unlearning algorithms aim to operate more efficiently than a full retraining, they are challenged to distinguish what part of the model should be unlearned (corresponding to deleted data) and what remembered (corresponding to remaining data). 
So, on one extreme, one may opt to ensure unlearning the deleted data (achieving low \error), at the potential cost of also (mistakenly) ``forgetting'' some of the remaining data (further decreasing accuracy \accuracy) -- or the opposite, on the other extreme.

For all unlearning methods in this paper, the trade-off is controlled via noise injection, and specifically via a \emph{noise parameter} \noiseParamter that determines the amount of injected noise.
Simply expressed, noise injection deliberately adds randomness to an \ml model, both during training and unlearning.
On one end of this trade-off, when large amounts of noise are injected, the predictions of the \ml model are effectively random -- therefore ensuring low disparity \error and high certifiability, but at the cost of low effectiveness, as the noise leads unlearning all the data.
On the other end, when no noise is injected, the unlearning method strives to optimize the objective function over the remaining data, thus prioritising effectiveness.

We note that the trade-off between effectiveness and certifiability, along with noise injection as a control mechanism, have already been introduced as concepts in the literature.
For further discussion on noise injection, we refer the interested reader to~\cite{golatkarEternalSunshineSpotless2020,chaudhuriPrivacyPreservingLogistic2009,guoCertifiedDataRemoval2020}.

% The two parameters \QoA and \noiseParamter give rise to the third trade-off between certifiability and efficiency.
% %
% First, increasing \noiseParamter and injecting more noise leads to higher certifiability at the cost of the computations required to inject the noise decreasing efficiency. 
% %
% Second, decreasing \QoA produces an updated model \wunlearned that better approximates the fully retrained model \worig, decreasing disparity \error, leading to higher certifiability at the cost of efficiency.
%
% This gain in effectiveness is directly correlated with higher certifiability because \wunlearned and \worig are less dissimilar.
%
% As we will see in the following sections, the computational complexity of noise injection mechanism of the unlearning methods described in this paper are not dependent on the value of \noiseParamter.
%
% Therefore, the certifiability-efficiency trade-off reported in the experiments arises only due to the \QoA parameter and is directly related to the effectiveness-efficiency trade-off. 

\medskip
Having defined the \ml pipeline, we now proceed to specify the unlearning methods.
For each unlearning method, we describe its three main components, namely the \emph{training algorithm} (used to train a model on a dataset), the \emph{unlearning algorithm} (used for incremental model updates after deletion), and the \emph{parameters} that control trade-offs between efficiency, effectiveness and certifiability.

\input{golatkar.tex}

\input{guo.tex}
\input{deltagrad.tex}

\input{audit.tex}

%% file: golatkar.tex
\subsection{\fisher Unlearning Method}
\label{subsec:fisher}
The \fisher unlearning method is described in \citet{golatkarEternalSunshineSpotless2020}.

\para{The training algorithm} for this method proceeds in two steps: in the first step, it invokes \sgd to optimize the objective \obj (Eq.~\ref{eq:objectiveFunc}); and in the second step it performs noise injection.
The output model \worig is expressed as
\begin{equation}
    \worig :=  \w^{opt} + \sigma \fMatrix^{-1/4}\noiseb, \label{eqn:fisher-train}
\end{equation}
where,
\begin{align}
    \w^{opt} &= \argmin^{\sgd}_{\w} \obj\left( \w,\D \right),\label{eqn:optimized-model}\\
    \fMatrix &= \nabla^{2}\obj(\w^{opt},\D)\label{eqn:fisher-matrix-training},\\
    \noiseb &\sim \mathcal{N}(0,1)^{\dimensions}.\label{eqn:fisher-noise}
\end{align}
As shown in \Cref{eqn:optimized-model}, $\w^{opt}$ is the model that optimizes the objective function \obj using \sgd.
Moreover, \fMatrix is the Fisher matrix of $L$, defined as the covariance of the objective function. 
For logistic regression, \fMatrix is equal to the Hessian of $L$, as reflected in Equation~\eqref{eqn:fisher-matrix-training}.
The second term in \Cref{eqn:fisher-train} corresponds to the noise injection that adds standard normal noise (see \cref{eqn:fisher-noise}) to the optimal model $\w^{opt}$ in the direction of the Fisher matrix.

\para{The unlearning algorithm}
takes as input the currently employed model \w, the deleted subset of the training data $\Dm \subset \D$, and outputs an updated model \wunlearned given by
\begin{equation}
    \wunlearned :=\underbrace{ \wEmployed - \fMatrix^{-1}\Delta}_{\text{Newton Correction}} + \underbrace{\noiseParamter\fMatrix^{-1/4}\noiseb }_{\text{Noise Injection}}, \label{eq:fisherUnlearning}
\end{equation}
where 
\begin{align}
    \Delta &= \nabla \obj(\wEmployed,\Dprime)\label{eqn:fish-gradient},\\
    \fMatrix &= \nabla^{2}\obj(\wEmployed,\Dprime)\label{eqn:fisher-matrix},
\end{align}
and \noiseb is the same as in \Cref{eqn:fisher-noise}.
As shown in Equation~\eqref{eqn:fish-gradient}, $\Delta$ is the gradient of the objective function $L$ (Eq.~\ref{eq:objectiveFunc}).
And, similar to \Cref{eqn:fisher-train}, \fMatrix is the Fisher matrix, now computed on the remaining training data after deletion (\Dprime).
The first term in Equation~\eqref{eq:fisherUnlearning} corresponds to the corrective Newton step that aims to unlearn the deleted data \Dm.
The second term corresponds to noise injection, and adds standard normal noise $\noiseb$ (see Eq.~\eqref{eqn:fisher-noise}) to the updated model \wunlearned in the direction of the Fisher matrix (see \Cref{eqn:fisher-matrix}).

% \note[AM]{Shoule we just use \QoA to denote $\nremovals^{\prime}$ for \fisher \& \infl and $T_0$ for \deltagrad? Or is it better to have them unique to imply that they cannot be set to the same value?}
As defined in Equation~\eqref{eq:fisherUnlearning}, the unlearning algorithm computes an updated model in a single step.
A more elaborate approach is to split the deleted data in mini-batches of size $\nremovals^{\prime}\leq\nremovals$ and use Equation~\eqref{eq:fisherUnlearning} sequentially for each of them.
This approach leads to multiple and smaller corrective Newton steps, which in turn lead to a more effective \ml model at the cost of efficiency.
For this experimental study, we'll be using this mini-batch version of the unlearning algorithm, as shown in Algorithm~\ref{alg:fisher-mini-batch}.
\input{fisher_algorithm.tex}

\para{Trade-off parameters} 
As explained earlier (Section~\ref{sec:pipeline}), the noise parameter \noiseParamter controls the trade-off between effectiveness and certifiability.
Moreover, the size of the mini-batches $\QoA_{_{\fisher}} = \nremovals^{\prime}$ serves as the efficiency parameter that controls the trade-offs between efficiency, on one hand, and effectiveness and certifiability, on the other.
The lowest efficiency is achieved when $\nremovals^{\prime}=1$, i.e., unlearning one deleted data point at a time incrementally -- however, this comes at the massive cost of recomputing the Fisher matrix after every single deleted data point.
The highest efficiency is achieved when $\nremovals^{\prime}=\nremovals$, i.e., unlearning all deleted data at once -- which comes at the cost of effectiveness due to a single and crude corrective Newton step.
In typical real settings, one would choose a value $\nremovals^{\prime}$ between the two extremes.

%% file: fisher_algorithm.tex
\begin{algorithm}%[h!] 
    \small
    % \footnotesize
    \SetKwInOut{Input}{Input}
    \SetKwInOut{Output}{Output}
    \Input{Employed model \wEmployed, Current training data \D, Deleted data \Dm, Parameter \noiseParamter, Mini-batch size $\nremovals^{\prime}$, objective function \obj }
    \Output{Updated model parameters \wunlearned}
    $s \leftarrow \left\lceil \frac{\nremovals}{\nremovals^{\prime}}\right\rceil$; Split \Dm into $s$ mini-batches $\{\D_{\nremovals^{\prime}}^1,\D_{\nremovals^{\prime}}^2,\dots,\D_{\nremovals^{\prime}}^s\}$

    $\D^{\prime} \leftarrow \D$; $\wunlearned \leftarrow \wEmployed$

    \For{$i=1;i<=s; i++$}{

        $\D^\prime \leftarrow \D^{\prime}
        \setminus \D_{\nremovals^{\prime}}^{i}$; $\Delta \leftarrow \nabla \obj\left(\wunlearned,\D^{\prime}\right)$; $\fMatrix \leftarrow \nabla^{2}\obj\left( \wunlearned,\D^{\prime} \right)$

        $\wunlearned \leftarrow \wunlearned - \fMatrix^{-1}\Delta$

        \If{$\noiseParamter>0$}{
            \label{alg:fisher-alg-noise}
            Sample $\noiseb \sim \mathcal{N}(0,1)^{d}$
            
            $\wunlearned \leftarrow \wunlearned + \noiseParamter\fMatrix^{-1/4}\noiseb$
        }
    
    }
    \Return \wunlearned
    \caption{\fisher mini-batch}
    \label{alg:fisher-mini-batch}
    \end{algorithm}

%% file: guo.tex
\subsection{\infl Unlearning Method}
\label{subsec:guo}
The \infl unlearning method is follows \citet{guoCertifiedDataRemoval2020}.
Its approach is based on \ml influence theory \cite{kohUnderstandingBlackboxPredictions2017}.
At a high level, unlearning is performed by computing the influence of the deleted data on the parameters of the trained \ml model and then updating the parameters to remove that influence. 
%
% Influence theory literature  describes methods to approximate the \textit{influence} of a training data point on the parameters of a \ml model.
%
% They also propose injecting random noise to the objective function rather than the \ml model parameters (as seen in \fisher and \deltagrad) to trade effectiveness for certifiability.
%
% We first describe the training algorithm for the \infl method.
%
% noisy objective function
Moreover, it uses a modified objective function that incorporates noise injection: 
\begin{equation}
    \label{eqn:noisyObjectiveFunc}
    \noisyobj(\w;\D) = \obj(\w,D) + \frac{\sigma\noiseb^{\T}\w}{|\D|},
\end{equation}
where $\obj$ and \noiseb are the same as in \Cref{eq:objectiveFunc,eqn:fisher-noise} respectively.
The second term in \Cref{eqn:noisyObjectiveFunc} describes the noise injection where \noiseParamter is the noise parameter.
The amount of noise is scaled wrt the size of the training data \D.

%Training 
%
\para{The training algorithm} uses \sgd to optimize the noisy objective.
\begin{equation}
    \worig := \argmin_{\w}^{\sgd} \noisyobj\left( \w,\D \right)\label{eqn:influence-train}
\end{equation}
Note that, when \noiseParamter is increased, the effectiveness of the \ml model decreases as the \sgd algorithm prioritizes minimizing the second term in Equation~\eqref{eqn:noisyObjectiveFunc} rather than the original objective function captured by the first term.
%
% A benefit of such a noise injection mechanism is the ability to select smaller values of the regularization parameter $\lambda$ (see \cref{eq:objectiveFunc}) without incurring a large cost to the effectiveness of the \ml model \cite{chaudhuriPrivacyPreservingLogistic2009}.

% Unlearning algorithm
\para{The unlearning algorithm} approximates the \emph{influence} of the deleted subset $\Dm\subset \D$ on the parameters of the currently employed model \wEmployed and performs the update as:
\begin{equation}
    \wunlearned := \wEmployed + H^{-1}\Delta^{(\nremovals)}, \label{eqn:influenceUnlearning}
\end{equation}
where 
\begin{align}
    \Delta^{(\nremovals)} &= \nabla \obj(\wEmployed,\Dm)\label{eqn:influence-gradient},\\
    H &= \nabla^{2}L(\wEmployed,\Dprime).\label{eqn:influence-hessian}
\end{align}
As seen in \cref{eqn:influence-gradient,eqn:influence-hessian}, $\Delta^{(m)}$ is the gradient of the objective function \obj (see \cref{eq:objectiveFunc}) computed on the deleted data and $H$ is the Hessian matrix computed on the remaining training data.
The second term in \Cref{eqn:influenceUnlearning} is known as the \textit{influence function} of the deleted data \Dm on the model parameters \wEmployed.

Similar to \fisher, when the unlearning algorithm is performed in mini-batches of $\nremovals^{\prime}\leq\nremovals$, we obtain a more effective \ml model at the cost of the efficiency.
This is because, we compute the influence function on smaller mini-batches of deleted data multiple times.
For this experimental study, we'll be using this mini-batch version of the unlearning algorithm, as shown in \Cref{alg:influence-mini-batch}
\input{influence_algorithm.tex}

\para{The trade-off parameters} are similar to those in the \fisher method.
The size of $\QoA_{_{\infl}} = \nremovals^{\prime}$ serves as the efficiency parameter and \noiseParamter as the noise parameter.

%% file: influence_algorithm.tex
\begin{algorithm}%[h!] 
    \small
    % \footnotesize
    \SetKwInOut{Input}{Input}
    \SetKwInOut{Output}{Output}
    \Input{Employed model \wEmployed, Current training data \D, Deleted data \Dm, Mini-batch size $\nremovals^{\prime}$, objective function \obj }
    \Output{Updated model parameter \wunlearned}
    $s \leftarrow \left\lceil \frac{\nremovals}{\nremovals^{\prime}}\right\rceil$; 
    Split \Dm into $s$ mini-batches $\{\D_{\nremovals^{\prime}}^1,\D_{\nremovals^{\prime}}^2,\dots,\D_{\nremovals^{\prime}}^s\}$

    $\D^{\prime} \leftarrow \D$; $\wunlearned \leftarrow \wEmployed$

    \For{$i=1;i<=s; i++$}{

        $\D^\prime \leftarrow \D^{\prime}
        \setminus \D_{\nremovals^{\prime}}^{i}$

        $\Delta^{(\nremovals^{\prime})} \leftarrow \nabla \obj\left(\wunlearned,\D_{\nremovals^{\prime}}^{i}\right)$; $H \leftarrow \nabla^{2}\obj\left( \wunlearned,\D^{\prime} \right)$
        
        $\wunlearned \leftarrow \wunlearned + H^{-1}\Delta^{(\nremovals^{\prime})}$
    
    }
    \Return \wunlearned
    \caption{\infl mini-batch}
    \label{alg:influence-mini-batch}
    \end{algorithm}

%% file: deltagrad.tex
\subsection{\deltagrad Unlearning Method}
\label{subsec:deltagrad}
The \deltagrad unlearning method is described in \citet{wuDeltaGradRapidRetraining2020}.
Its approach is to approximate the \sgd steps that would have happened if the deleted data had not been present, using the information from the initial \sgd training steps.
%

% Training algorithm 
\para{The training algorithm} uses \sgd followed by noise injection,
\begin{equation}
    \worig := \argmin_{\w}^{\sgd} \obj\left( \w,\D \right) + \noiseParamter\cdot\noiseb. \label{eqn:deltagrad-train}
\end{equation}
where \noiseb is defined as in \cref{eqn:fisher-noise}). 
This noise injection mechanism is a Gaussian version of the one described by \citet{wuPrIUProvenanceBasedApproach2020} using results from \citet{dwork2014algorithmic}.
In contrast to \fisher method's noise injection (Eq.~\ref{eq:fisherUnlearning}), there is no Fisher matrix to guide the random Gaussian noise in this mechanism.
Therefore, a large value of $\sigma$ will indiscriminately remove information from the employed model which in turn drastically reduces the effectiveness of the \ml model.

At every iteration of the \sgd algorithm (\cref{eq:sgd-step}), the parameters $\w_{t}$ and objective function gradients $\nabla\obj(\w_{t},\D)$ are \textit{stored to disk}.
%
% The resulting \ml model becomes the trained model of the \ml pipeline, i.e.\ $\w:=\worig$ (Figure~\ref{fig:pipeline})

% Unlearning algorithm
\para{The unlearning algorithm} for this method proceeds in two steps: in the first step, it approximately updates the stored sequence $(\iw_t)$ of parameters computed by \sgd; in the second step, it injects noise.
In summary, and slightly abusing notation, we write
\begin{equation}
    \wunlearned := \dgapprox(\{\iw_t\}) + \noiseParamter\cdot\noiseb
    \label{eqn:deltagradUnlearning}
\end{equation}
The first term corresponds to the approximate update of \sgd steps, and the second to noise injection, with \noiseb defined as in \cref{eqn:fisher-noise}.

Let us provide more details about how the first term is computed.
Upon the deletion of the current subset of the training data $\Dm \subset \D$, the unlearning algorithm aims to obtain, approximately, the \ml model that would have resulted from \sgd if \Dm had never been used for training.
By definition (Eq.~\ref{eq:sgd-step}), in the absence of \Dm, the \sgd steps would have been:
\begin{equation}
    \iw_{t+1} = \iw_{t} - \frac{\eta_{t}}{(n-\nremovals)}\left[ n\nabla \obj(\iw_{t},\D) -\nremovals\nabla \obj(\iw_{t},\Dm)  \right],
    \label{eq:deltagrad-leave-m-out}
\end{equation}
leading to a different sequence $(\iw_t)$ of model parameters than the one obtained before deletion from \cref{eq:sgd-step}.
As a consequence, the value of $\nabla \obj(\iw_{t};\D)$ differs between the executions of \cref{eq:sgd-step} (before deletion) and \cref{eq:deltagrad-leave-m-out} (after deletion).
\deltagrad's approach is to obtain a fast approximation of the latter from the former, thus approximately unlearning the deleted data without performing a full-cost \sgd on the remaining data.

\input{dg_algorithm.tex}
The unlearning algorithm is shown for reference in Algorithm~\ref{alg:deltagrad-algorithm}.
As seen in \cref{alg:deltagrad-approx-1,alg:deltagrad-approx-2,alg:deltagrad-approx-3,alg:deltagrad-approx-4,alg:deltagrad-approx-5}, the term $\nabla \obj(\iw_{t};\D)$ is approximated using the Quasi-Newton L-BFGS optimization algorithm with the terms $\w_{t}$ and $\nabla \obj(\w_{t},\D)$ that were stored during training.
However, there exist two issues with this approximation.
First, the L-BFGS algorithm requires a history of accurate computations to produce an effective approximation.
Second, consecutive approximations lead to errors accumulating after several iterations in \sgd.
The first issue is addressed by using a \emph{burn-in period} of $j_0$ iterations, during which the exact gradient on the remaining dataset, $\nabla \obj(\iw_{t},\Dprime)$, is computed.
The latter issue is addressed by periodically computing the exact gradient after every $T_0$ iterations (following the burn-in period).
These are seen in \cref{alg:deltagrad-exact-1,alg:deltagrad-exact-2,alg:deltagrad-exact-3,alg:deltagrad-exact-4,alg:deltagrad-exact-5}.
Moreover, in order to use the above \deltagrad algorithm for subsequent data deletions, the terms $\w_{t}$ and $\nabla \obj\left(\w_{t},\D\right)$ that were previously stored in disk are updating after unlearning the deleted data \Dm.
This is described in \cref{alg:deltagrad-exact-4,alg:deltagrad-exact-5} and \cref{alg:deltagrad-approx-4,alg:deltagrad-approx-5}.

\para{Trade-off parameters} The unlearning algorithm has several parameters that control its efficiency.
The burn-in period $j_0$, periodicity $T_0$, learning rate $\eta_{t}$, number of \sgd iteration $T$ and the length of historical computations for L-BFGS optimization algorithm are all potential \QoA parameters for the \deltagrad algorithm.
Due to page-limit constraints, we choose the periodicity $\QoA_{_\deltagrad} =  T_0$ as the primary efficiency parameter, while keeping all other secondary parameters fixed for a given dataset.
%
\begin{comment}
%
For instance, a large $\eta_{t}$ and long number of \sgd iterations $T$, amplify the approximation error in each step for many iterations reducing effectiveness.
%
Similarly, a small burn-in period and a large periodicity will offer a significant increase in efficiency, but at the cost of the effectiveness due to approximation errors accumulating in each iteration.
%
\end{comment}
%
We use $T_0=2$ as its lower value, which corresponds to computing the exact gradient every alternate iteration, leading to minimum efficiency.
Conversely, large values of $T_0$ lead to higher efficiency.
%
% This is due to more number of efficient L-BFGS approximation iterations used in Algorithm~\ref{alg:deltagrad-algorithm} before an inefficient exact gradient computation is needed.
% %
% However, this leads to accumulating approximation errors which results in a lower effectiveness of the \ml model.
Finally, the noise parameter \noiseParamter controls the trade-off between effectiveness and certifiability.

%% file: dg_algorithm.tex
\begin{algorithm}%[h!] 
    \small
    % \footnotesize
    \SetKwInOut{Input}{Input}
    \SetKwInOut{Output}{Output}
    \Input{Current training data \D, Deleted data \Dm, model weights saved during the training stage or updated later $\{\w_{0}, \w_{1}, \dots, \w_{t}\}$ and corresponding gradients $\{\nabla \obj\left(\w_{0},\D\right), \nabla \obj\left(\w_{1},\D\right), \dots, \nabla \obj\left(\w_{t},\D\right)\}$, period $T_0$, total iteration number $T$, ``burn-in'' iteration number $j_0$, learning rate $\eta_t$}
    \Output{Updated model parameter $\iw_{t}$}
    Initialize $\iw_{0} \leftarrow \w_{0}$
    
    \For{$t=0;t<T; t++$}{
    
    \eIf{$[((t-j_0) \mod T_0) == 0]$ or $t \leq j_0$} 
    { \label{alg:deltagrad-exact-1}
        compute $\nabla F\left(\iw_{t},\Dprime\right)$ exactly \label{alg:deltagrad-exact-2}
        
        compute $\iw_{t+1}$ by using exact update (\cref{eq:sgd-step}) \label{alg:deltagrad-exact-3}

        Update $\w_{t}$ with $\iw_{t}$ \label{alg:deltagrad-exact-4}

        Update $\nabla F\left(\w_{t},\D\right)$ with $\nabla F\left(\iw_{t},\Dprime\right)$ \label{alg:deltagrad-exact-5}
    }
    { 
        Approximate $\nabla \obj\left(\iw_{t},\Dprime\right)$ with L-BFGS algorithm using stored terms $\w_{t}$ and $\nabla \obj\left( \w_{t},\D \right)$ \label{alg:deltagrad-approx-1}
        
        Compute $\nabla \obj\left( \iw_{t},\Dm \right)$ \label{alg:deltagrad-approx-2}

        Compute $\iw_{t+1}$ via the modified gradient formula (\cref{eq:deltagrad-leave-m-out}) \label{alg:deltagrad-approx-3}

        Update $\w_{t}$ with $\iw_{t}$ \label{alg:deltagrad-approx-4}

        Update $\nabla F\left(\w_{t},\D\right)$ with approximated $\nabla F\left(\iw_{t},\Dprime\right)$ \label{alg:deltagrad-approx-5}
    }
    }
    
    \Return $\iw_{t}$
    \caption{\dgapprox}
    \label{alg:deltagrad-algorithm}
    \end{algorithm}

%% file: audit.tex
\subsection{Auditing}
\label{sec:audit}
% As discussed in \Cref{sec:introduction}, when deletions occur, a requirement of certifiability may be needed.
% %
% Certifiability ensures that after a subset of the current data $\Dm \subset \D$ is deleted, the currently employed model \wEmployed is similar to a fully retrained model \worig on the remaining training data \Dprime.
% %
% In particular, the \ml pipeline provides a maximum disparity \error, of the dissimilarity between the employed model \wEmployed and a fully retrained model \worig.
% %
% To verify these claims of certifiability, an audit may be requested at any time after deletions have occurred.

During an audit, the auditor first obtains a fully retrained model \worig using the training algorithm of the corresponding unlearning method (see \cref{eqn:fisher-train,eqn:deltagrad-train,eqn:influence-train}) on the available training data.
Next, the auditor measures the disparity \error between \worig and the currently employed model \wEmployed.
If the measured disparity does not exceed a given threshold, then \ml pipeline passes the audit and is allowed to resume.
Otherwise, the \ml pipeline does not satisfy the certifiability claimed and therefore fails the audit.
Such failed certifiability audits may result in fines or other regulatory issues.
Therefore, it is in the best interest of the deployer of the \ml pipeline to correctly state the certifiability requirements and ensure that the pipeline is able to pass an audit after any number of deletions.

%% file: setup.tex
\section{Experimental Setup}
\label{sec:evalutaion}

% intro paragraph
In this section, we describe the datasets, the implementation of the \ml pipeline and metrics we use for evaluation.

\begin{table}[h]
    \caption{Datasets}
    \label{tab:datasets}
    \begin{small}
        \begin{tabular}{lcccrr}
            \toprule
            \multirow{2}{*}{Dataset} & \multicolumn{2}{c}{Dimensionality} & Classes & Train Data    & Test Data    \\ 
            \cmidrule(l){2-3} 
            & d & level & $k$ & $\ntrain$ & $\ntest$ \\ 

            \midrule
            $\bmnist$ &  784 & moderate & 2       & 11\,982   & 1\,984      \\
            $\cifar$  & 3072 & high  & 2      & 20\,000   & 2\,000     \\
            $\mnist$  &  784 &moderate  & 10      & 60\,000   & 10\,000     \\
            $\covtype$    & 54 & low   & 2       & 522\,910  & 58\,102     \\
            $\eps$       &2000 & high   & 2       & 400\,000 & 100\,000\\
            $\higgs$        &28 & low   & 2       & 9\,900\,000 & 1\,100\,000\\
            \bottomrule
        \end{tabular}
    \end{small}
\end{table}

\subsection{Datasets}
\label{subsec:datasets}

We perform experiments over six datasets, retrieved from the public LIBSVM repository~\cite{libsvm}.
The datasets cover a large range of size and dimensionality, as summarily shown in Table~\ref{tab:datasets}, allowing us to effectively explore the trends and trade-offs of the unlearning methods.
In addition, to have a uniform experimental setting with comparable results, 
% we \emph{reduce} the number of considered classes to $2$ for some of the datasets, and
we focus on the task of binary classification.
Towards this end, most datasets were chosen to include $2$ predictive classes (or if the original dataset contained more classes, the experiments focused on two of them, as reported in Table~\ref{tab:datasets}).
Nevertheless, we also include one multi-class dataset (\mnist, with $10$ classes).

In more detail, 
{\bf\mnist}~\cite{mnist} consists of $28\times28$ black and white images of handwritten digits ($0$-$9$), each digit corresponding to one class.
{$\bmnist$} is the binary-class subset of the $\mnist$ dataset, consisting only of digits 3 and 8 for both training and test data.
\cifar consists of $32\times32$ RGB color images,  belonging to the ``cat'' or ``ship'' categories from the original ten category CIFAR-10~\cite{CIFAR10} dataset.
\covtype~\cite{covtype} consists of 54 cartographic features used to categorize forest cover types. 
We use the binary version from LIBSVM.
The \higgs~\cite{higgs} dataset consists of kinematic features from Monte Carlo simulation of particle detectors for binary classification.
\eps~\cite{epsilon} is obtained from the PASCAL Large Scale Learning Challenge 2008.

\subsection{ \ml Pipeline}
\label{subsec:pipeline-setup}
We now provide implementation details for the \ml pipeline (Figure~\ref{fig:pipeline}).
The pipeline is designed so that it is suitable to all the three chosen unlearning methods discussed in Section~\ref{sec:unlearning-methods}.
The pipeline is implemented in Python 3.6 using PyTorch 1.8 \cite{PytorchNEURIPS2019}.
All experiments are run on a machine with 24 CPU cores and 180 GB RAM.
Our full code base is publicly available\footnote{\url{https://version.helsinki.fi/mahadeva/unlearning-experiments}}.

\para{Preprocessing.} 
% The precursor stage of the \ml pipeline involves the data processing steps applied to all datasets described in \Cref{subsec:datasets}.
%
The \infl unlearning method requires all data points $\sample_{i}$ of a dataset to have a Euclidean norm at most 1, i.e., $\|\sample_i\|_2\leq 1$ (see \citet{guoCertifiedDataRemoval2020}).
To satisfy this requirement, we perform a max-$L_2$ normalization for all datasets as a pre-processing step, where we divide each data point with the largest $L_2$ norm of any data point in the dataset.
This normalization does not affect the performance of other methods. %, as it preserves the relative distances between data points.

\para{\bf Training.} As mentioned in \Cref{sec:unlearning-methods}, we use the mini-batch \sgd algorithm for training.
In all cases, we use fixed learning rate $\eta=1$ and ridge regularization parameter $\lambda=10^{-4}$.
Moreover, we use standard SGD since \deltagrad does not support momentum-based \sgd algorithms such as Adam~\cite{kingma2015Adam}.
Note that both the \fisher and \infl unlearning algorithms require the Hessian matrix to be positive definite to compute the inverse (see \cref{eq:fisherUnlearning,eqn:influenceUnlearning}).
This is ensured by running the \sgd algorithm for a sufficiently large number of iterations during training to achieve convergence. 
Towards this end, we use a small subset of the training data as a validation dataset, to identify an optimal mini-batch size and total number of \sgd iterations.
Moreover, we control the data points selected in each mini-batch by fixing the random seed used to produce mini-batches in the \sgd algorithm.
This ensures reproducibility of the experiments across various unlearning methods.
Finally, for multi-class classification with $k>2$ classes on the \mnist dataset, we train $k$ independent binary logistic regression classifiers in a One vs Rest (OVR) fashion.

\para{\bf Unlearning.} When a subset \Dm of the current training data \D is deleted, the unlearning algorithm of the employed method (\fisher, \deltagrad, or \infl) is invoked.
We modify and extend the code provided in \citet{guoCertifiedDataRemoval2020}\footnote{\url{https://github.com/facebookresearch/certified-removal}} to implement the \infl unlearning algorithm as described in Section~\ref{subsec:guo}.
We further extend this code to also implement the mini-batch version of the \fisher unlearning algorithm as seen in \Cref{subsec:fisher}.
For the \deltagrad method, we use the code provided by the authors in \citet{wuDeltaGradRapidRetraining2020}\footnote{\url{https://github.com/thuwuyinjun/DeltaGrad}} and modify it to add the noise injection mechanism described in Section~\ref{subsec:deltagrad} to trade-off effectiveness for certifiability.

\subsection{Evaluation Metrics}
\label{subsec:metrics}
In this section, we define the metrics we use to report the performance of different unlearning methods in terms of effectiveness, certifiability and efficiency.
For uniformity of presentation, we'll be reporting the performance achieved by a given model as relative to the performance of a baseline model.
Towards this end, we'll be using the \emph{Symmetric Absolute Percentage Error} (\sape) defined as
\begin{equation}
    \sape(a,b) = \frac{|b-a|}{|b|+|a|}\cdot 100\%.
    \label{eqn:sape}
\end{equation}
For the function to be continuous, we define \sape(0,0) = 0.

\spara{Effectiveness} is measured in terms of predictive \emph{accuracy}, i.e., as the fraction of data points correctly classified by a given \ml model on a particular dataset. 
We will write \acctest to denote the accuracy on the test dataset \Dtest and \accremoved to denote the accuracy on the deleted data \Dm.
Let $\acctest^{u}$ be the accuracy of the updated model \wunlearned on the test dataset; and $\acctest^{*}$ be the optimal accuracy that may be obtained via logistic regression on the same data (in other words, the latter is the test accuracy of the fully trained model with $\noiseParamter=0$).
We will report \accDrop as the error in test accuracy of the updated model \wunlearned compared to the optimal one, i.e.,
\begin{equation}
    \accDrop = \sape(\acctest^{*},~\acctest^{u}).\label{eqn:effectiveness-metric}
\end{equation}
A low value of \accDrop implies that the updated model \wunlearned is more effective, i.e., the predictive accuracy of the updated model is close to optimal for the available data.

\spara{Certifiability} 
is measured in terms of how well the updated model has unlearned the deleted data relatively to a fully retrained model by the same method.
Specifically, let $\accremoved^{u}$ and $\accremoved^{*}$ be the accuracy on the deleted data for the updated model and the fully retrained model, respectively, for the same noise value \noiseParamter.
We report \accDis as the disparity in accuracy of the two models, i.e.,
\begin{equation}
    \accDis = \sape(\accremoved^{*},~\accremoved^{u}).\label{eqn:certifiability-metric}
\end{equation}
A lower value of \accDis implies that the updated model \wunlearned has higher certifiability, i.e., the updated model is more similar to the fully retrained model, which had never seen the deleted data.
Note that the symmetry of \sape is essential here, because both under- and over-performance of the updated model contributes towards disparity wrt the fully retrained model.

\spara{Efficiency} 
is measured as the speed-up in running time to obtain the updated model \wunlearned relative to the running time to obtain fully retrained model: 
\begin{equation}
    \text{speed-up} = \frac{\text{time taken to obtain }\worig}{\text{time taken to obtain }\wunlearned}~\text{x}. \label{eqn:efficiency-metric}
\end{equation}
A speed-up of 2x indicates that the unlearning stage is able to produce an updated model twice as fast as it takes for the training stage takes to produce a fully retrained model.

\subsection{Experimental Roadmap}
\label{subsec:exp-roadmap}
In this subsection, we provide a brief overview of the experiments in the upcoming sections.
The experiments analyze the stages of the \ml pipeline presented in \Cref{sec:pipeline} in an incremental manner.

In \Cref{sec:deletion-distribution}, before we evaluate any unlearning methods, we explore how the quantity and quality of deleted data affect the accuracy of the fully retrained model.
Then, in \Cref{sec:results}, we demonstrate the trade-offs between efficiency, effectiveness and certifiability, for different values of the \QoA and \noiseParamter parameters for each unlearning method.
Finally, in \Cref{sec:when-to-retrain} we 
propose an evaluation strategy to decide whether the incrementally updated model produced by the unlearning method has diverged enough from the available data to warrant a full-retrain.

%% file: deletion_distributions.tex
\section{Effect of Deletion Distribution}
\label{sec:deletion-distribution}

Before we compare unlearning methods, let us explore how the volume and distribution of the deleted data affect the accuracy of fully trained models.
This will allow us to separate the effects of data deletion from the effects of a specific unlearning method.

% For the experiments of this section, we vary the volume of deleted data as a fraction of the initial training data size.
%
We implement a two-step process to generate different deletion distributions.
The process is invoked once for each deleted data point, for a predetermined number of deletions.
In the first step, one \emph{class} is selected.
For example, for binary-class datasets (see Table~\ref{tab:datasets}), the first step selects one of the two classes.
The selection may be either \emph{uniform}, where one of the $k$ classes is selected at random, each with probability $1/k$; or \emph{targeted}, where one class is randomly predetermined and subsequently always selected.
%
% When the deleted data points are equally likely to come from any class in a dataset, then it is called a uniform deletion distribution.
% %
% Conversely, if all the deleted data points belong to a particular class, then it is defined as a targeted deletion distribution and the targeted class is called the \emph{deleted class}.

Once the class has been selected in the first step, a data point from that class is selected in the second step. % is \emph{information-based} choice to select deleted data points of the selected class of data.
%
% The aim of the second step is to compromise the integrity of the \ml model through the deleted samples.
%
The selection may be either \emph{random}, where one data point is selected uniformly at random; or \emph{informed}, where one point is selected so as to decrease the model's accuracy the most.
Ideally, for the \emph{informed} selection, and for each data point, we would compute exactly the drop in the accuracy of a fully trained model on the remaining data after the single-point removal, and we would repeat this computation after every single selection.
In practice, however, such an approach would be extremely expensive computationally, even for experimental purposes.
Instead, for the \emph{informed} selection, we opt to heuristically select the outliers in the dataset, as quantified by the $L_2$ norm of each data point.
This heuristic is inspired by \citet{izzoApproximateDataDeletion2020}, who state that deleting data points with a large $L_2$ norm negatively affects the approximation of Hessian-based unlearning algorithms.

As described above, the two-step process yields four distinct deletion distributions namely \emph{\unirand}, \emph{\targrand}, \emph{\uniinfo} and \emph{\targinfo}.
In the experiments that follow, we select data to delete for different choices of deletion distribution and volume. %, the latter specified as a fraction of the total number of data points in the original training dataset. % 
For each set of deleted data, we report the accuracy of the fully trained model after deletion (this is the accuracy achieved by the model that optimizes \cref{eq:objectiveFunc} using \sgd).

\begin{figure*}
    \centering
    \includegraphics[width=\textwidth]{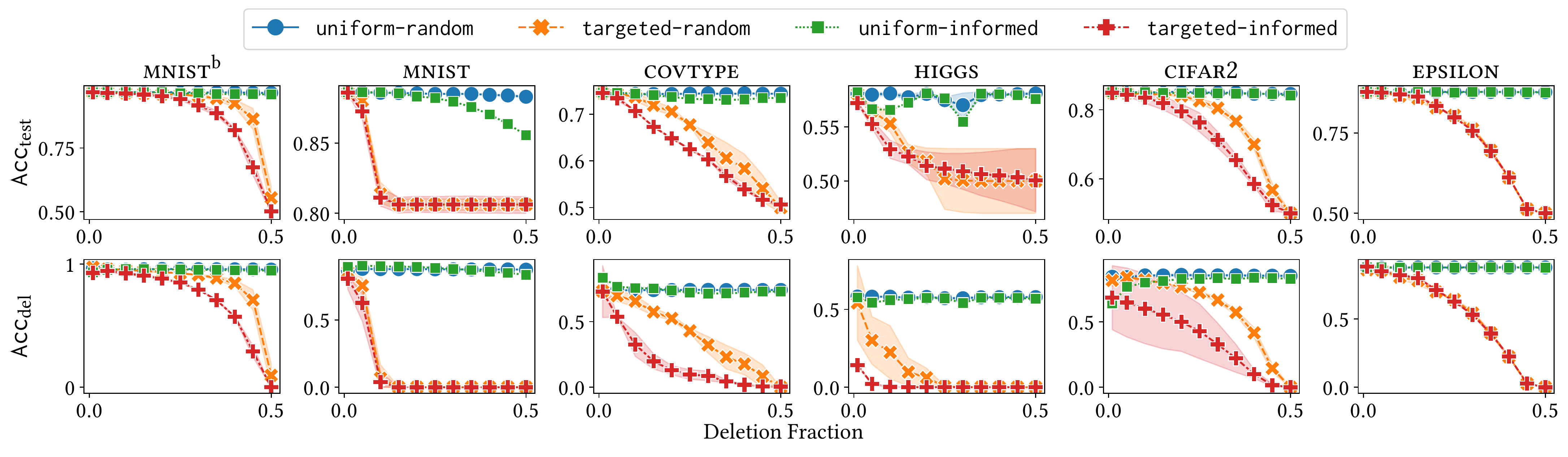}
    \caption{Effect of different deletion distributions on the test accuracy (\acctest) and deleted data accuracy (\accremoved) as the fraction of deleted data is varied for different datasets.}
    \label{fig:sampling-distributions}
\end{figure*}

The results are shown in Figure~\ref{fig:sampling-distributions}.
Each plot in the figure corresponds to one dataset.
The first row of plots reports the accuracy on the test dataset, and the second row on the deleted data.
Accuracy values correspond to the $y$-axis while the volume of deletion (as fraction of the original dataset size) to the $x$-axis.
Different deletion distributions are indicated with different markers and color.
% , we see the effect of the four types of deletion distribution on the performance of the fully retrained model \worig as the ratio of deleted data is varied.
%
The variance seen in \Cref{fig:sampling-distributions} is a consequence of the randomness in the selection of deleted points (2 random runs were performed).

There are three main takeaways from these results.
First, uniform deletion distributions (\unirand and \uniinfo) do not adversely affect the test accuracy of a fully retrained \ml model even at deletion fractions close to $0.5$.
\begin{fullpaper}
In fact, we see that the accuracy decreases only after more than $90\%$ of the data are deleted (see \cref{fig:extended-deletion-distribution}).
\end{fullpaper}
This is due to the redundancy present in real-world datasets, in the sense that only a small number of data points from each class is sufficient to separate the classes as well as possible.
And therefore, evaluating unlearning methods on deletions from uniform distributions will not offer significant insights on the effectiveness and efficiency trade-offs.
Moreover, notice that the test accuracy for the \uniinfo distribution is lower than the \unirand distribution, indicating that the informed deletions remove outlier data points that decrease the accuracy of the \ml model.

Second, targeted deletion distributions (\targrand and \targinfo) provide a worst-case scenario of deletions that leads to large drops in test accuracy.
This is because deleting data points from one targeted class eventually leads to class imbalance, and causes the \ml model to be less effective in classifying data points from that class.
\begin{fullpaper}
In addition, we observe that the variance resulting from the selection of the deleted class is low in all datasets apart from \higgs.
We postulate this is because on the particular way that missing values have been treated for this dataset: data point that have missing feature values disproportionately belong to class 1.
Therefore, this tends to cause a steeper drop in accuracy, when data from class 1 is targeted.
\end{fullpaper}
Next, we observe that the drop in accuracy is steeper for \targinfo compared to \targrand, indicating the deletion of informed points results in a less effective model at the same deletion fraction.
This highlights the \targinfo distribution as a worst-case deletion scenario: to validate their performance, machine unlearning methods should be tested on \targinfo or similar distributions, where data deletions quickly affect the accuracy of the learned model.

Thirdly, we see across deletion fractions that the accuracy on the test and deleted dataset (\acctest and \accremoved, respectively) follow a similar trend (i.e., their values are highly correlated).
Hence, the test accuracy \acctest, which can always be computed for a model on the test data, can be used as a good proxy for the \accremoved of a \ml model, which may be impossible to compute after data deletion but is required in order to assess certifiability.
This observation will be useful to decide when to trigger a model retraining in the \ml pipeline (\Cref{sec:when-to-retrain}).

\begin{fullpaper}
Additionally, we note that the rate of drop in \acctest and \accremoved with respect to deletion fraction varies from dataset to dataset.    
\end{fullpaper}

%% file: experiments.tex
\section{Experimental Evaluation}
\label{sec:results}
In this section, we demonstrate the trade-offs exhibited by the unlearning methods in terms of the qualities of interest (effectiveness, efficiency, certifiability), for different values of their parameters \QoA and \noiseParamter .
For each dataset, we experiment with three volumes of deleted data points, \emph{small}, \emph{medium}, and \emph{large}, measured as a fraction of the initial training data as shown in Table~\ref{tab:deletion-ratios}.
The deletion volumes correspond different values of accuracy drop, as encountered in Section~\ref{sec:deletion-distribution}.
Specifically, they correspond to a 1\%, 5\% and 10\% drop in \acctest for a fully retrained model when using a \targinfo deletion distribution with class 0 as the deleted class (see Figure~\ref{fig:sampling-distributions}).
Moreover, we group the datasets presented in \Cref{tab:datasets} into three categories,  \emph{low}, \emph{moderate} and \emph{high} based on their dimensionality.
\covtype and \higgs are \emph{low} dimensional datasets, \bmnist and \mnist are \emph{moderate} dimensional datasets and \cifar and \covtype are \emph{high} dimensional datasets.

\begin{submission}
    For some experiments, we'll be reporting partial results due to page-limit constraints.
    In such cases, the results will be available in the extended version of the paper \cite{mahadevan2021certifiable}.
\end{submission}

\begin{table}[htbp]
    \caption{Volume of deleted data, as fraction of initial data volume,  corresponding to different drops in \acctest for fully retrained model and \targinfo deletion (as per Section~\ref{sec:deletion-distribution}, Figure~\ref{fig:sampling-distributions}).}
    \label{tab:deletion-ratios}
    \begin{small}
    \begin{center}
    \begin{tabular}{l lr lr lr}
    \toprule
    \multirow{3}{*}{Dataset} &
    \multicolumn{2}{c}{Small} &
    \multicolumn{2}{c}{Medium} &
    \multicolumn{2}{c}{Large} \\
     &
    \multicolumn{2}{c}{$1$\% Drop} &
    \multicolumn{2}{c}{$5$\% Drop} &
    \multicolumn{2}{c}{$10$\% Drop} \\
    \cmidrule(lr){2-3} \cmidrule(lr){4-5} \cmidrule(lr){6-7}
    & fraction & $|\Dm|$ & fraction & $|\Dm|$ & fraction & $|\Dm|$ \\
    \midrule
    $\bmnist$ & 0.2 & 2396 &0.3 &3594 &0.375 &4493 \\
    $\mnist$ & 0.01 & 600&0.05 &3000 &0.075 &6000 \\
    $\covtype$ & 0.05 &26145 &0.10 &52291 &0.15 &78436 \\
    $\higgs$ & 0.01 & 99000&0.05 &495000 &0.10 & 990000\\
    $\cifar$ & 0.05 &500 &0.125 &1250 &0.2 &2000 \\
    $\eps$ & 0.1 &4000&0.2 & 8000&0.25 &10000 \\
    \bottomrule
    \end{tabular}
    \end{center}
    \end{small}
\end{table}

\input{efficiency_certifiability.tex}
\input{efficiency_effectiveness.tex}
\input{effectiveness_certifiability.tex}

%% file: efficiency_certifiability.tex
\begin{figure*}
    \centering
    \begin{subfigure}{\textwidth}
        \centering
        \includegraphics[width=0.7\linewidth]{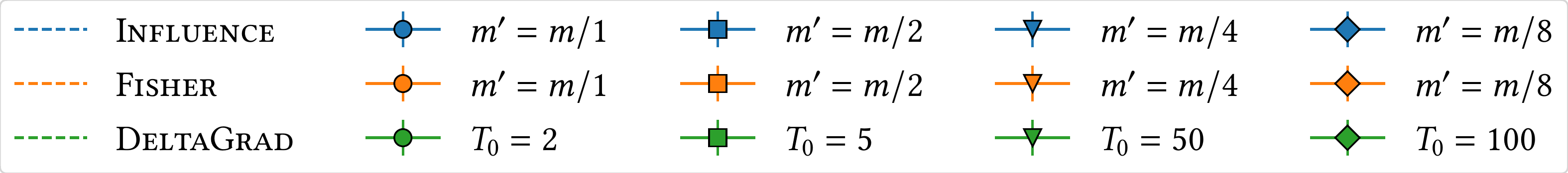}
    \end{subfigure}

    \begin{subfigure}{\textwidth}
        \includegraphics[width=\linewidth]{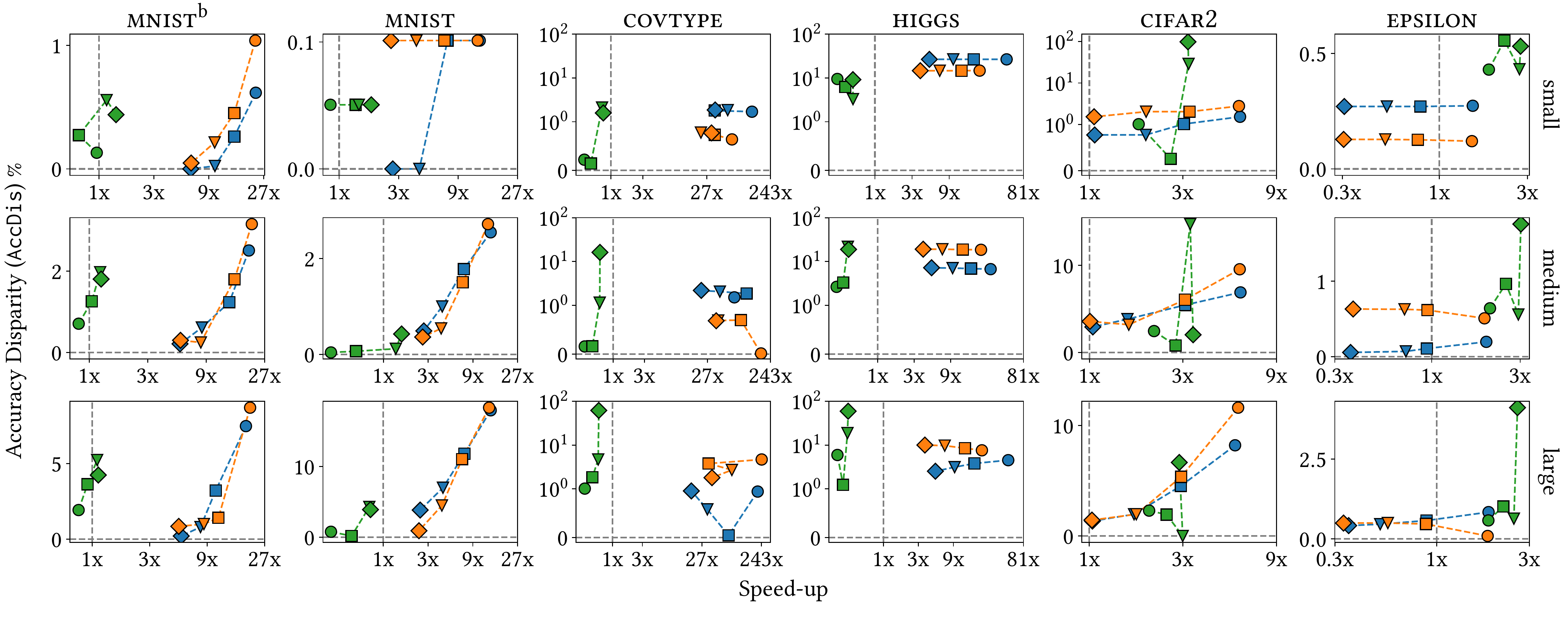}
        \caption{\noiseParamter=0}
        \label{fig:cert-effic-sigma-0}
    \end{subfigure}
    
    \begin{subfigure}{\textwidth}
        \includegraphics[width=\linewidth]{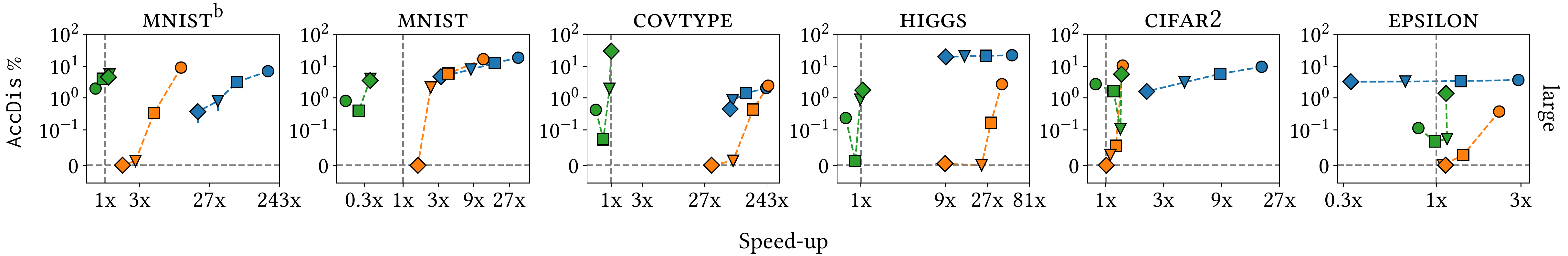}
        \caption{\noiseParamter=1}
        \label{fig:cert-effic-sigma-1}
    \end{subfigure}
    \caption{Efficiency-Certifiability trade-offs for (a) $\noiseParamter=0$ at all volumes and (b) $\noiseParamter=1$ at the largest volume of deletion as efficiency parameter is varied. The y-axis reports certifiability (\accDis) and the x-axis reports the efficiency (speed-up).}
    \label{fig:certifiability-efficiency}
\end{figure*}

\begin{figure*}
    \centering
    \includegraphics[width=\textwidth]{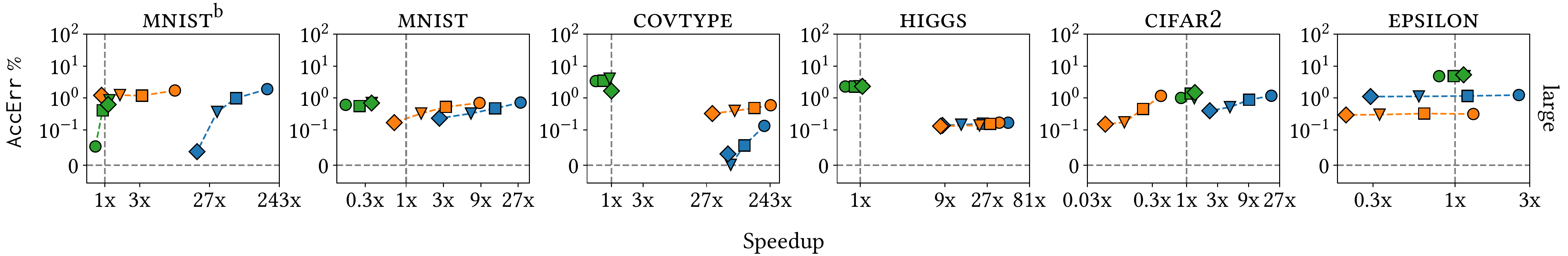}
    \caption{Efficiency-Effectiveness trade-off at $\noiseParamter=1$ for the largest volume of deletion as \QoA is varied. The y-axis reports effectiveness (\accDrop) and the x-axis reports efficiency (speed-up). The legend is same as in \Cref{fig:certifiability-efficiency}. }
    \label{fig:effectiveness-efficiency-sigma-1}
\end{figure*}

\subsection{Efficiency and Certifiability Trade-Off}
\label{subsec:efficiency-certifiability}
In this experiment, we evaluate the trade-off between certifiability and efficiency of each unlearning method when the noise parameter \noiseParamter is kept constant and the efficiency parameter \QoA is varied.
The efficiency parameter for \fisher and \infl is the size of the unlearning mini-batch $\nremovals^{\prime}$ and its values are
\begin{align*}
    \QoA_{_{\fisher}} , \QoA_{_{\infl}} = \nremovals^{\prime} \in \left\{\nremovals,\left\lfloor\frac{\nremovals}{2}\right\rfloor,\left\lfloor\frac{\nremovals}{4}\right\rfloor,\left\lfloor\frac{\nremovals}{8}\right\rfloor\right\},
\end{align*}
where \nremovals is the volume of deleted data.
For the \deltagrad method the efficiency parameter is the periodicity $T_0$ of the unlearning algorithm, and its values are 
\begin{align*}
    \QoA_{_{\deltagrad}} =T_0 \in \{2,5,50,100\}.
\end{align*}
We obtain the updated model \wunlearned and the fully retrained model \worig for each unlearning method as described in \Cref{subsec:fisher,subsec:deltagrad,subsec:guo} at a fixed  value of \noiseParamter.

The results are shown in \Cref{fig:certifiability-efficiency}.
We report the results for two fixed values of the noise parameter and specific volumes of  deletion.
First in \Cref{fig:cert-effic-sigma-0}, we fix $\noiseParamter=0$ and present results for all volumes of deletion corresponding to each row in the figure.
Second, in \Cref{fig:cert-effic-sigma-1}, we fix $\noiseParamter=1$ and present results corresponding only to the largest deletion volume (due to page-limit constraints).
For each plot in the figure, the y-axis reports certifiability (\accDis) and the x-axis reports efficiency (speed-up).
Different unlearning methods and values of \QoA are indicated with different colors and markers respectively in the legend.
\begin{fullpaper}
For extensive results covering different values of \noiseParamter and volumes of deletion, please refer to \Cref{app:certifiability-efficiency} and \Cref{tab:efficiency-extended-links}
\end{fullpaper} 
%
% \begin{submission}
% \end{submission}
%

% \para{Noise parameter $\noiseParamter=0$.} 
% The results in \Cref{fig:cert-effic-sigma-0} display how \QoA trades-off efficiency for effectiveness in the absence of any noise injection.
%

We observe three main trends from \Cref{fig:certifiability-efficiency}.
First, we observe the general trade-off between efficiency and certifiability: higher efficiency (i.e., higher speedup) is typically associated with lower certifiability (i.e., higher \accDis) in the plots.
Some discontinuity in the plotlines, especially for \deltagrad, is largely due to the convergence criteria, particularly since \deltagrad employs \sgd not only for training but also for unlearning.

Second, the \infl and \fisher methods have a roughly similar trend for each dataset.
For the low dimensional datasets they provide large speed-ups of nearly 200x and 50x for each dataset respectively when performing bulk removals (i.e., $\nremovals^{\prime}=\nremovals$), while \deltagrad provides speed-up $<1\text{x}$, i.e., requiring more time than the fully retrained model.
This is because the cost of computing the inverse Hessian matrix (see \cref{eqn:fisher-noise,eqn:influence-hessian}) for \infl and \fisher, is much lower when dimensionality is low, compared to the cost of approximating a large number of \sgd iterations for the \deltagrad method.
Conversely, for the high dimensional datasets, \infl and \fisher provide a smaller speed-up, even when bulk removals are performed (5x and 1.8x respectively for each dataset); and when \QoA is decreased to $\nremovals^{\prime}=\lfloor\nremovals/8\rfloor$, the efficiency is further reduced (1.03x and 0.35x).
Whereas \deltagrad at $T_0=50$ provides comparable and better speed-ups of 2.9x and 2.5x respectively, with similar values of \accDis as compared to the other methods.

Third, as the volume of deletions increases (see the rows of \Cref{fig:cert-effic-sigma-0}), the range of \accDis increases as well.
This is because as we delete more data points, the updated model \wunlearned diverges from the fully retained model \worig due to the approximations in the unlearning algorithm of each method, leading to higher disparity.
This is clearly seen at the largest values of the efficiency parameter ($\nremovals^{\prime}=\nremovals$ and $T_0=100$), where the unlearning algorithm sacrifices the most certifiability, indicated by the largest disparity in accuracy.

Lastly, the trends for \deltagrad are similar to the previous results at $\noiseParamter=0$.
However, the efficiency for the high dimensional datasets is lower, due to the computational cost of noise injection (see \cref{eqn:deltagradUnlearning}).
Furthermore, the certifiability has slightly improved for the low dimensional datasets, as indicated by lower values of \accDis, due to the injected noise.
We also note that increasing the efficiency parameter offers only minor improvements in speed-up, at much larger values of accuracy disparity.

%% file: efficiency_effectiveness.tex
\subsection{Efficiency and Effectiveness Trade-Off}
In this experiment, we evaluate how varying the efficiency parameter \QoA trades-off efficiency for effectiveness when the volume of data deleted and \noiseParamter are kept constant.
The range of \QoA for each unlearning method is the same as in \Cref{subsec:efficiency-certifiability}.
\begin{fullpaper}
In \Cref{fig:effectiveness-efficiency-sigma-1}, we discuss the results for $\noiseParamter=1$ and the large volume of deletion for each dataset.
For extensive results covering all different values of \noiseParamter and volumes of deletion, please refer to \Cref{app:certifiability-efficiency} and \Cref{tab:efficiency-extended-links}.
\end{fullpaper}
\begin{submission}
Due to page-limit constraints, we present results for $\noiseParamter=1$ and the largest volume of deletion for each dataset in \Cref{fig:effectiveness-efficiency-sigma-1}.
\end{submission}
In each plot, effectiveness is reported as the test accuracy error \accDrop, and efficiency is reported as the speed-up in running time (metrics defined in \Cref{subsec:metrics}).
We observe the following trends.
First, we observe the general trade-off: higher efficiency (i.e., higher speed-up) is typically associated with lower accuracy error (i.e., \accDrop) for the same method.

Second, we observe that the \infl offers the best efficiency and effectiveness trade-off among all the methods.
Especially, for the high dimensional datasets, the highest efficiency offered is 20x and 2.5x respectively compared to 0.4x and 1.3x of \fisher, at a slightly larger test accuracy error ($1.17\%$ and $1.2\%$ respectively compared to $1.16\%$ and $0.3\%$ of \fisher).
For the low dimensional datasets, \infl and \fisher offer similar efficiency and effectiveness.
Lastly, for the moderate dimensional datasets, the largest efficiency \infl offers is 168x and 29x respectively compared to 9x and 8.5x of \fisher, at a lower test accuracy error.
Furthermore, decreasing the \QoA in the unlearning algorithm leads to lower test accuracy error as seen clearly in \bmnist and \covtype datasets, because the noise is injected only in the training algorithm.

Third, we again see that \deltagrad is mostly stable both in terms of efficiency and effectiveness as seen in \Cref{subsec:efficiency-certifiability}.
However, note that the test accuracy error for all datasets is larger compared to the other methods due to the direct noise injection (see \cref{eqn:deltagradUnlearning}) and hence offers a lower effectiveness even at $\noiseParamter=1$.

\begin{fullpaper}
Lastly, \fisher offers much lower efficiency for the high dimensional datasets compared to \Cref{subsec:efficiency-certifiability}, where the speed-up is $\leq$1x for nearly all choices of the efficiency parameter.

This is due to the reduced computational effort required to obtain the fully retrained model \worig as no noise injection is done.
Therefore, it takes longer to inject noise using the \fisher unlearning algorithm and obtain an updated model compared to obtaining the fully retrained model, especially when dimensionality is high.
Also, we see that due to the amount of noise injected in the unlearning algorithm, reducing the efficiency parameter only slightly reduces the test accuracy error of the method.
\end{fullpaper}

%% file: effectiveness_certifiability.tex
\begin{figure*} 
    \centering
    \includegraphics[width=\linewidth]{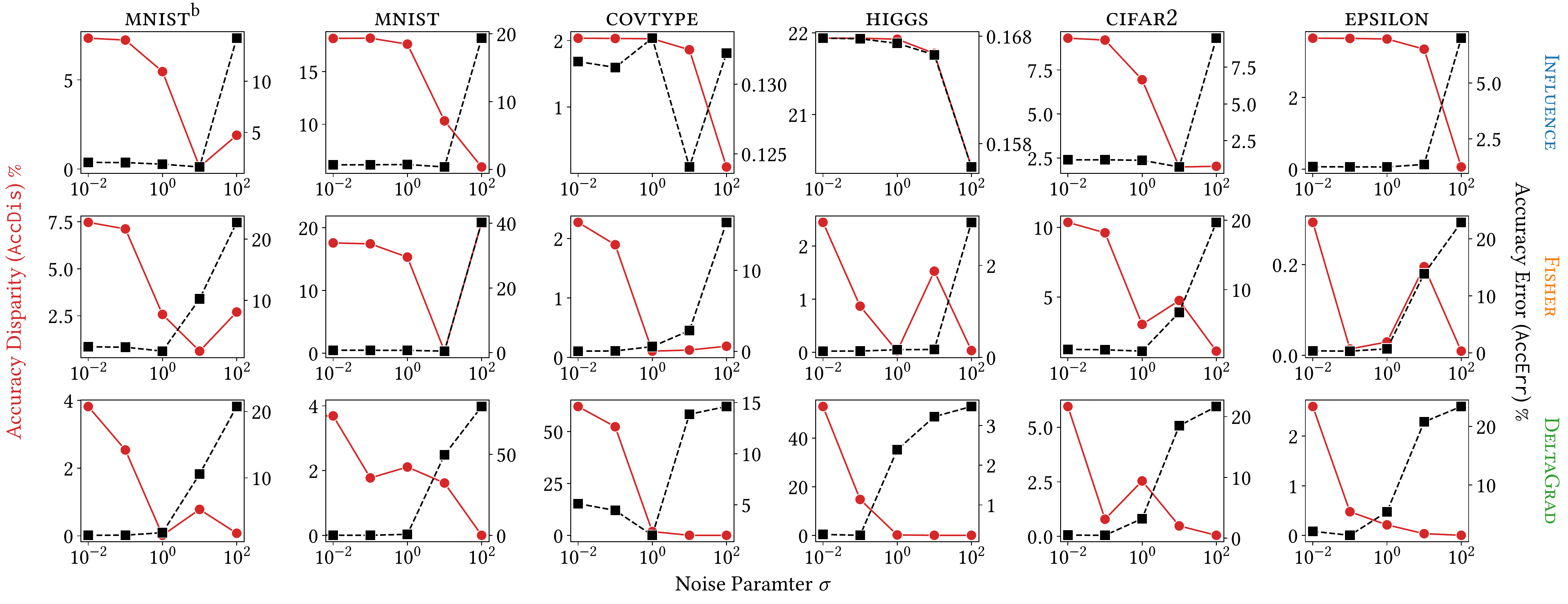}
    \caption{Effectiveness-Certifiability trade-offs for largest volume of deletion. Each row corresponds to an unlearning method. Efficiency parameter is fixed at $\nremovals^{\prime}=m$ for \infl and \fisher and $T_0=100$ for \deltagrad. Left y-axis reports certifiability (\accDis), right y-axis reports effectiveness (\accDrop) and x-axis varies noise parameter \noiseParamter. Lower is better for both y-axes.}
    \label{fig:certifiability-effectiveness}
\end{figure*}

\subsection{Effectiveness and Certifiability Trade-Off}
\label{subsec:certifiability-effectiveness}   
\begin{fullpaper}
    \begin{figure*}
        \centering
        \includegraphics[width=\textwidth]{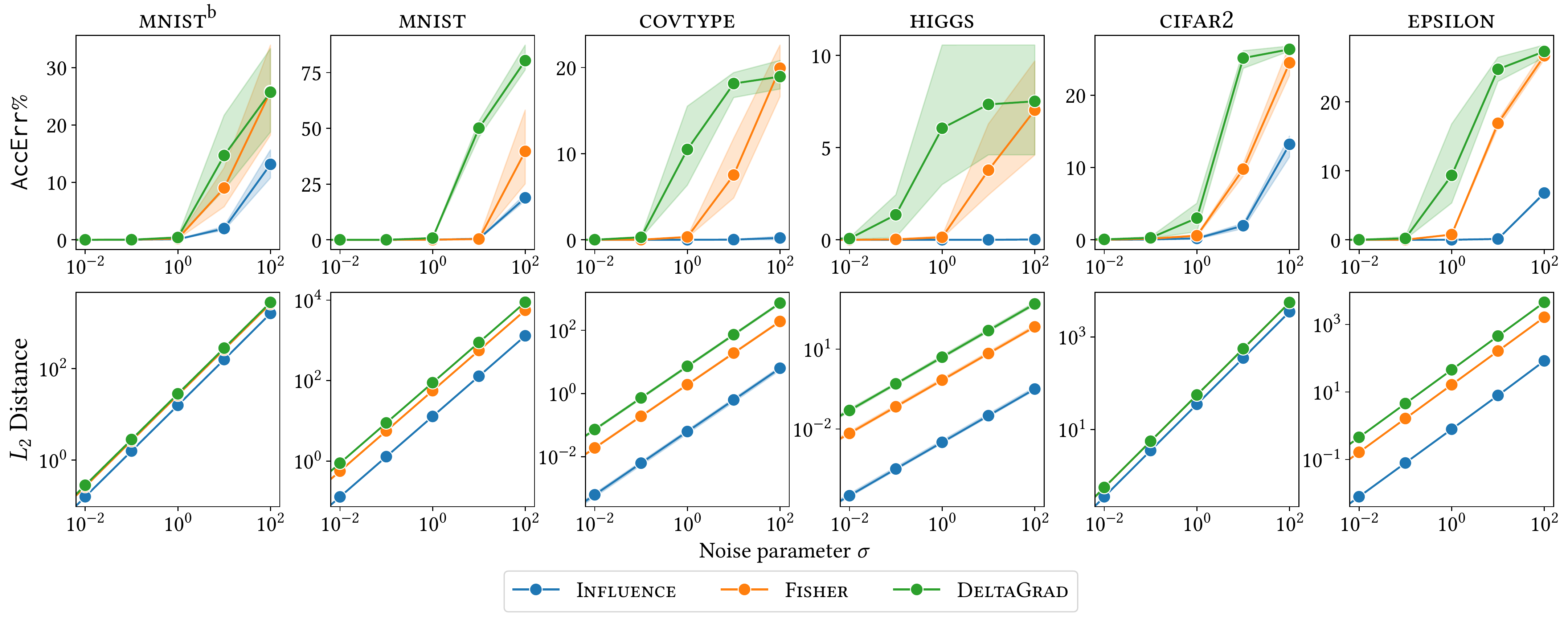}
        \caption{Effect of noise parameter \noiseParamter on the updated model \wunlearned when no data points are deleted. The first row reports the effectiveness (\accDrop) and second row reports the $L_2$ distance between \wunlearned and the optimal model}
        \label{fig:noise-parameter}
    \end{figure*}
    
    In this section, we first study the effect of the noise parameter \noiseParamter on the effectiveness of a fully trained \ml model \worig.
    Next, we evaluate how varying the noise parameter \noiseParamter trades-off effectiveness for certifiability for each unlearning method when the volume of data deleted and the \QoA parameter are kept constant.
    
    \spara{Effect of \noiseParamter on \worig}
    To isolate the effect of data deletion and for simplicity, we select a deletion fraction of 0, i.e., no data is deleted.
    We vary \noiseParamter from $10^{-2}$ till $10^2$ and obtain a fully trained model \worig using the training algorithm of each unlearning method (see \cref{eqn:fisher-train,eqn:deltagrad-train,eqn:influence-train}) on the initial dataset \Dtrain.
    Then, we compare this model with the optimal model (a fully trained model with $\noiseParamter=0$) on the initial dataset. 
    The results are presented in \Cref{fig:noise-parameter}, where we report in the first row the effectiveness using accuracy error \accDrop and in the second row the $L_2$ distance between the two models.
    As no data is deleted, we cannot report the accuracy disparity, hence the $L_2$ distance acts a proxy for the disparity between the fully trained model and the optimal model.

    In \Cref{fig:noise-parameter}, we notice the difference in the extent of the test accuracy error and the $L_2$ distance as a consequence of the noise injection in each unlearning method at different values of \noiseParamter.
    First, the \infl method has the smallest \accDrop and $L_2$ distance from the optimal model -- a direct consequence of noise injection in its objective function (see \cref{eqn:noisyObjectiveFunc}) and the subsequent optimization by the \sgd algorithm during the training stage.
    Second, the \deltagrad method has the largest \accDrop and $L_2$ distance from the optimal model, even at small values of \noiseParamter, due to random noise directly injected into the model parameters (see \cref{eqn:deltagrad-train}).
    Lastly, the \fisher method sits in-between the other methods, both in terms \accDrop and $L_2$ distance due to the injection of random noise in the direction of the Fisher matrix (see \cref{eqn:fisher-train}).
    This lessens the impact on the model parameters in comparison to the \deltagrad method, however, the impact is still greater compared to the optimized noise injection in the \infl method.

    \spara{Trade-off experiments} 
    Here we only present the results for the largest deletion volume from \Cref{tab:deletion-ratios}.
    For more extensive experiments covering different \QoA parameters and volumes of deletion, refer \Cref{app:certifiability-effectiveness-results,tab:certifiability-effectiveness-extended-links}.
\end{fullpaper}
\begin{submission}
In this experiment, we evaluate how varying the noise parameter \noiseParamter trades-off effectiveness for certifiability when the volume of data deleted and the efficiency parameter \QoA are kept constant.
For each dataset, we experimented with each of the three deletion volumes from \Cref{tab:deletion-ratios}.
However, due to page-limit constraints, here we only present the results for the largest one.
\end{submission}
The efficiency parameter \QoA was set as follows: for \infl and \fisher, we set
\begin{align*}
    \QoA_{_{\infl}}=\QoA_{_{\fisher}}=\nremovals^{\prime} := \nremovals,    
\end{align*}
i.e., the size of the unlearning mini-batch was set to be equal to the volume of deleted data; and for \deltagrad, we set 
\begin{align*}
    \QoA_{_{\deltagrad}} = T_0 := 100,    
\end{align*}
i.e., the periodicity is set to 100 \sgd steps.

For different values of the noise parameter \noiseParamter, we obtain the updated models \wunlearned corresponding to each unlearning method as described in \Cref{subsec:fisher,subsec:deltagrad,subsec:guo}.
For baselines, first we obtain the fully retrained model \worig at the same \noiseParamter to measure certifiability and a second fully retrained model \worig at $\noiseParamter=0$ to measure effectiveness, as per Section~\ref{subsec:metrics}.
The results are shown in \Cref{fig:certifiability-effectiveness}.
For each plot in the figure, the left y-axis reports the certifiability (\accDis), the  right y-axis reports effectiveness (\accDrop). as the \noiseParamter is varied from $10^{-2}$ till ${10^{2}}$ for the different unlearning methods.

First, notice that we observe the trade-off between effectiveness and certifiability: higher effectiveness (lower \accDrop) is typically associated with lower certifiability (higher \accDis).

Another clear observation is that for \infl method, the test accuracy error \accDrop remains low for most of the range of \noiseParamter, and increases only at higher values of \noiseParamter ($ \geq 10^1$).
Moreover, we see its largest \accDrop is lower than other methods across all datasets.
For example, in the \mnist dataset, the maximum \accDrop (at $\noiseParamter=100$) is approximately $19\%$, $40\%$ and $79\%$ and for \infl, \fisher and \deltagrad respectively.
At the same time, however, improved certifiability (i.e., decreased \accDis) is achieved for high values of $\noiseParamter$.
Therefore, to obtain a good combination of effectiveness and efficiency, one must select higher values of \noiseParamter, based on the dataset.

Moreover, for \fisher near $\noiseParamter=1$, the trade-off between \accDrop and \accDis is the best amongst all methods across all datasets, \accDrop is 1.7, 0.7, 0.6, 0.16, 1.15 \& 0.7 percent and \accDis is 2.5, 15, 0.1, 0.02, 3 \& 0.03 percent for the datasets respectively as seen in \Cref{fig:certifiability-effectiveness}.
If a good effectiveness-certifiability trade-off is required, then the \fisher method appears to be a very suitable method.

Note that, because \infl and \fisher share the same efficiency parameter \QoA, their experimental results in this section are directly comparable.
However, generally that's not the case with \deltagrad. 
As we saw earlier in this section, \deltagrad is typically quite slower than the other two methods, as evidenced in the achieved speed-ups.
Therefore, for these experiments, we invoked it with the largest value of $\QoA_{_{\deltagrad}} = T_0 := 100$, so that its running time is as small as possible and closer to the running time of the other two methods (and, in fact, comparable for the high-dimensional datasets).
\begin{fullpaper}
As discussed in \citet{wuDeltaGradRapidRetraining2020}, the effectiveness of the \deltagrad method decreases only slightly when the periodicity $T_0$ is set to larger values.
Therefore, for the same computational budget as the other methods, the trade-off between certifiability and effectiveness for \deltagrad will be similar to that shown in \Cref{fig:certifiability-effectiveness}.
    We can observe this in \cref{fig:cert-effec-deltagrad-small,fig:cert-effec-deltagrad-medium,fig:cert-effec-deltagrad-large}, where the rows correspond to different \QoA parameters for \deltagrad.
    We see across the rows for each dataset, the \accDrop is similar, while small differences exist in \accDis due to the randomness of the noise injected.
\end{fullpaper}

%% file: retrain.tex
\section{When to Retrain}
\label{sec:when-to-retrain}
% \note[AM]{Needs a pass to polish to improve readability}
%
When the updated model \wunlearned is obtained after data deletion (see Figure~\ref{fig:pipeline}), a decision is made on whether to employ the model for inference -- or discard it, restart the pipeline, and train a new model on the remaining data.
Specifically, if the test accuracy \accuracy and certifiability disparity \error of the updated model meet certain pre-determined thresholds, then the model is employed for inference, otherwise the pipeline restarts.
Notice, though, that measuring \error directly as per~\cref{eqn:certifiability-metric} would require the fully retrained model \worig, which is not readily available.
In fact, actually computing \worig after every deletion would defeat the purpose of utilizing an unlearning method in the first place.
Moreover, if data deletions occur instantaneously, the current subset of deleted data \Dm will not be available for the computation of \error (\cref{eqn:certifiability-metric}).

Therefore, in practice, \error needs to be estimated.
We propose an estimate based on the empirical observation that, as more data are deleted, accuracy disparity grows proportionally to how much the test accuracy error in comparison to the initial model.
\begin{equation}
    \errorPred=\slope\cdot\accdropinit,
    \label{eqn:disparity-LR-model}
\end{equation}
where \errorPred is the estimate of \accDis for the updated model, \accdropinit is the test accuracy error relative to the initial model \worig,
\begin{equation}
    \accdropinit = \sape(\acctestinit,~\acctest^{u})
    \label{eqn:accuracy-drop-init}
\end{equation}
and \slope is a constant proportion learned from the data \emph{before} the pipeline starts.
Specifically, \slope is calculated in three steps.
First, after training the model \winit on the initial dataset, its test accuracy \acctestinit is calculated.
Second, an updated model \wunlearned and a fully retrained model \worig is obtained, at the highest value of the efficiency parameter \QoA, for a sufficiently large deletion fraction $\theta$, say $\theta=0.5$.
Third, the proportion \slope calculated as 
\begin{align*}
    \slope = \left[ \frac{\accDis}{\accdropinit} \right]_{\text{deletion fraction}=\theta}
\end{align*}

\begin{table}
    \caption{Correlation between \accdropinit  and \error with \targrand deletion distribution for the \fisher unlearning method ($\nremovals^{\prime}=\nremovals$).}
    \label{tab:when-to-retrain}
    \begin{small}
        \begin{center}
            \begin{tabular}{lcc}
            \toprule
            Dataset & Pearson Corr. & Spearman Corr. \\
            \midrule
            $\bmnist$ & 0.963 & 0.612  \\
            $\mnist$ & 0.999 & 1 \\
            $\covtype$ & 0.881 & 0.964  \\
            $\higgs$ & 0.478 &0.892 \\
            $\cifar$ & 0.938& 0.976 \\
            $\eps$ & 0.8787 & 0.891\\
            \bottomrule
            \end{tabular}
        \end{center}
    \end{small}
\end{table}

The empirical measurements in support of this estimate are shown in \Cref{tab:when-to-retrain}, that reports the Pearson and Spearman correlations between \accDis and \accdropinit under a \targrand deletion distribution and varying the deletion fraction from $0.01$ till $0.45$ ($0.095$ for \mnist) utilizing the \fisher unlearning method ($\nremovals^{\prime}=\nremovals$).
From \Cref{tab:when-to-retrain}, we see that apart from the \higgs dataset, there is a large and positive correlation between \accdropinit and \error, which supports the linear estimate of~\cref{eqn:disparity-LR-model}.
\begin{fullpaper}
    The low correlation values of the \higgs dataset can be attributed to the issue regarding variance that was observed when using the \targrand deletion distribution as discussed in \Cref{sec:deletion-distribution}.
\end{fullpaper}
\begin{submission}
    \newpage
\end{submission}
% \balance
Now, when an updated model \wunlearned is obtained from the unlearning stage, the \accdropinit is measured using the stored \acctestinit, and the certifiability disparity is predicted using \Cref{eqn:disparity-LR-model}.
%
% \balance
If the predicted \errorPred exceeds the given threshold for certifiability, then the pipeline restarts and trains a new initial model from scratch on the remaining data and the steps are repeated.

Admittedly, the estimate of \cref{eqn:disparity-LR-model} is highly simplistic, but it is also easy to compute, and we have found empirically that it provides conservative estimates (i.e., mild overestimation of the true value)
\begin{submission}
     -- related measurements are shown in the extended version of this paper \cite{mahadevan2021certifiable}.
\end{submission}
\begin{fullpaper}
    .
    The mean squared errors (MSE) of the proposed estimator of \accDis for the \fisher unlearning method is shown in \Cref{tab:when-to-retrain-mse}.
    We observe apart from the \higgs and \covtype dataset, the errors are low for the proposed estimator.
    The large error for the \higgs data are expected from the discussion above and the low Pearson correlation we observed in \Cref{tab:when-to-retrain}.
    Similarly, the slightly lower Pearson correlation of the \covtype dataset leads to larger MSE when utilizing the simple linear estimator.
    However, the larger Spearman correlation for the \covtype dataset as seen in \Cref{tab:when-to-retrain}, indicates that a non-linear estimator can better capture the relationship between \accdropinit and \accDis.
\end{fullpaper}
%
% However, due to its simplicity, the predicted disparity \errorPred is an overestimate of the true disparity \error of the updated model.
%
The design of more sophisticated models or algorithms to better estimate the certification disparity is left for future work.

\begin{fullpaper}
    \begin{table}
        \caption{Mean Squared Error (MSE) of the proposed estimator of \accDis for the \fisher unlearning method ($\nremovals^{\prime}=\nremovals$). Proportion $c$ is computed at deletion fraction $\theta =0.45$ using \targrand deletions.}
        \label{tab:when-to-retrain-mse}
        \begin{small}
            \begin{center}
                \begin{tabular}{lr}
                \toprule
                Dataset & MSE \\
                \midrule
                $\bmnist$ & 0.15    \\
                $\mnist$ & 3.249 \\
                $\covtype$ & 427.77  \\
                $\higgs$ & 2445.24 \\
                $\cifar$ & 47.088\\
                $\eps$ & 35.775 \\
                \bottomrule
                \end{tabular}
            \end{center}
        \end{small}
    \end{table}
\end{fullpaper}

%% file: conclusion.tex
\section{Conclusion}
We provided an experimental evaluation of three state-of-the-art machine unlearning methods for linear models.
We found that, for the right parameterization, \fisher and \infl offer a combination of good performance qualities, at significant speed-up compared to full retraining; and that \deltagrad offers stable, albeit not as competitive performance.
% We implement and extend the unlearning methods in a common \ml pipeline and compare them on six real-world datasets.
% %
% We analyze each method's trade-offs between efficiency, effectiveness and certifiability in a variety of setting and identify the effect of the volume and distribution of deleted data on the performance of linear \ml models in general.
% %
Moreover, we proposed an online strategy to evaluate an updated \ml model and determine if it satisfies the requirements of the pipeline and when a full-retraining on the remaining data is required.

Our work falls within a wider research area in which \ml tasks (e.g., machine unlearning, in this paper) are studied not only in terms of predictive performance (effectiveness), but also in terms of system efficiency.
For future work, several possibilities are open ahead. 
One direction is to extend the study to general data updates (i.e., not only data deletions) and other \ml settings (e.g., to update more complex models, such as neural networks).
And another direction is to develop more elaborate mechanisms to determine when a full retraining of the updated models is needed.

%% file: appendix.tex
\section{Experimental Setup}
\label{app:setup}
In this section we discuss the additional details regarding the experiments and the implementation of the common \ml pipeline.

\subsection{Training}
\label{app:training}
Ensuring that the training phase of the common \ml pipeline, especially the optimization of each unlearning method is a difficult task.
As mentioned in \Cref{subsec:pipeline-setup}, \infl and \fisher require \sgd convergence and \deltagrad can only use vanilla \sgd.
The additional constraints come from the \deltagrad method.
\citet{wuDeltaGradRapidRetraining2020} describes that smaller mini-batch size leads to lower approximation and hence lower effectiveness.
However, choosing a full-batch gradient descent update as described in \Cref{eq:sgd-step} to ensure best performance of the \deltagrad method leads to the requirement of a large number of epochs to achieve converge for \infl and \fisher.
This is computationally expensive both in calculation of full-batch gradients for the large datasets such as \eps and \higgs and the number of epochs required in total to reach convergence.
Ideally to reduce the impact of the latter, we would fix a number of epochs and then select a larger learning rate $\eta$ to compensate for the slower average gradient updates.
However, we experimentally found that increasing the learning rate $\eta$ beyond 1 has a significant impact on the performance of \deltagrad.
This is primarily because the error in the approximate \sgd step is amplified as the learning rate is increased beyond the optimal learning rate (which results in an increased number of epochs to achieve the same convergence).
These constraints and limitations led us to fix the learning rate to 1 and choose large enough mini-batches (for \deltagrad performance) while keeping the number of epochs low (for computational effort) using a small validation dataset of the initial training data \Dtrain.
The chosen values of the mini-batch size and the number of epochs for each dataset is described in \Cref{tab:training-parameters}.
\begin{table}[h]
    \caption{Values of training parameters common for all unlearning methods}
    \label{tab:training-parameters}
    \begin{tabular}{ccc}
    \toprule
    Dataset & epochs & mini-batch size \\
    \midrule
    \bmnist& 1000& 1024\\
    \mnist& 200&512\\
    \covtype&200&512\\
    \higgs&20&512\\
    \cifar&500&512\\
    \eps&60&512\\
    \bottomrule
    \end{tabular}
\end{table}

\subsection{\deltagrad unlearning method}
\label{app:deltagrad}
As described in \Cref{subsec:deltagrad}, the primary \QoA parameter chosen for the \deltagrad method was the periodicity $T_0$.
Based on the discussion of the hyper-parameter of the \deltagrad in \citet{wuDeltaGradRapidRetraining2020}, the ideal \QoA parameter would be the training mini-batch size. 
However, this would result in a non-standard training stage in the \ml pipeline for the \deltagrad which in turn prevents any comparison with the other unlearning methods.
Therefore, upon fixing the common training stage, we choose the hyper-parameter $T_0$ that best represents the trade-off between effectiveness and efficiency.
The remaining candidate \QoA parameters are the burn-in period $j_0$ and the size of the history for the L-BFGS algorithm $h$.
Following \citet{wuDeltaGradRapidRetraining2020}, we fix $h=2$ for all datasets and the values of $j_0$ are presented in \Cref{tab:deltagrad-parameter}.

\begin{table}[h]
    \caption{Values of \deltagrad burn-in period parameter $j_0$.}
    \label{tab:deltagrad-parameter}
    \begin{tabular}{cc}
    \toprule
    Dataset & $j_0$ \\
    \midrule
    \bmnist&10\\
    \mnist& 20\\
    \covtype&10\\
    \higgs&500\\
    \cifar&20\\
    \eps&10\\
    \bottomrule
    \end{tabular}
\end{table}

\section{Extended Deletion Distribution Results}

\begin{figure*}
    \centering
    \includegraphics[width=\textwidth]{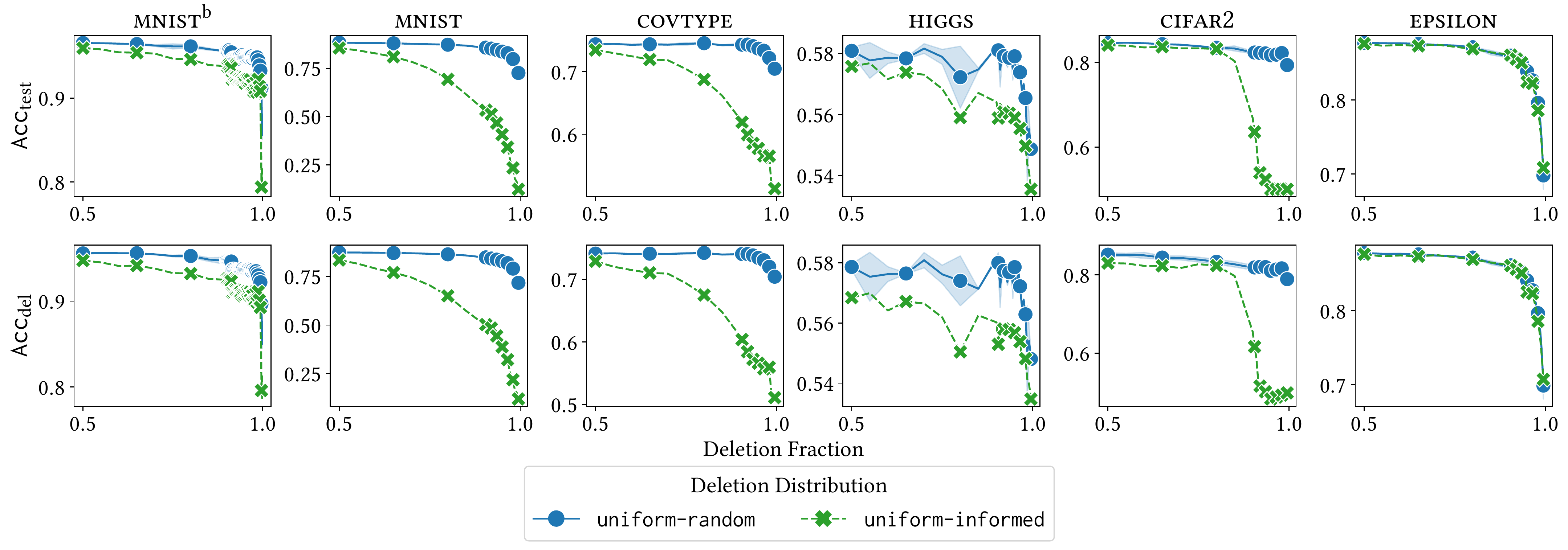}
    \caption{Extended deletion distribution results. Deletion fraction varied from $0.5$ till $0.995$. Only \unirand and \uniinfo deletion distribution results are reported.}
    \label{fig:extended-deletion-distribution}
\end{figure*}

In \Cref{fig:extended-deletion-distribution}, we present the extended results for the \unirand and \uniinfo deletion distribution.
We increase the fraction of data deleted from $0.5$ till $0.995$.
We see that the drop in both \acctest and \accremoved only occur when we delete beyond $90\%$ of the initial training data.
We also clearly see that the drop in both metric is much steeper for the \uniinfo distribution compared to the \unirand distribution.
This indicates that the informed deletions are deleting outliers that are required by the \ml model to effectively classify samples.

\begin{comment}
\begin{table*}
    \caption{For all methods at $\noiseParamter=1$ and largest value of \QoA parameter. Efficiency is measured wrt fully-trained model with noise injection.}
    \begin{tabular}{@{}llllllllll@{}}
    \toprule
\multirow{3}{*}{Method} & \multicolumn{9}{c}{dimensionality}\\
\cmidrule(l){2-10} 
& \multicolumn{3}{c}{low} & \multicolumn{3}{c}{medium} & \multicolumn{3}{c}{high} \\
\cmidrule(lr){2-4}\cmidrule(lr){5-7}\cmidrule(lr){8-10}
& Effic  & Cert  & Effec  & Effic   & Cert   & Effec   & Effic   & Cert  & Effec  \\ \midrule
    \infl &
    \best & \good& \better& 
    \best & \good& \best & 
    \best &\good &\better        \\ 
    \fisher &
    \better & \good & \good &
    & & & 
    & & \\
    \infl               &        &       &        &         &        &         &         &       &       
    \end{tabular}
\end{table*}
\end{comment}

\section{Certifiability-Efficiency Trade-off results}
\begin{figure*}
    \begin{subfigure}{\textwidth}
        \centering
        \includegraphics[width=0.75\textwidth]{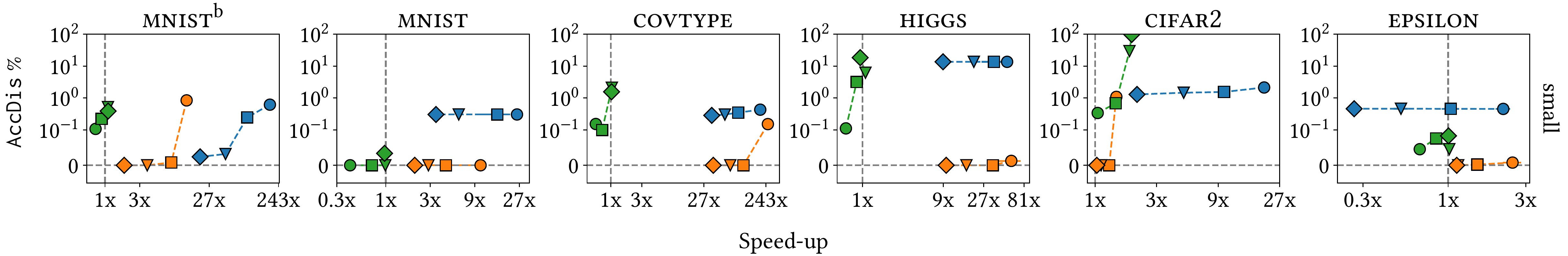}
        \caption{small deletion fraction}
        \label{fig:cert-effic-small-0.01}
    \end{subfigure}
    
    \begin{subfigure}{\textwidth}
        \centering
        \includegraphics[width=0.75\textwidth]{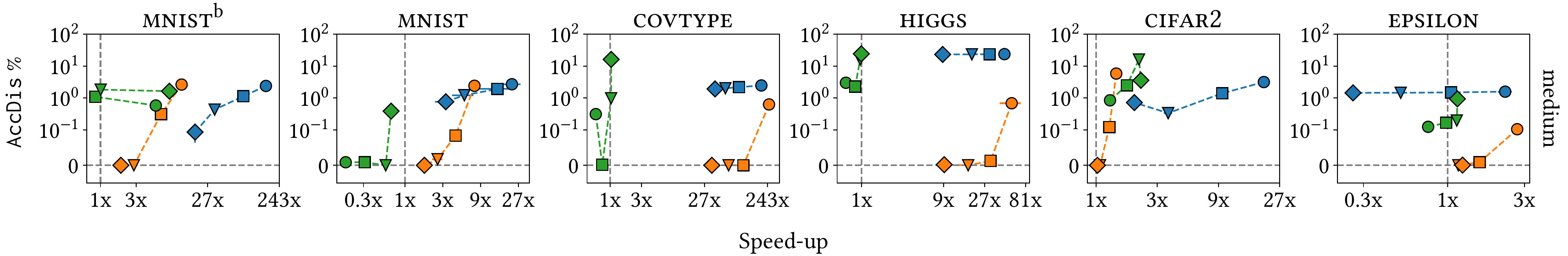}
        \caption{medium deletion fraction}
        \label{fig:cert-effic-medium-0.01}
    \end{subfigure}
    
    \begin{subfigure}{\textwidth}
        \centering
        \includegraphics[width=0.75\textwidth]{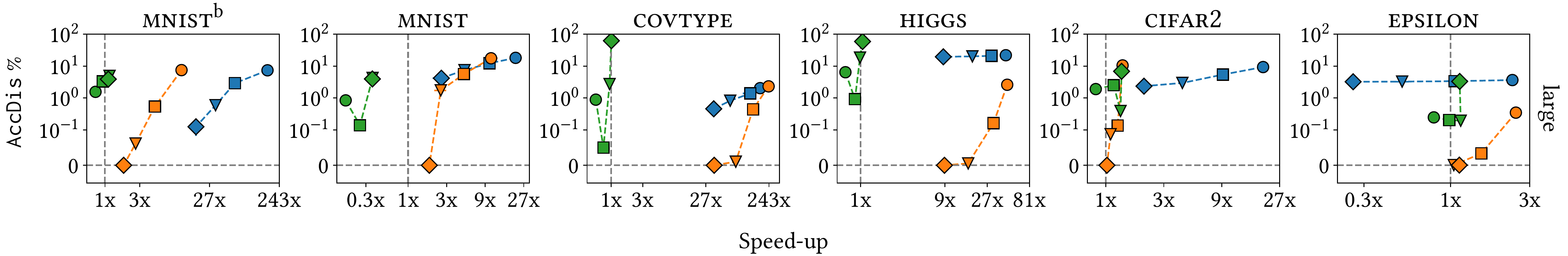}
        \caption{large deletion fraction}
        \label{fig:cert-effic-large-0.01}
    \end{subfigure}
    \caption{Certifiability and efficiency trade-off results for $\noiseParamter=0.01$ at different volumes of deletion : (a) small (b) medium and (c) large. \accDis is reported on the y-axis and the speed-up in running time on the x-axis. Legend is same as in \cref{fig:certifiability-efficiency}}
    \label{fig:cert-effic-0.01}
\end{figure*}

In this section, we present the additional results of the trade-off between certifiability and efficiency as the \QoA parameter is varied at different volumes of deletion and values of \noiseParamter.
In \Cref{tab:certifiability-effectiveness-extended-links}, we provide an interface to easily navigate to the results corresponding to each value of \noiseParamter.
In each figure, there are sub-figures corresponding to different volumes of deletion.
For example, \Cref{fig:cert-effic-0.01} presents results for when $\noiseParamter=0.01$ and sub-figures \cref{fig:cert-effic-small-0.01,fig:cert-effic-medium-0.01,fig:cert-effic-large-0.01} correspond to the \emph{small}, \emph{medium} and \emph{large} deletion volumes described in \Cref{tab:deletion-ratios}.
The legend for the range of the \QoA parameter is the same as that found in \Cref{fig:certifiability-efficiency}.

There are two interesting trends to note from these results.
First, is that for values of $\noiseParamter<1$, we see little to no difference in the trend of the trade-off offered.
This is due to the smaller quantities of injected noise that does not increase certifiability by reducing the accuracy disparity.
Second, is that for smaller volumes of deletion, all of the unlearning methods have lower \accDis and offer higher certifiability, especially at lower values of \QoA.
This is because the unlearning algorithms of each method are better able approximate the fully-retrained model as the number of deleted points is fewer.

\label{app:certifiability-efficiency}
\begin{table}[H]
    \caption{Links to results for the trade-offs for efficiency.}
    \label{tab:efficiency-extended-links}
    \begin{small}
    \begin{tabular}{lccccc}
        \toprule
        \multicolumn{1}{c}{\multirow{2}{*}{Trade-Off}} & \multicolumn{5}{c}{\noiseParamter} \\
        \cmidrule{2-6}
        \multicolumn{1}{c}{}                                & 0.01     & 0.1    & 1    & 10    & 100    \\
        \midrule
        Certifiability-Efficiency &   \cref{fig:cert-effic-0.01}       &    \cref{fig:cert-effic-0.1}      &\cref{fig:cert-effic-1}      &   \cref{fig:cert-effic-10}      &  \cref{fig:cert-effic-100} \\
        Effectiveness-Efficiency &   \cref{fig:effec-effic-0.01}       &    \cref{fig:effec-effic-0.1}      &\cref{fig:effec-effic-1}     &   \cref{fig:effec-effic-10}      &  \cref{fig:effec-effic-100} \\ 
        \bottomrule     
    \end{tabular}
    \end{small}
\end{table}

\begin{figure*}
    \begin{subfigure}{\textwidth}
        \centering
        \includegraphics[width=0.75\textwidth]{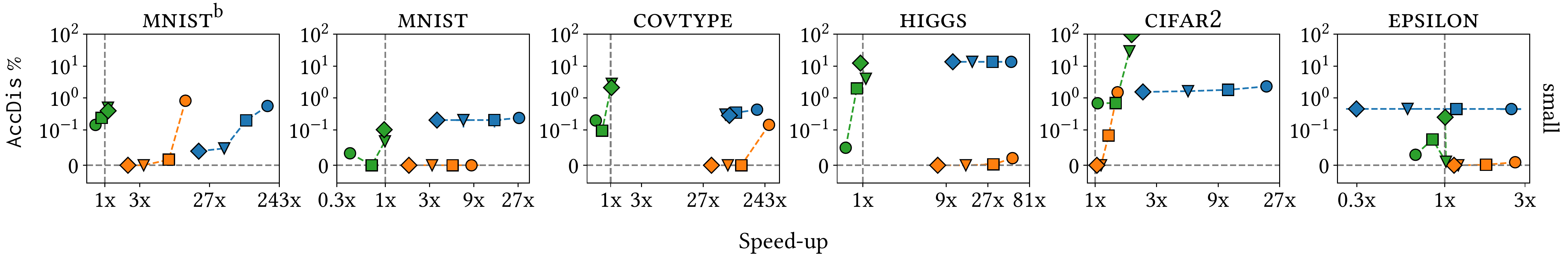}
        \caption{small deletion fraction}
        \label{fig:cert-effic-small-0.1}
    \end{subfigure}
    
    \begin{subfigure}{\textwidth}
        \centering
        \includegraphics[width=0.75\textwidth]{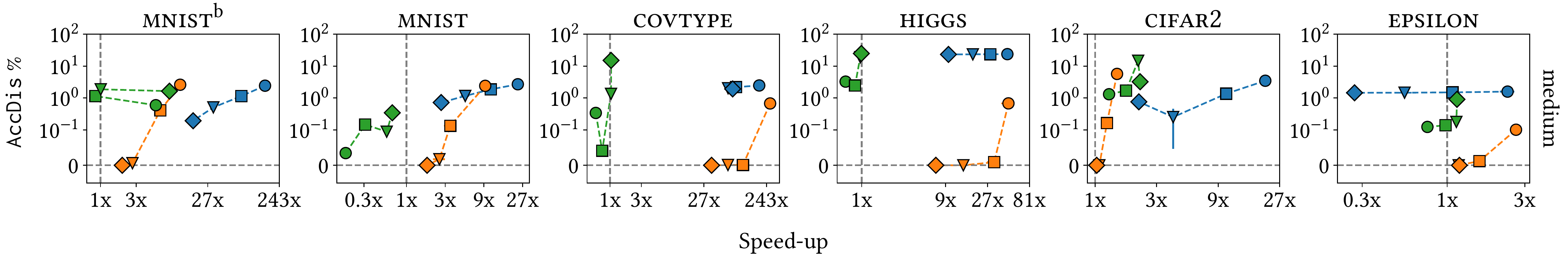}
        \caption{medium deletion fraction}
        \label{fig:cert-effic-medium-0.1}
    \end{subfigure}
    
    \begin{subfigure}{\textwidth}
        \centering
        \includegraphics[width=0.75\textwidth]{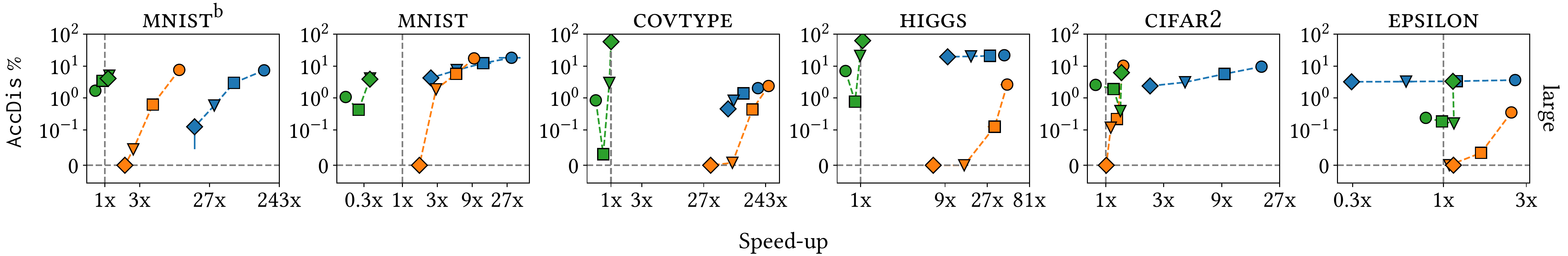}
        \caption{large deletion fraction}
        \label{fig:cert-effic-large-0.1}
    \end{subfigure}
    \caption{Certifiability and efficiency trade-off results for $\noiseParamter=0.1$ at different volumes of deletion : (a) small (b) medium and (c) large. \accDis is reported on the y-axis and the speed-up in running time on the x-axis. Legend is same as in \cref{fig:certifiability-efficiency}}
    \label{fig:cert-effic-0.1}
\end{figure*}

\begin{figure*}
    \begin{subfigure}{\textwidth}
        \centering
        \includegraphics[width=0.75\textwidth]{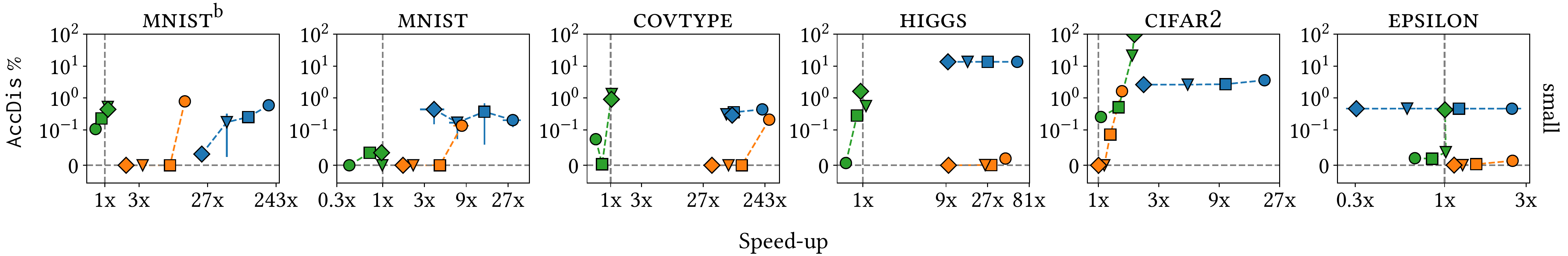}
        \caption{small deletion fraction}
        \label{fig:cert-effic-small-1}
    \end{subfigure}
    
    \begin{subfigure}{\textwidth}
        \centering
        \includegraphics[width=0.75\textwidth]{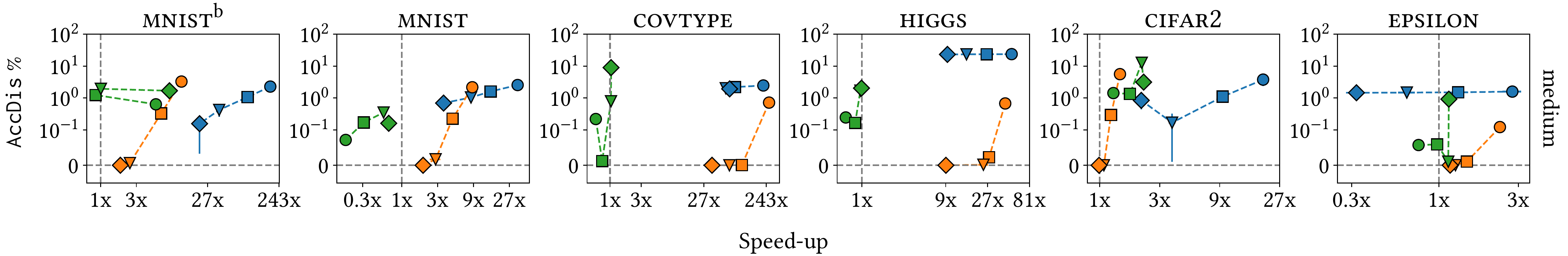}
        \caption{medium deletion fraction}
        \label{fig:cert-effic-medium-1}
    \end{subfigure}
    
    \begin{subfigure}{\textwidth}
        \centering
        \includegraphics[width=0.75\textwidth]{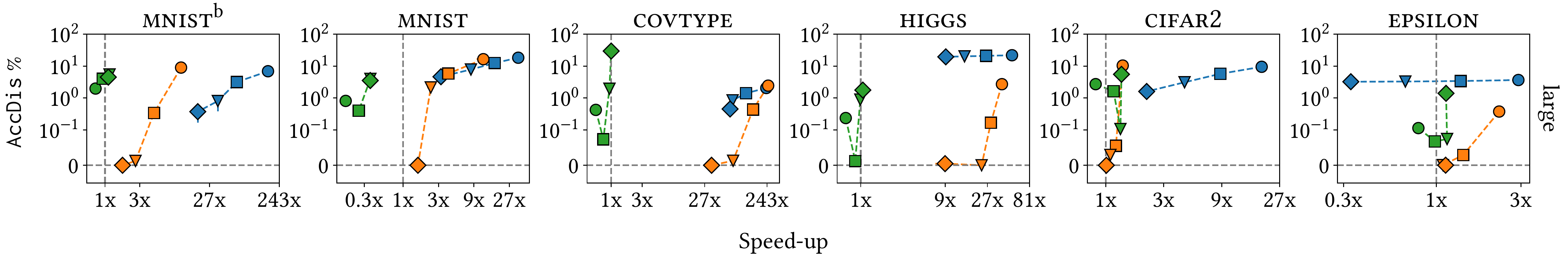}
        \caption{large deletion fraction}
        \label{fig:cert-effic-large-1}
    \end{subfigure}
    \caption{Certifiability and efficiency trade-off results for $\noiseParamter=1$ at different volumes of deletion : (a) small (b) medium and (c) large. \accDis is reported on the y-axis and the speed-up in running time on the x-axis. Legend is same as in \cref{fig:certifiability-efficiency}}
    \label{fig:cert-effic-1}
\end{figure*}

\begin{figure*}
    \begin{subfigure}{\textwidth}
        \centering
        \includegraphics[width=0.75\textwidth]{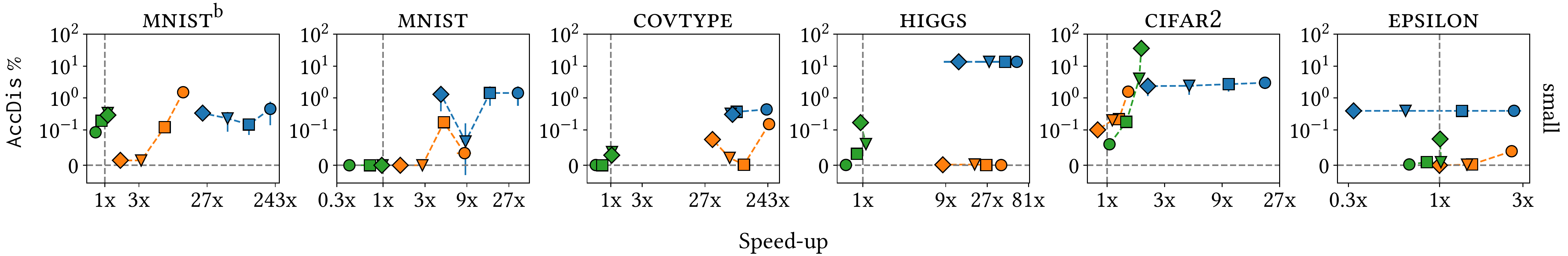}
        \caption{small deletion fraction}
        \label{fig:cert-effic-small-10}
    \end{subfigure}
    
    \begin{subfigure}{\textwidth}
        \centering
        \includegraphics[width=0.75\textwidth]{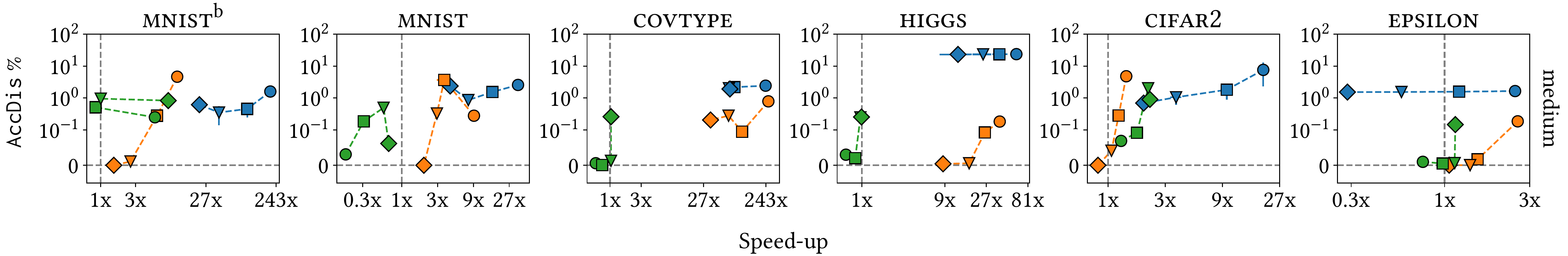}
        \caption{medium deletion fraction}
        \label{fig:cert-effic-medium-10}
    \end{subfigure}
    
    \begin{subfigure}{\textwidth}
        \centering
        \includegraphics[width=0.75\textwidth]{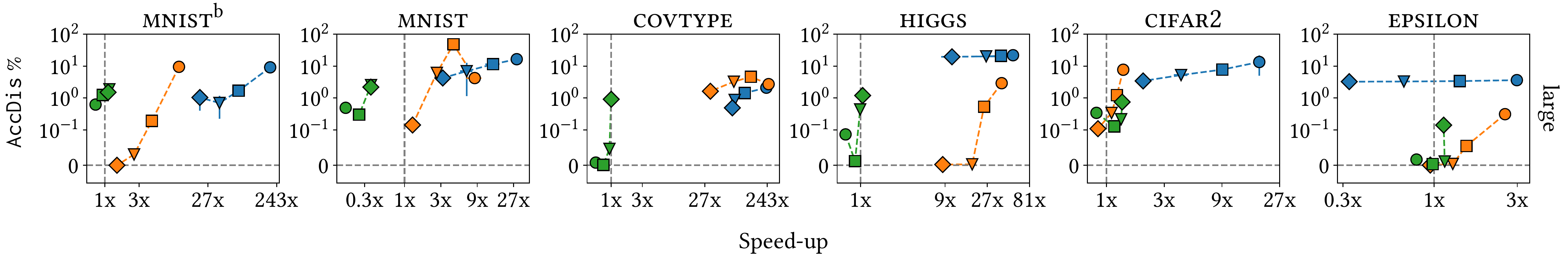}
        \caption{large deletion fraction}
        \label{fig:cert-effic-large-10}
    \end{subfigure}
    \caption{Certifiability and efficiency trade-off results for $\noiseParamter=10$ at different volumes of deletion : (a) small (b) medium and (c) large. \accDis is reported on the y-axis and the speed-up in running time on the x-axis. Legend is same as in \cref{fig:certifiability-efficiency}}
    \label{fig:cert-effic-10}
\end{figure*}

\begin{figure*}
    \begin{subfigure}{\textwidth}
        \centering
        \includegraphics[width=0.75\textwidth]{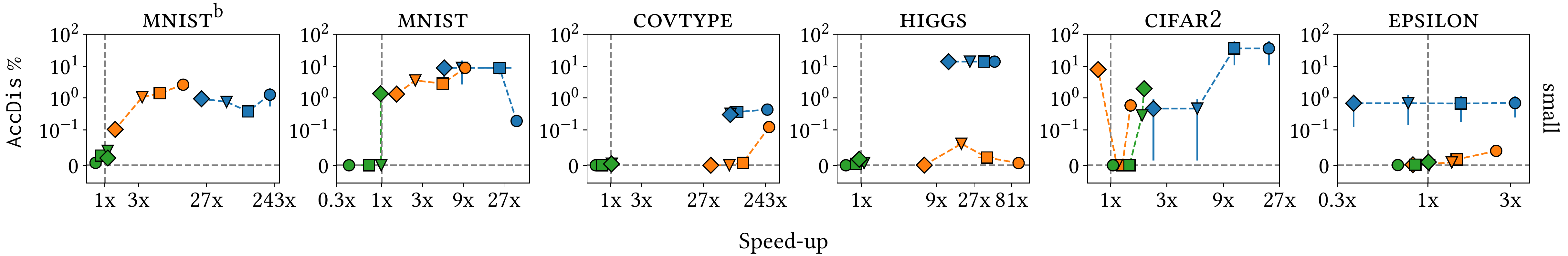}
        \caption{small deletion fraction}
        \label{fig:cert-effic-small-100}
    \end{subfigure}
    
    \begin{subfigure}{\textwidth}
        \centering
        \includegraphics[width=0.75\textwidth]{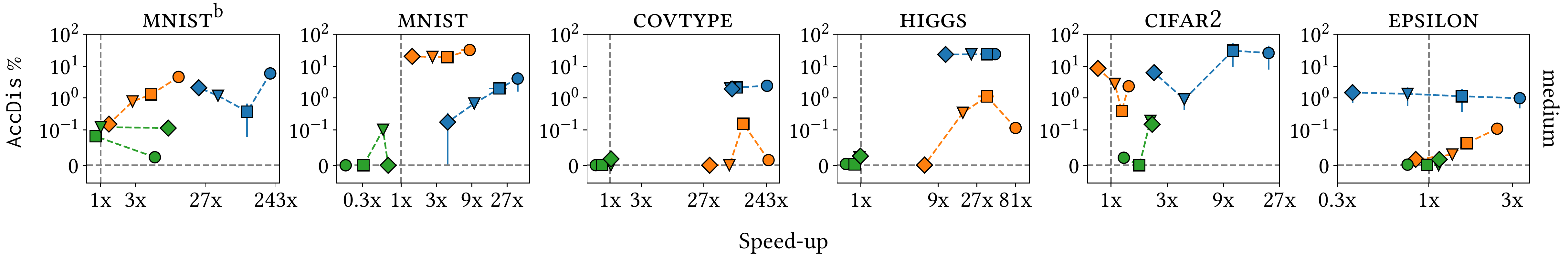}
        \caption{medium deletion fraction}
        \label{fig:cert-effic-medium-100}
    \end{subfigure}
    
    \begin{subfigure}{\textwidth}
        \centering
        \includegraphics[width=0.75\textwidth]{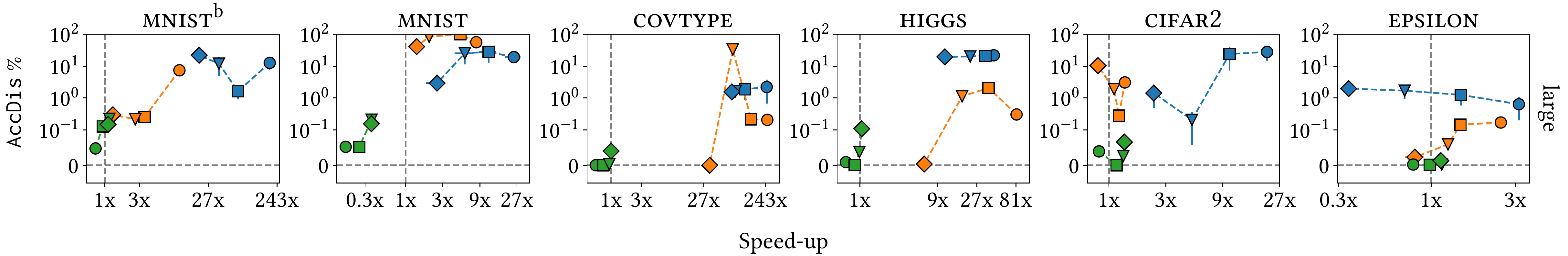}
        \caption{large deletion fraction}
        \label{fig:cert-effic-large-100}
    \end{subfigure}
    \caption{Certifiability and efficiency trade-off results for $\noiseParamter=100$ at different volumes of deletion : (a) small (b) medium and (c) large. \accDis is reported on the y-axis and the speed-up in running time on the x-axis. Legend is same as in \cref{fig:certifiability-efficiency}}
    \label{fig:cert-effic-100}
\end{figure*}

\section{Effectiveness-Efficiency Trade-off Results}

\begin{figure*}
    \begin{subfigure}{\textwidth}
        \centering
        \includegraphics[width=0.75\textwidth]{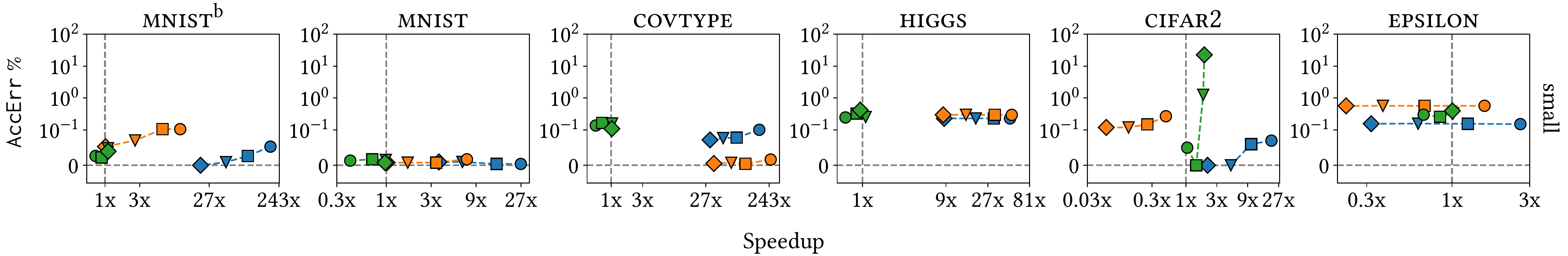}
        \caption{small deletion fraction}
        \label{fig:effec-effic-small-0.01}
    \end{subfigure}
    
    \begin{subfigure}{\textwidth}
        \centering
        \includegraphics[width=0.75\textwidth]{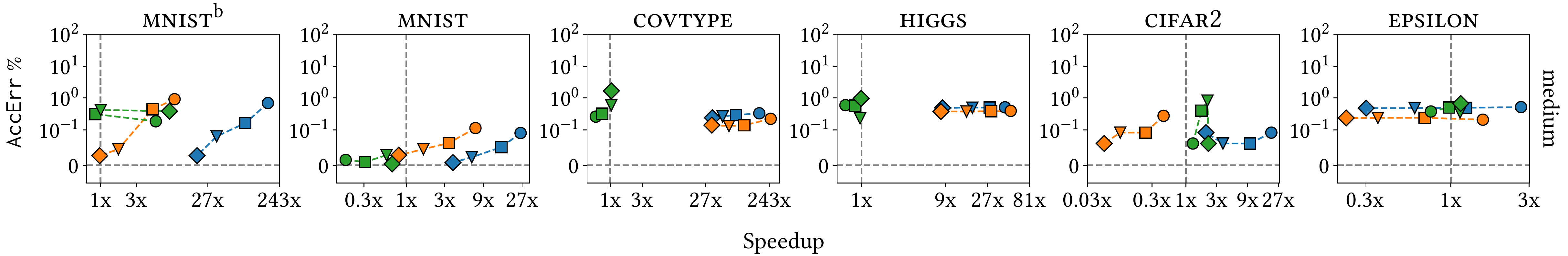}
        \caption{medium deletion fraction}
        \label{fig:effec-effic-medium-0.01}
    \end{subfigure}
    
    \begin{subfigure}{\textwidth}
        \centering
        \includegraphics[width=0.75\textwidth]{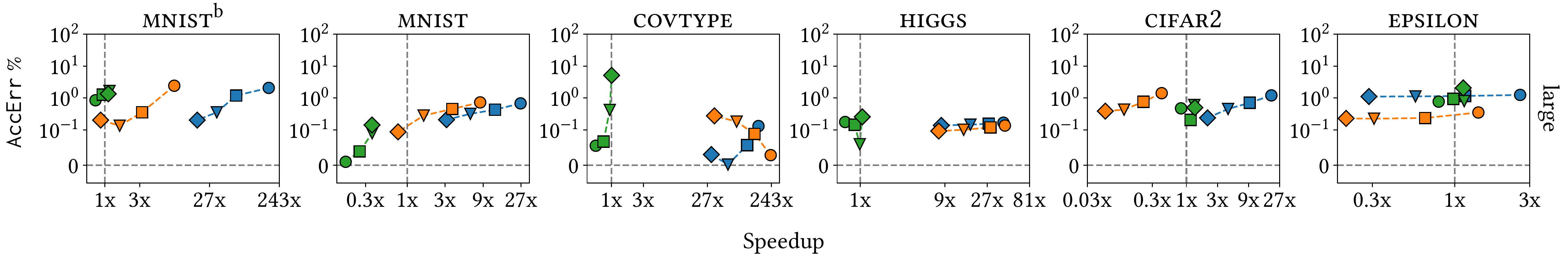}
        \caption{large deletion fraction}
        \label{fig:effec-effic-large-0.01}
    \end{subfigure}
    \caption{Effectiveness and efficiency trade-off results for $\noiseParamter=0.01$ at different volumes of deletion : (a) small (b) medium and (c) large. \accDrop is reported on the y-axis and the speed-up in running time on the x-axis. Legend is same as in \cref{fig:certifiability-efficiency}}
    \label{fig:effec-effic-0.01}
\end{figure*}

In this section, we present the additional results of the trade-off between efficiency and effectiveness as the \QoA parameter is varied for different volumes of deletion and values of \noiseParamter.
In \Cref{tab:certifiability-effectiveness-extended-links}, we provide an interface to easily navigate to the results corresponding to each value of \noiseParamter.
In each figure, there are sub-figures corresponding to different volumes of deletion.
For example, \Cref{fig:effec-effic-0.01} presents results for when $\noiseParamter=0.01$ and sub-figures \cref{fig:effec-effic-small-0.01,fig:effec-effic-medium-0.01,fig:effec-effic-large-0.01} correspond to the \emph{small}, \emph{medium} and \emph{large} deletion volumes described in \Cref{tab:deletion-ratios}.
The legend for the range of the \QoA parameter is the same as that found in \Cref{fig:certifiability-efficiency}.

\begin{figure*}
    \begin{subfigure}{\textwidth}
        \centering
        \includegraphics[width=0.75\textwidth]{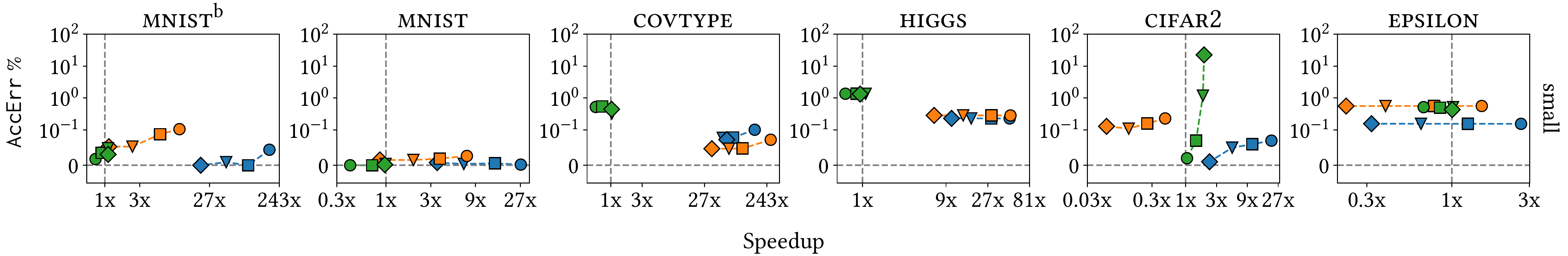}
        \caption{small deletion fraction}
        \label{fig:effec-effic-small-0.1}
    \end{subfigure}
    
    \begin{subfigure}{\textwidth}
        \centering
        \includegraphics[width=0.75\textwidth]{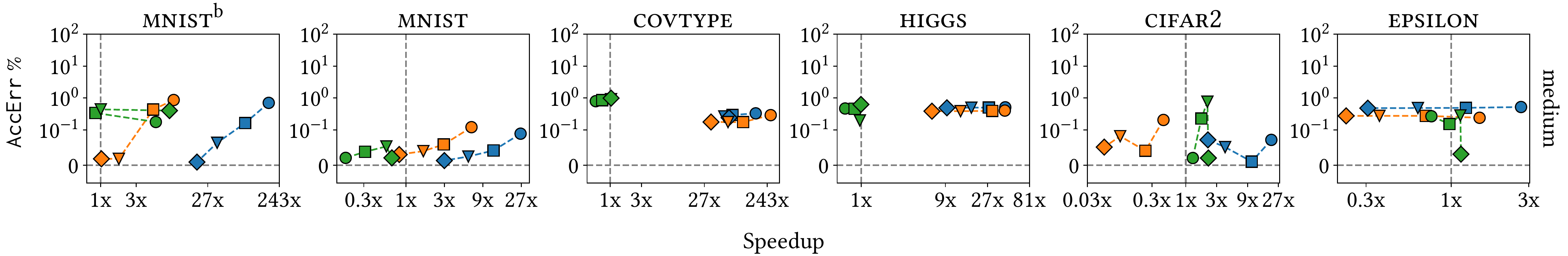}
        \caption{medium deletion fraction}
        \label{fig:effec-effic-medium-0.1}
    \end{subfigure}
    
    \begin{subfigure}{\textwidth}
        \centering
        \includegraphics[width=0.75\textwidth]{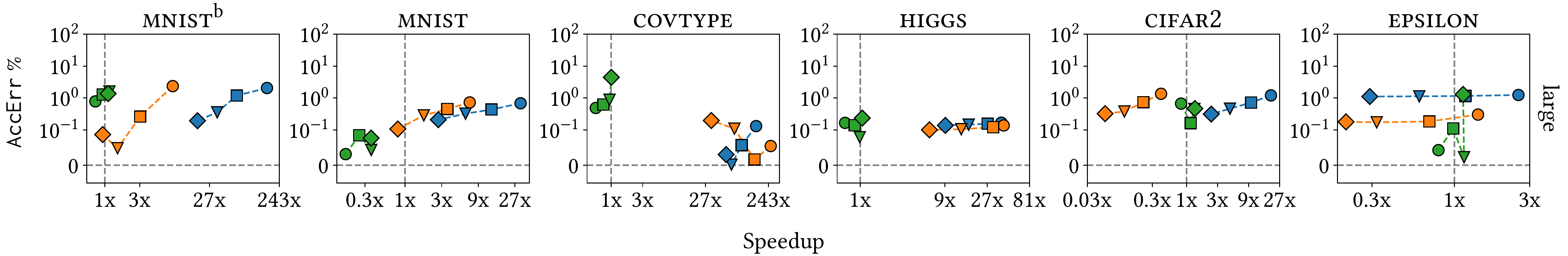}
        \caption{large deletion fraction}
        \label{fig:effec-effic-large-0.1}
    \end{subfigure}
    \caption{Effectiveness and efficiency trade-off results for $\noiseParamter=0.1$ at different volumes of deletion : (a) small (b) medium and (c) large. \accDrop is reported on the y-axis and the speed-up in running time on the x-axis. Legend is same as in \cref{fig:certifiability-efficiency}}
    \label{fig:effec-effic-0.1}
\end{figure*}

\begin{figure*}
    \begin{subfigure}{\textwidth}
        \centering
        \includegraphics[width=0.75\textwidth]{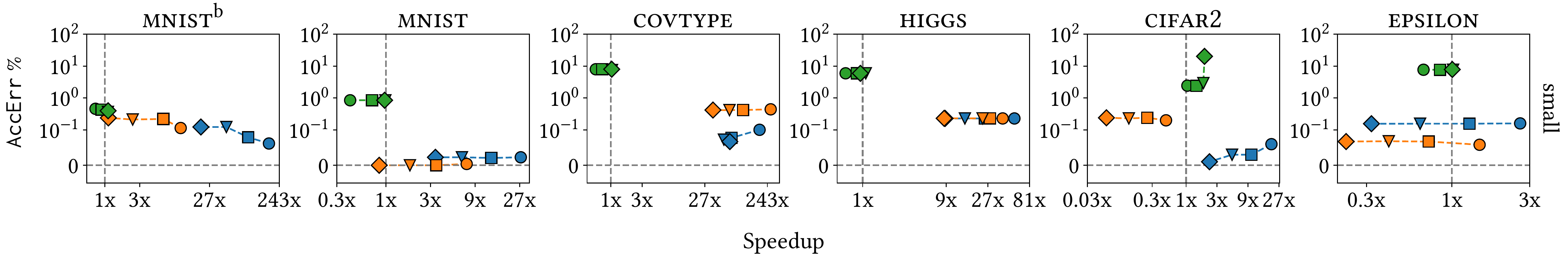}
        \caption{small deletion fraction}
        \label{fig:effec-effic-small-1}
    \end{subfigure}
    
    \begin{subfigure}{\textwidth}
        \centering
        \includegraphics[width=0.75\textwidth]{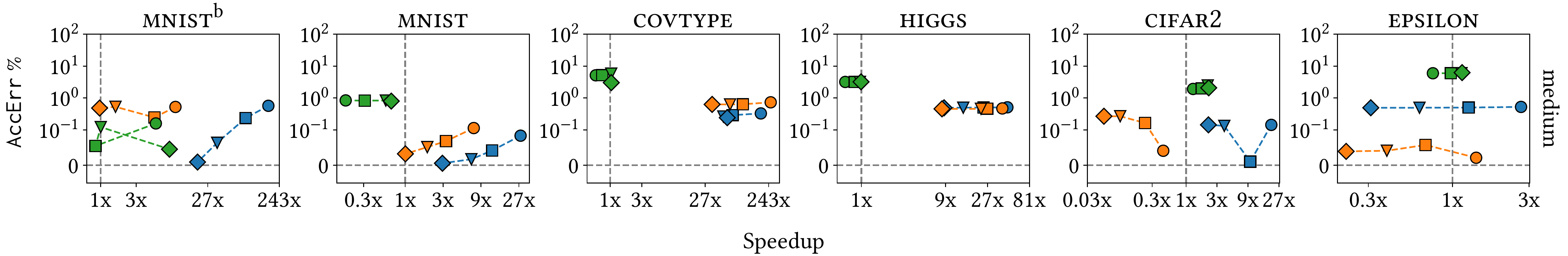}
        \caption{medium deletion fraction}
        \label{fig:effec-effic-medium-1}
    \end{subfigure}
    
    \begin{subfigure}{\textwidth}
        \centering
        \includegraphics[width=0.75\textwidth]{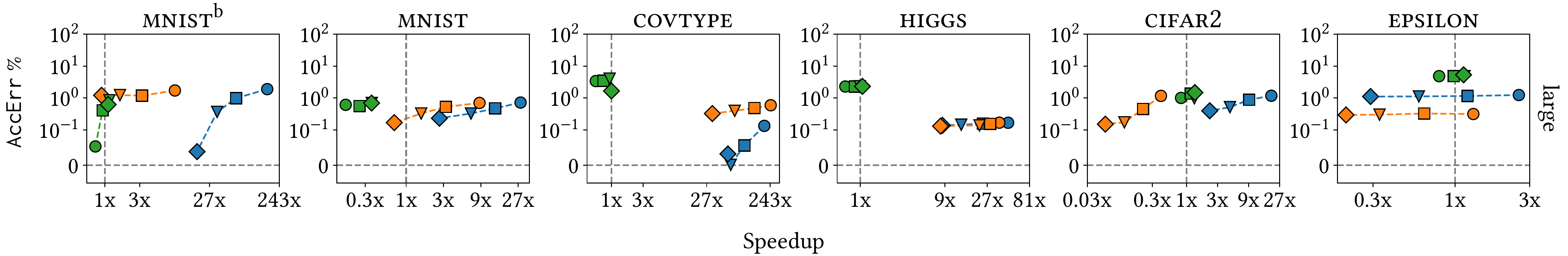}
        \caption{large deletion fraction}
        \label{fig:effec-effic-large-1}
    \end{subfigure}
    \caption{Effectiveness and efficiency trade-off results for $\noiseParamter=1$ at different volumes of deletion : (a) small (b) medium and (c) large. \accDrop is reported on the y-axis and the speed-up in running time on the x-axis. Legend is same as in \cref{fig:certifiability-efficiency}}
    \label{fig:effec-effic-1}
\end{figure*}

\begin{figure*}
    \begin{subfigure}{\textwidth}
        \centering
        \includegraphics[width=0.75\textwidth]{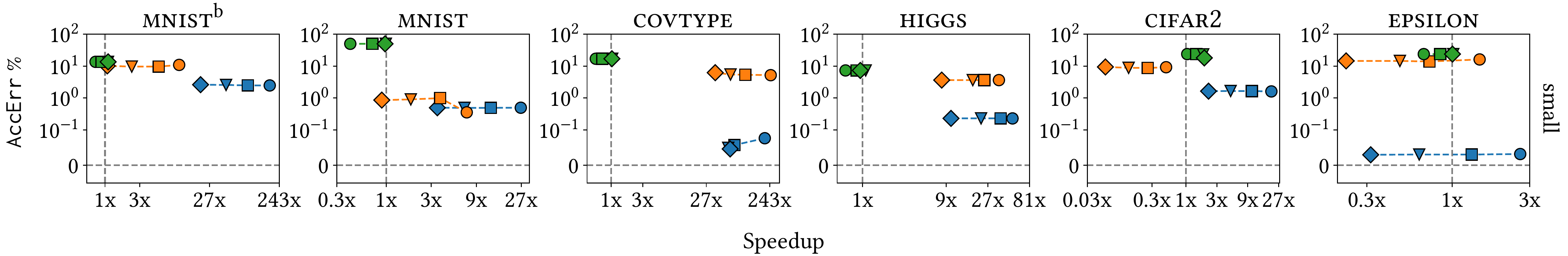}
        \caption{small deletion fraction}
        \label{fig:effec-effic-small-10}
    \end{subfigure}
    
    \begin{subfigure}{\textwidth}
        \centering
        \includegraphics[width=0.75\textwidth]{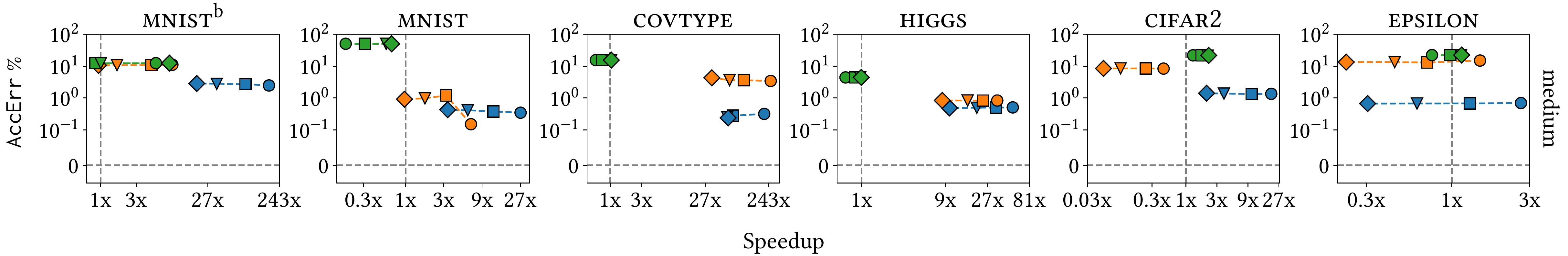}
        \caption{medium deletion fraction}
        \label{fig:effec-effic-medium-10}
    \end{subfigure}
    
    \begin{subfigure}{\textwidth}
        \centering
        \includegraphics[width=0.75\textwidth]{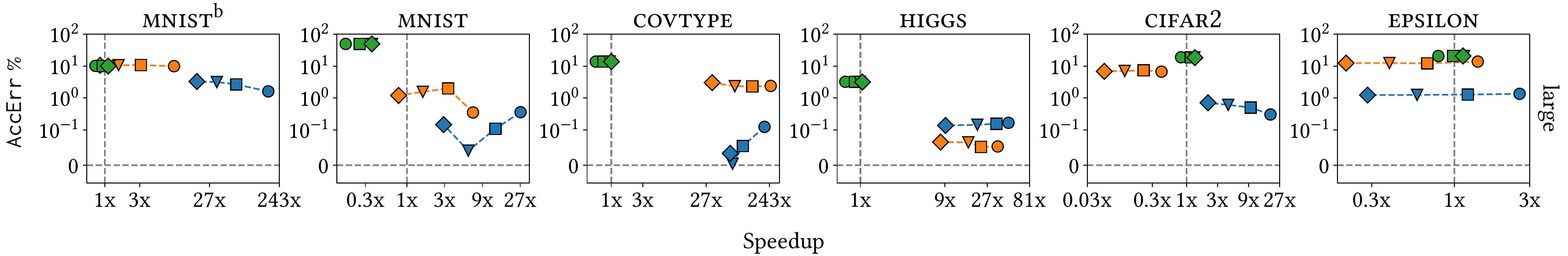}
        \caption{large deletion fraction}
        \label{fig:effec-effic-large-10}
    \end{subfigure}
    \caption{Effectiveness and efficiency trade-off results for $\noiseParamter=10$ at different volumes of deletion : (a) small (b) medium and (c) large. \accDrop is reported on the y-axis and the speed-up in running time on the x-axis. Legend is same as in \cref{fig:certifiability-efficiency}}
    \label{fig:effec-effic-10}
\end{figure*}

\begin{figure*}
    \begin{subfigure}{\textwidth}
        \centering
        \includegraphics[width=0.75\textwidth]{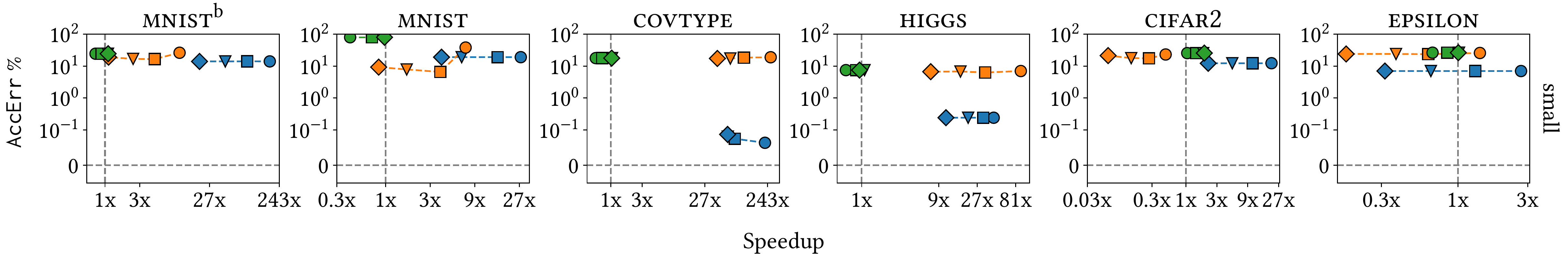}
        \caption{small deletion fraction}
        \label{fig:effec-effic-small-100}
    \end{subfigure}
    
    \begin{subfigure}{\textwidth}
        \centering
        \includegraphics[width=0.75\textwidth]{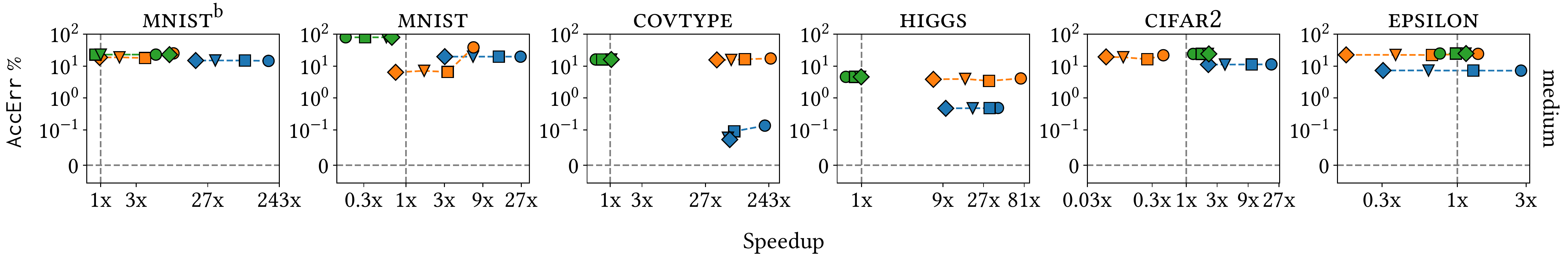}
        \caption{medium deletion fraction}
        \label{fig:effec-effic-medium-100}
    \end{subfigure}
    
    \begin{subfigure}{\textwidth}
        \centering
        \includegraphics[width=0.75\textwidth]{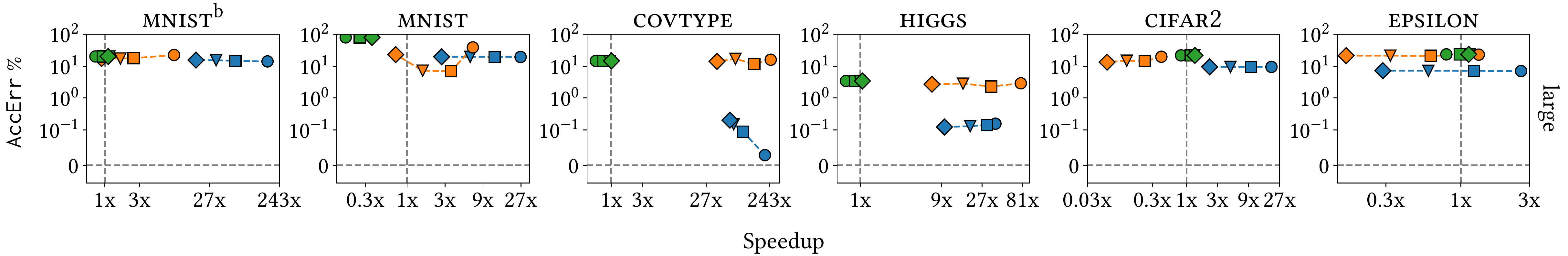}
        \caption{large deletion fraction}
        \label{fig:effec-effic-large-100}
    \end{subfigure}
    \caption{Effectiveness and efficiency trade-off results for $\noiseParamter=100$ at different volumes of deletion : (a) small (b) medium and (c) large. \accDrop is reported on the y-axis and the speed-up in running time on the x-axis. Legend is same as in \cref{fig:certifiability-efficiency}}
    \label{fig:effec-effic-100}
\end{figure*}

\section{Certifiability-Effectiveness Trade-off Results}
\label{app:certifiability-effectiveness-results}
In this section, we present the additional results for the trade-off between certifiability and effectiveness as \noiseParamter is varied for different volumes of deletions and values of the \QoA parameter.
In \Cref{app:cert-effec-small,app:cert-effec-medium,app:cert-effec-large} we present the results corresponding to the \emph{small}, \emph{medium} and \emph{large} deletion volumes described in \Cref{tab:deletion-ratios}.
In each subsection, we present the results for each unlearning method in a figure, where the rows of the figure correspond to different values of the \QoA parameter.
In \Cref{tab:certifiability-effectiveness-extended-links}, we provide an interface to easily navigate to the results corresponding to each unlearning method and volume of deletion.
\begin{table}
    \caption{Links to results of the trade-off between certifiability and effectiveness for different methods and deletion volumes}
    \label{tab:certifiability-effectiveness-extended-links}
    \begin{small}
    \begin{tabular}{cccc}
    \toprule
    \multirow{2}{*}{Method} &
    \multicolumn{3}{c}{Deletion Volume}\\
    \cmidrule{2-4}
     & small & medium & large \\
    \midrule
    \fisher    & \cref{fig:cert-effec-fisher-small}       & \cref{fig:cert-effec-fisher-medium}        & \cref{fig:cert-effec-fisher-large}      \\
    \deltagrad    & \cref{fig:cert-effec-deltagrad-small}       &  \cref{fig:cert-effec-deltagrad-medium}       & \cref{fig:cert-effec-deltagrad-large}     \\ 
    \infl   & \cref{fig:cert-effec-influence-small}      &  \cref{fig:cert-effec-influence-small}      &  \cref{fig:cert-effec-influence-small}    \\
    \bottomrule
    \end{tabular}
\end{small}
\end{table}

\subsection{Small deletion volume results}
\label{app:cert-effec-small}
\begin{figure*}
    \centering
    \includegraphics[width=\linewidth]{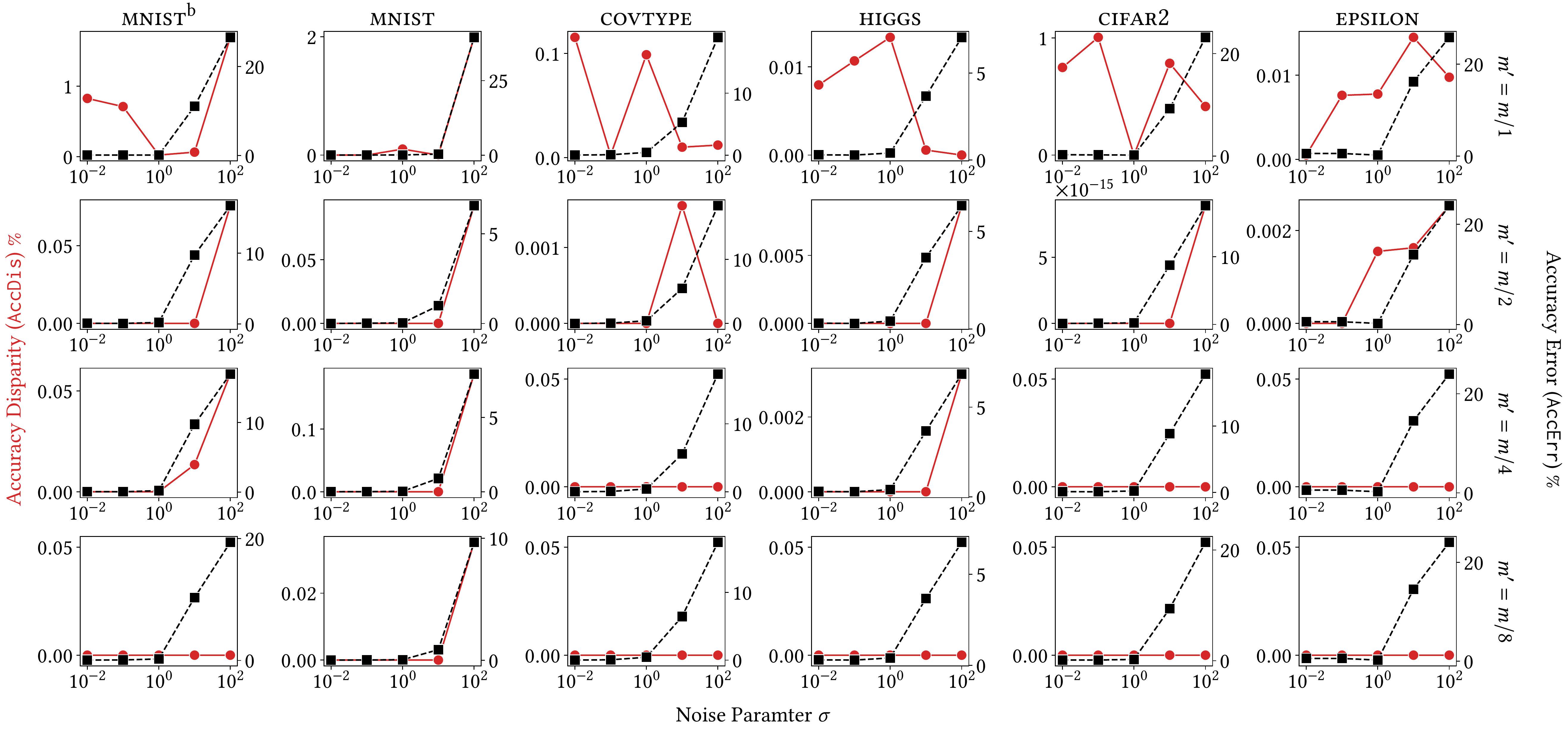}
    \caption{Certifiability-Effectiveness trade-off for \fisher method at small deletion volume. Each row corresponds to a value of the \QoA parameter}
    \label{fig:cert-effec-fisher-small}
\end{figure*}

\begin{figure*}
    \centering
    \includegraphics[width=\linewidth]{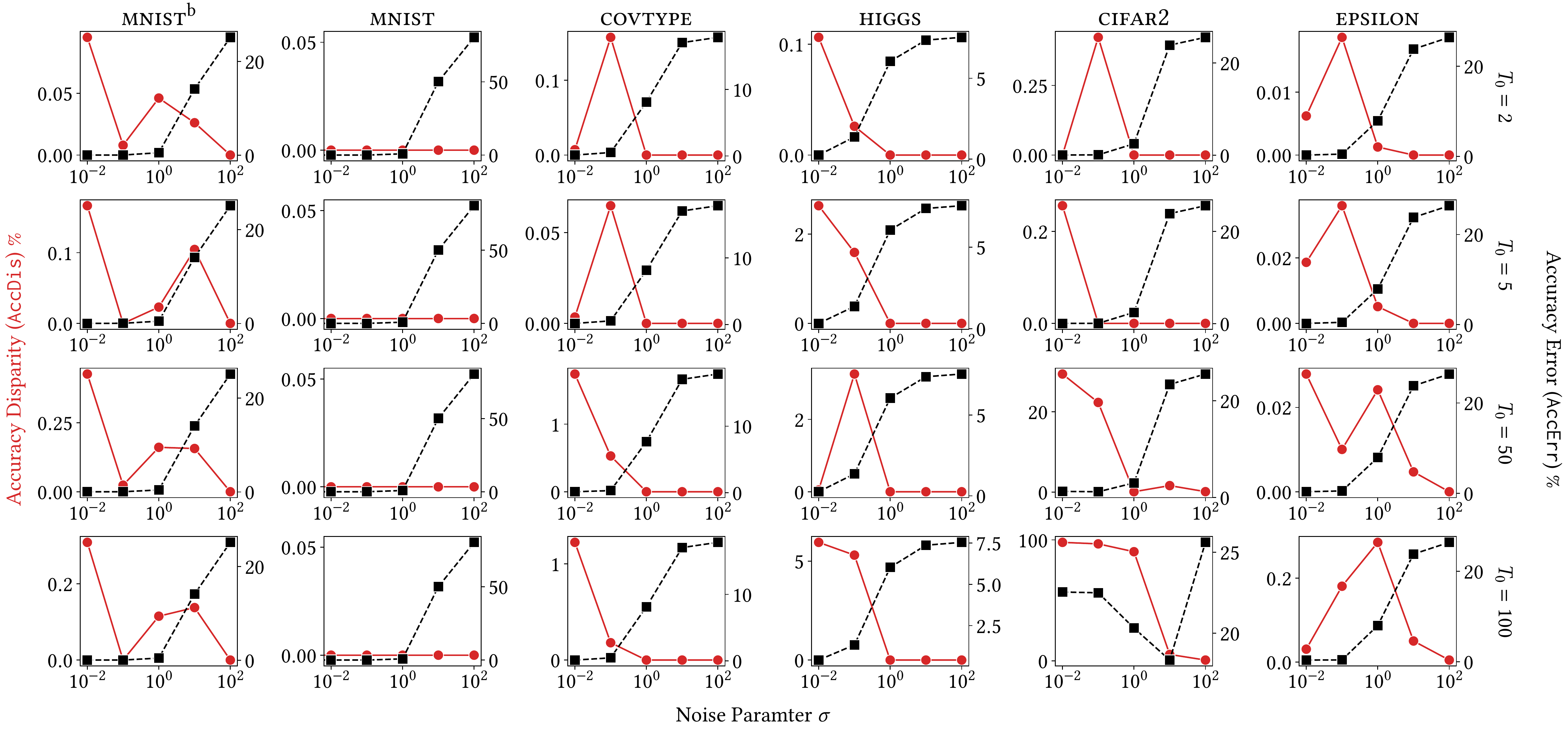}
    \caption{Certifiability-Effectiveness trade-off for \deltagrad method at small deletion volume. Each row corresponds to a value of the \QoA parameter}
    \label{fig:cert-effec-deltagrad-small}
\end{figure*}

\begin{figure*}
    \centering
    \includegraphics[width=\linewidth]{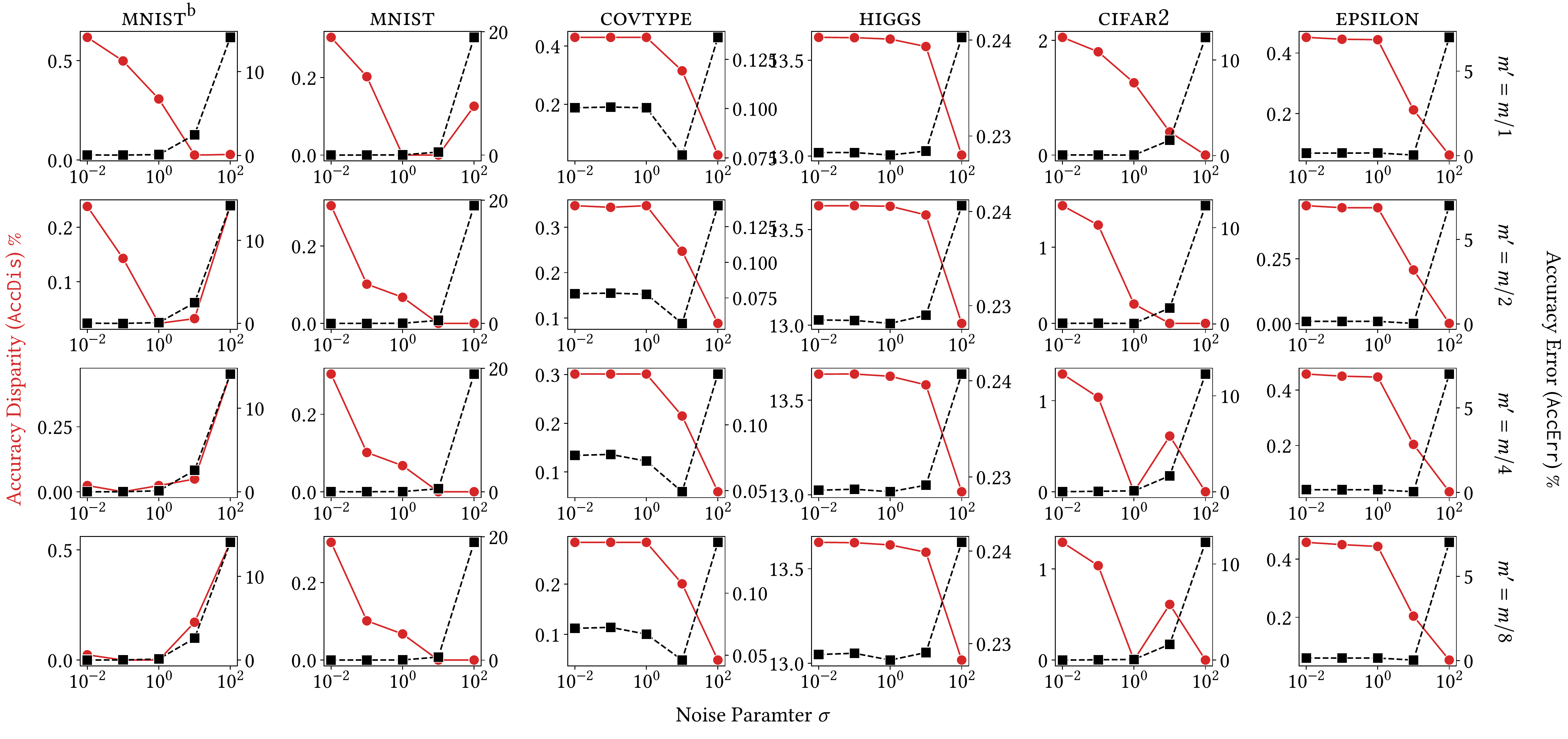}
    \caption{Certifiability-Effectiveness trade-off for \infl method at small deletion volume. Each row corresponds to a value of the \QoA parameter}
    \label{fig:cert-effec-influence-small}
\end{figure*}

In this subsection, \Cref{fig:cert-effec-fisher-small,fig:cert-effec-deltagrad-small,fig:cert-effec-influence-small} correspond to the results for the \fisher, \deltagrad and \infl methods for the smallest volume of deleted data.

\subsection{Medium deletion volume results}
\label{app:cert-effec-medium}
\begin{figure*}
    \centering
    \includegraphics[width=\linewidth]{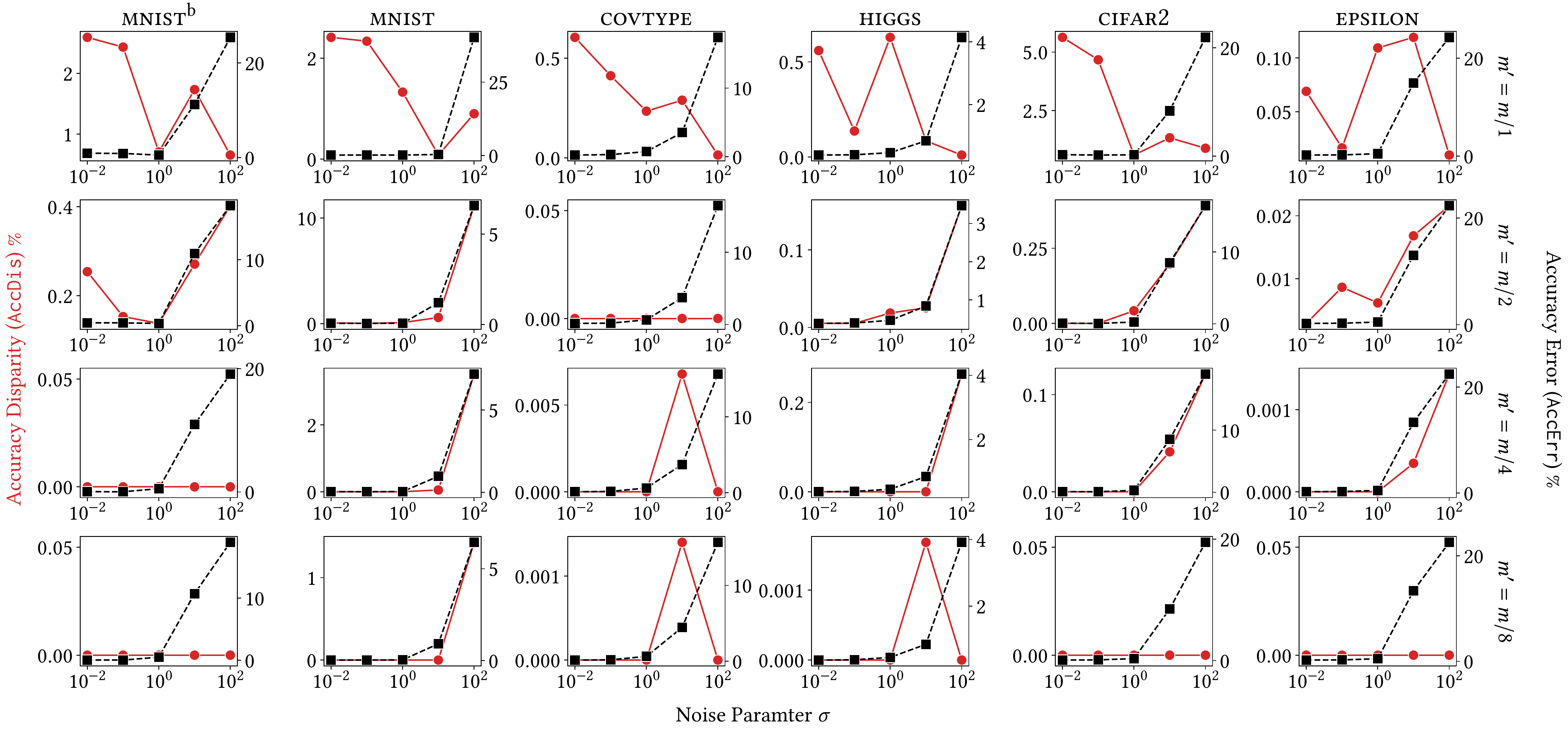}
    \caption{Certifiability-Effectiveness trade-off for \fisher method at medium deletion volume. Each row corresponds to a value of the \QoA parameter}
    \label{fig:cert-effec-fisher-medium}
\end{figure*}

\begin{figure*}
    \centering
    \includegraphics[width=\linewidth]{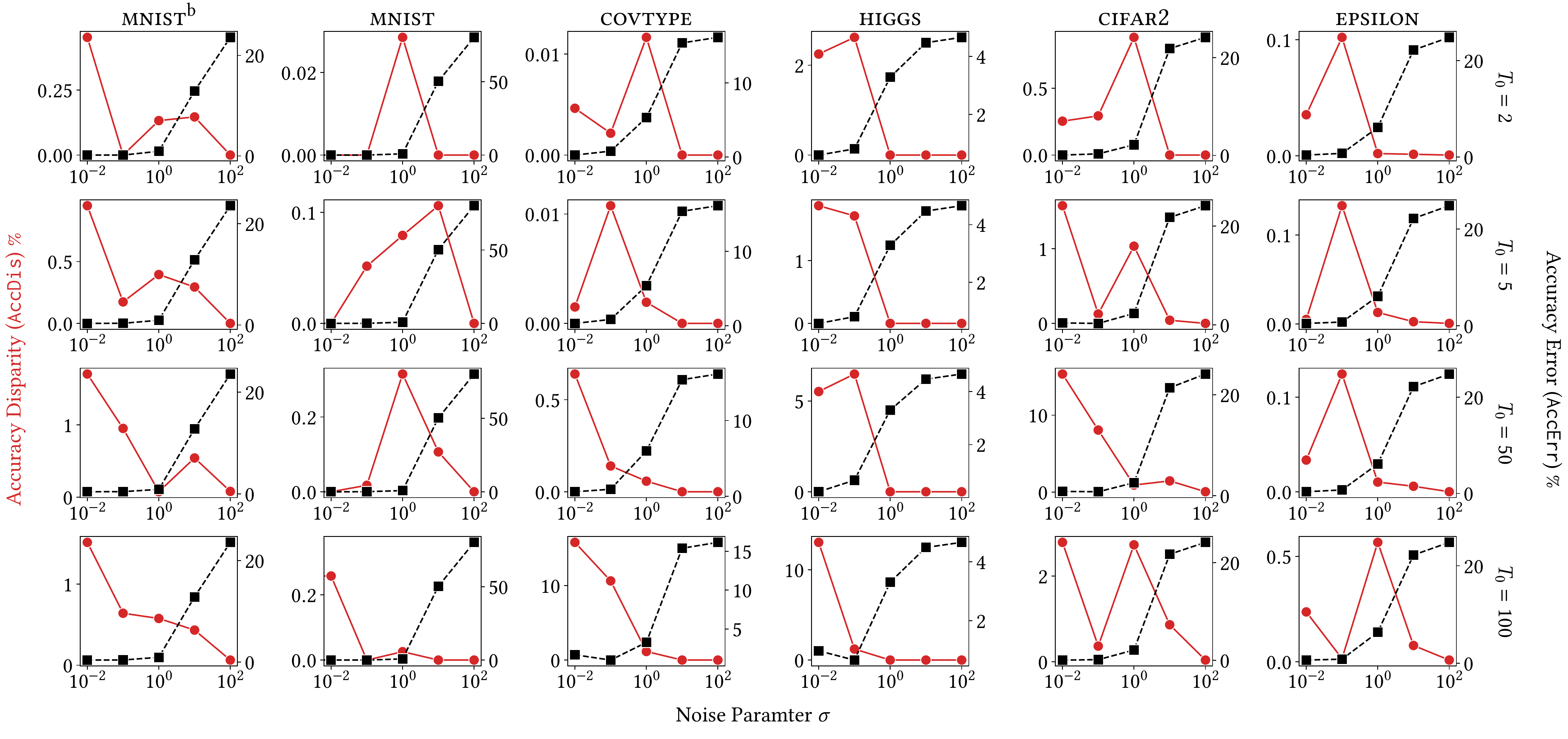}
    \caption{Certifiability-Effectiveness trade-off for \deltagrad method at medium deletion volume. Each row corresponds to a value of the \QoA parameter}
    \label{fig:cert-effec-deltagrad-medium}
\end{figure*}

\begin{figure*}
    \centering
    \includegraphics[width=\linewidth]{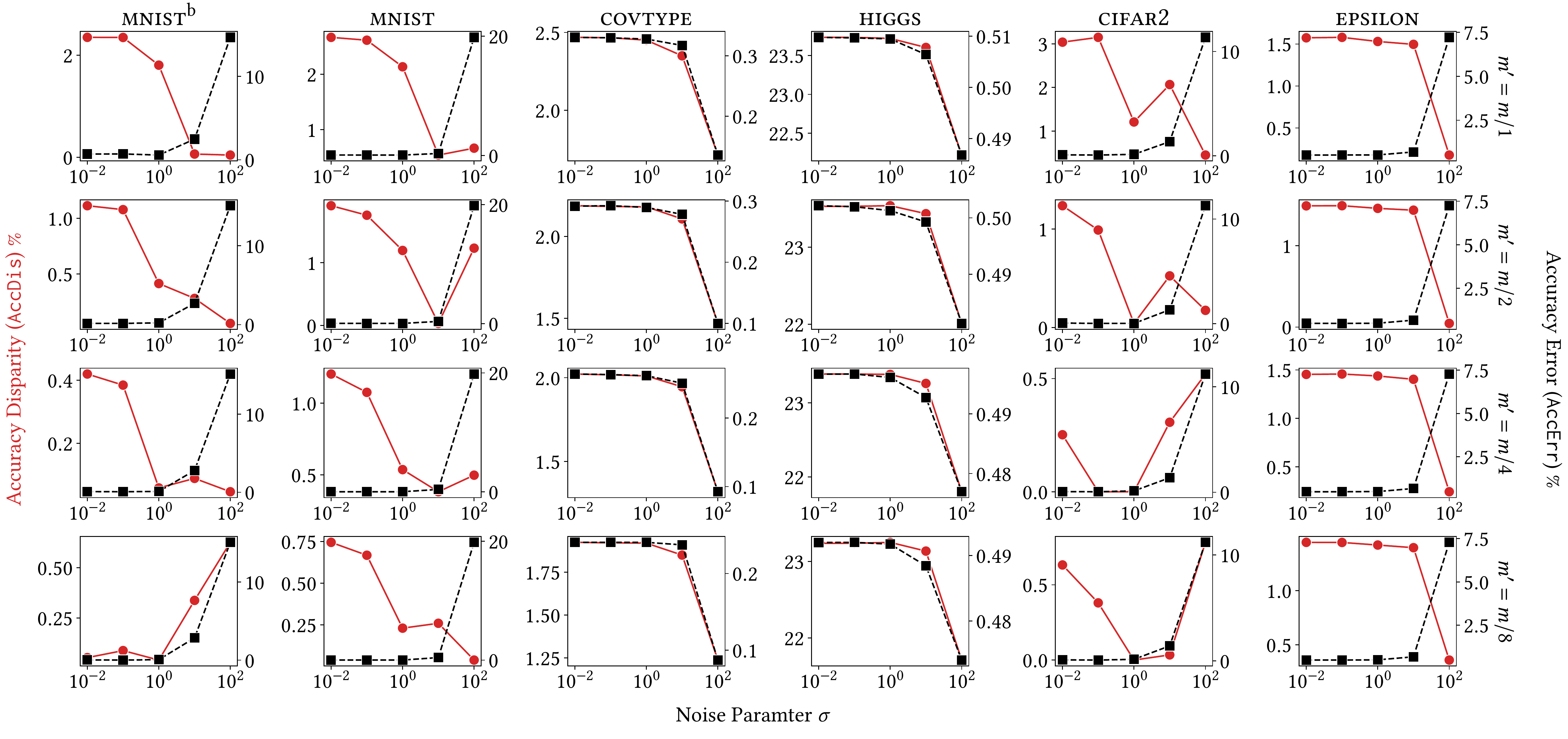}
    \caption{Certifiability-Effectiveness trade-off for \infl method at medium deletion volume. Each row corresponds to a value of the \QoA parameter}
    \label{fig:cert-effec-influence-medium}
\end{figure*}

In this subsection, \Cref{fig:cert-effec-fisher-medium,fig:cert-effec-deltagrad-medium,fig:cert-effec-influence-medium} correspond to the results for the \fisher, \deltagrad and \infl methods for the medium volume of deleted data.

\subsection{Large deletion volume results}
\label{app:cert-effec-large}
\begin{figure*}
    \centering
    \includegraphics[width=\linewidth]{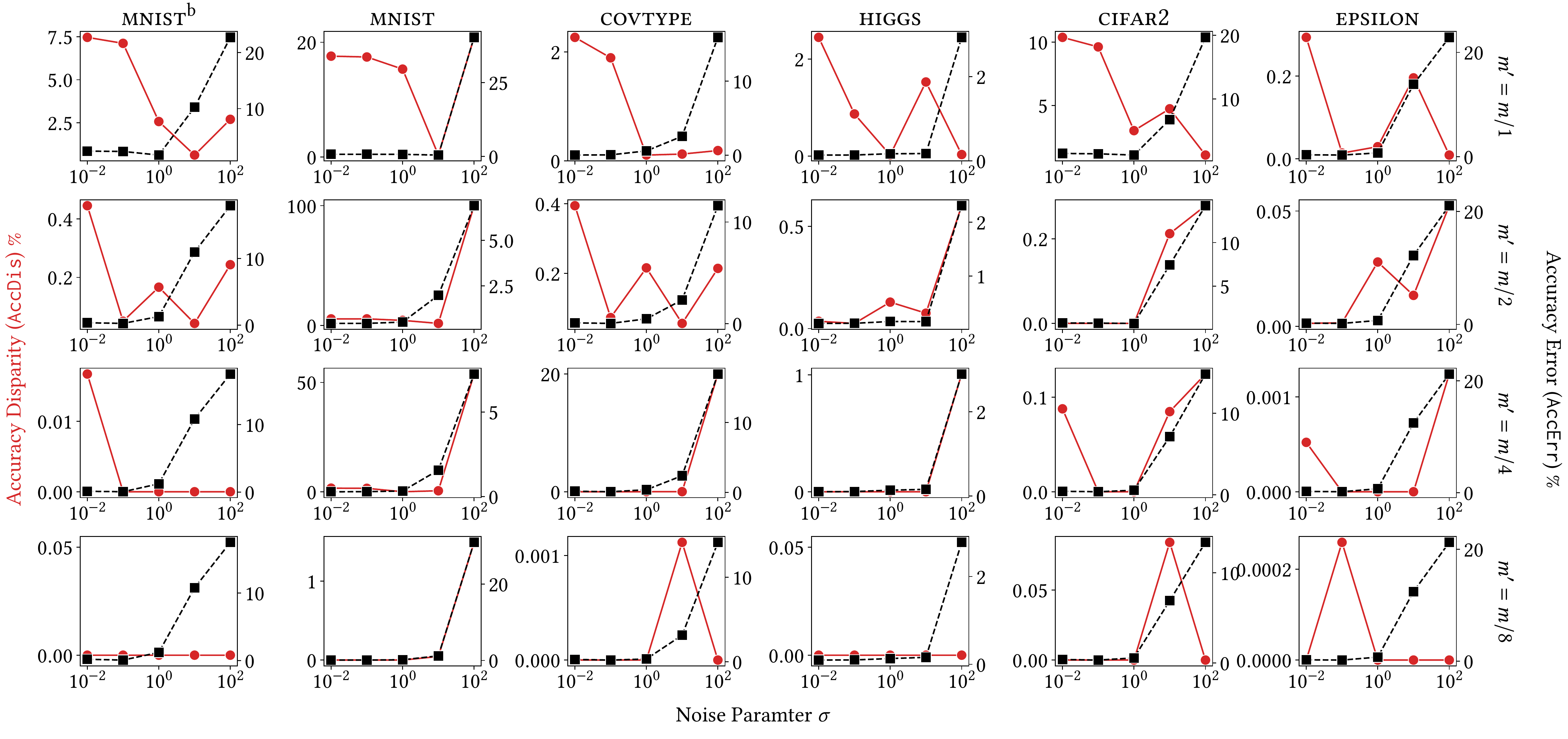}
    \caption{Certifiability-Effectiveness trade-off for \fisher method at large deletion volume. Each row corresponds to a value of the \QoA parameter}
    \label{fig:cert-effec-fisher-large}
\end{figure*}

\begin{figure*}
    \centering
    \includegraphics[width=\linewidth]{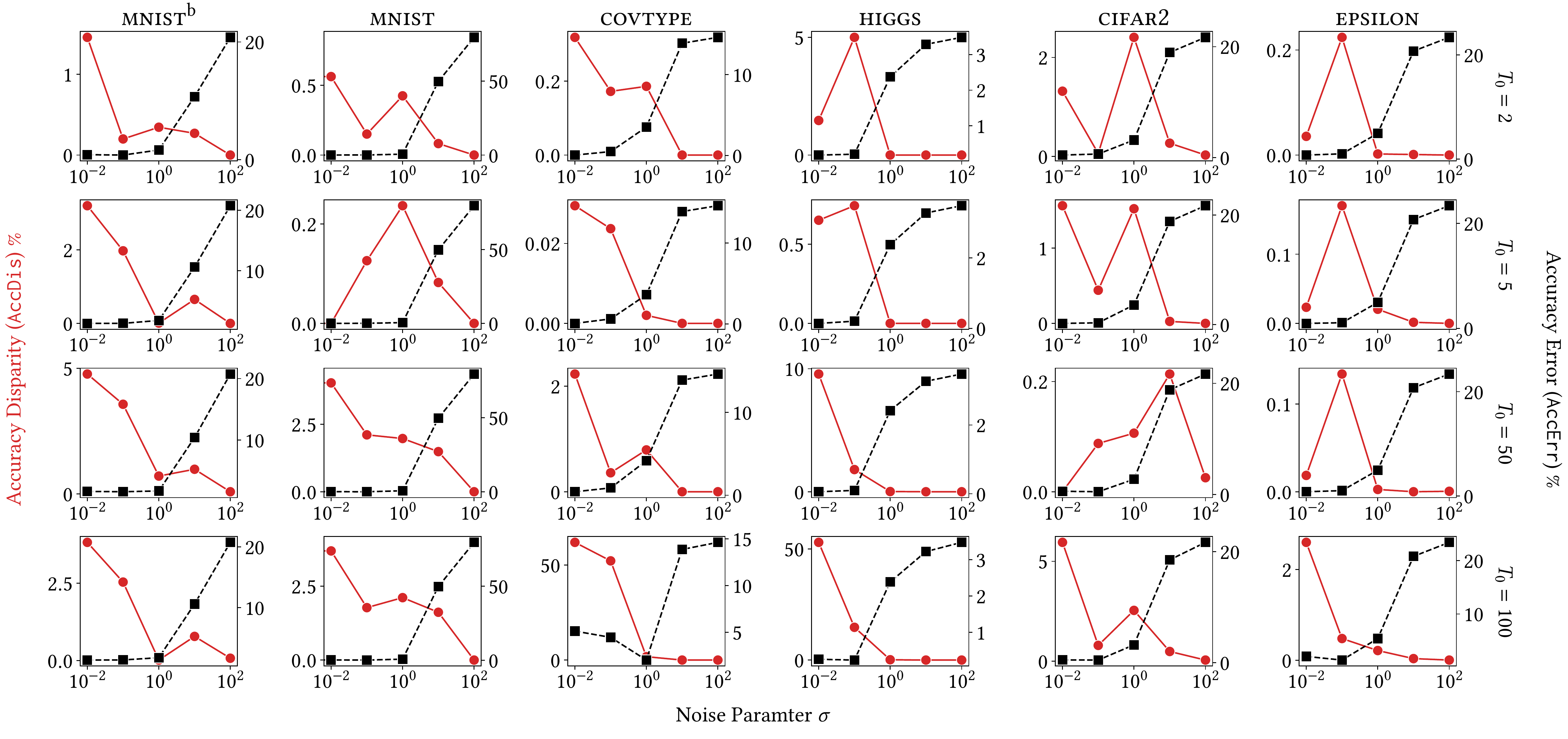}
    \caption{Certifiability-Effectiveness trade-off for \deltagrad method at large deletion volume. Each row corresponds to a value of the \QoA parameter}
    \label{fig:cert-effec-deltagrad-large}
\end{figure*}

\begin{figure*}
    \centering
    \includegraphics[width=\linewidth]{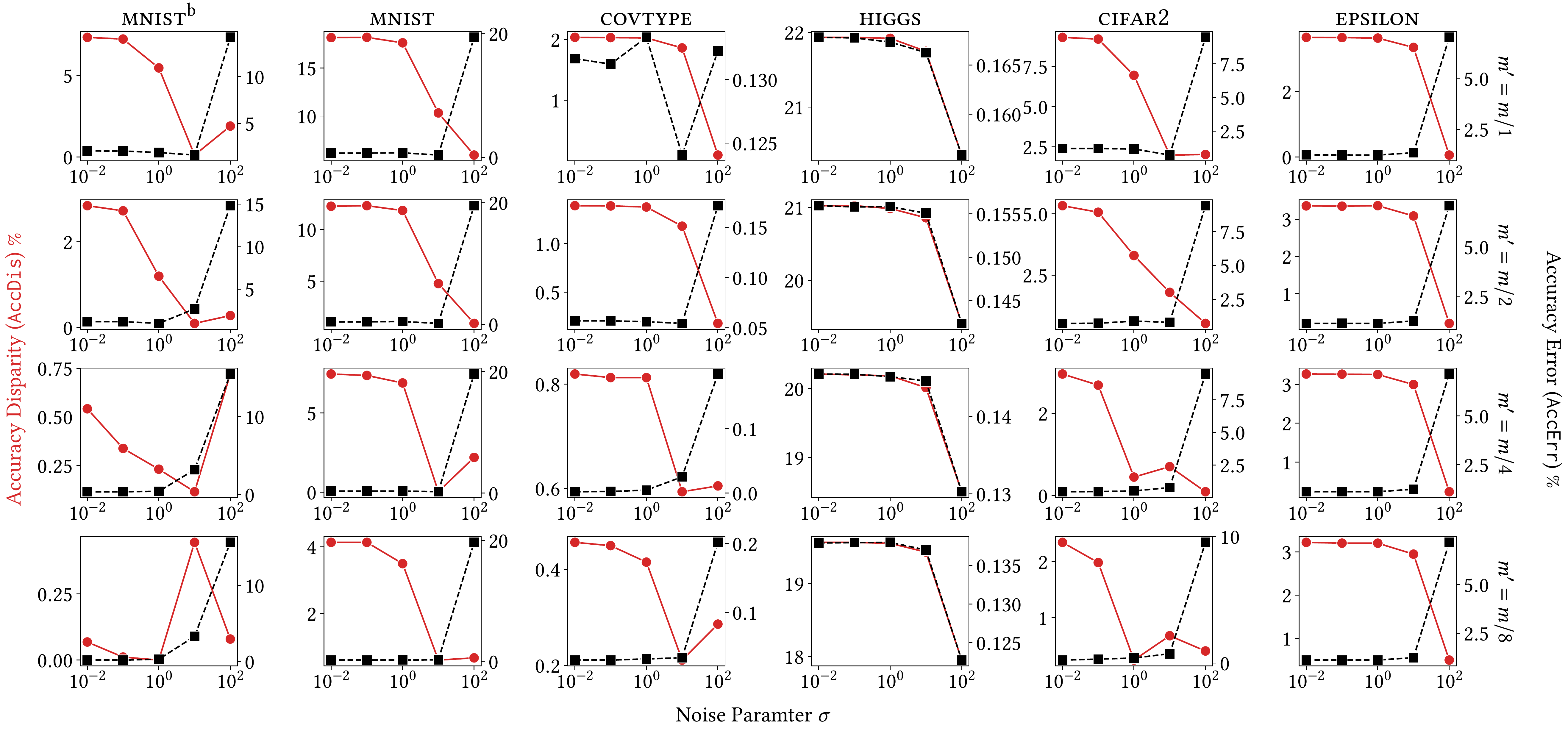}
    \caption{Certifiability-Effectiveness trade-off for \infl method at large deletion volume. Each row corresponds to a value of the \QoA parameter}
    \label{fig:cert-effec-influence-large}
\end{figure*}

In this subsection, \Cref{fig:cert-effec-fisher-large,fig:cert-effec-deltagrad-large,fig:cert-effec-influence-large} correspond to the results for the \fisher, \deltagrad and \infl methods for the largest volume of deleted data.

